\documentclass{article}

\usepackage{jfrExamplee}
\usepackage{apalike}
\let\bibhang\relax
\usepackage[sort&compress,comma,authoryear]{natbib}
\usepackage{setspace}
\usepackage{subcaption}

\usepackage{booktabs}
\usepackage{balance}
\usepackage{xcolor}
\usepackage{tabularx}
\usepackage{graphicx}\usepackage{amsmath}
\usepackage{amssymb}
\usepackage{listings}
\usepackage{textcomp}
\usepackage{url}
\usepackage{multirow}
\usepackage{hyperref}
\usepackage{cleveref}
\usepackage{wrapfig}
\crefformat{footnote}{#2\footnotemark[#1]#3}
\usepackage{ragged2e} 
\usepackage[subnum]{cases}
\newcolumntype{Y}{>{\RaggedRight\arraybackslash}X} 
\usepackage{algorithm}
\usepackage{bm}
\usepackage[noend]{algpseudocode}
\usepackage{hyperref}
\usepackage[toc,page]{appendix}
\usepackage{float}
\usepackage{tikz}
\usetikzlibrary{positioning}
\usetikzlibrary{calc}

\usepackage{ragged2e}
\newcolumntype{L}[1]{>{\raggedleft\arraybackslash}p{#1}}
\newcolumntype{R}[1]{>{\raggedright\arraybackslash}p{#1}}
\newcolumntype{K}[1]{>{\centering\arraybackslash}p{#1}}

\graphicspath{ {figures/}}

\newcommand{\degree}{$^{\circ}$}
\newcommand{\coo}{\ensuremath{\mathrm{CO_2}}}

\newcommand{\comment}[1]{}

\usepackage{todonotes}

\newcommand{\alloc}[1]{}

\title{Heterogeneous Ground and Air Platforms, Homogeneous Sensing: Team CSIRO Data61's Approach to the DARPA Subterranean Challenge}

\author{
Nicolas Hudson,
Fletcher Talbot,
Mark Cox,
Jason Williams,
Thomas Hines,
Alex Pitt,\\
\textbf{Brett Wood,
Dennis Frousheger,
Katrina Lo Surdo,
Thomas Molnar,
Ryan Steindl,}\\
\textbf{Matt Wildie,
Inkyu Sa,
Navinda Kottege\thanks{All correspondence should be addressed to N. Kottege at \texttt{navinda.kottege@csiro.au.} This research was developed with funding from the Defense Advanced Research Projects Agency (DARPA). The views, opinions and/or findings expressed are those of the author and should not be interpreted as representing the official views or policies of the Department of Defense or the U.S. Government.}, 
Kazys Stepanas, 
Emili Hernandez,}\\
\textbf{Gavin Catt,
William Docherty,
Brendan Tidd,
Benjamin Tam,
Simon Murrell,}\\
\textbf{Mitchell Bessell,
Lauren Hanson,
Lachlan Tychsen-Smith,
Hajime Suzuki,
Leslie Overs}\\
Robotics and Autonomous Systems Group\\
CSIRO, Pullenvale, Queensland, Australia \\
\And
Farid Kendoul, Glenn Wagner, Duncan Palmer, Peter Milani \\
Emesent, Milton, Queensland, Australia
\And
Matthew O'Brien, Shu Jiang, Shengkang Chen, Ronald C. Arkin \\
Georgia Tech, Atlanta, Georgia, USA
}

\begin{document}

\maketitle

\begin{abstract}
Heterogeneous teams of robots, leveraging a balance between autonomy and human interaction, bring powerful capabilities to the problem of exploring dangerous, unstructured subterranean environments. Here we describe the solution developed by Team CSIRO Data61, consisting of CSIRO, Emesent and Georgia Tech, during the DARPA Subterranean Challenge. These presented systems were fielded in the Tunnel Circuit in August 2019, the Urban Circuit in February 2020, and in our own Cave event, conducted in September 2020. A unique capability of the fielded team is the homogeneous sensing of the platforms utilised, which is leveraged to obtain a decentralised multi-agent SLAM solution on each platform (both ground agents and UAVs) using peer-to-peer communications. This enabled a shift in focus from constructing a pervasive communications network to relying on multi-agent autonomy, motivated by experiences in early circuit events. These experiences also showed the surprising capability of rugged tracked platforms for challenging terrain, which in turn led to the heterogeneous team structure based on a BIA5 OzBot Titan ground robot and an Emesent Hovermap UAV, supplemented by smaller tracked or legged ground robots. The ground agents use a common CatPack perception module, which allowed reuse of the perception and autonomy stack across all ground agents with minimal adaptation.
\end{abstract}

\section{Introduction}
\label{sec:introduction}
Heterogeneous teams of autonomous robots bring unique capabilities to the problem of exploring unknown, dangerous environments with limited communications, as occurs in subterranean environments or disaster scenarios. Steady progress has been made in relevant areas, showcased through programs such as DARPA Software for Distributed Robotics \citep{HowPar06} and MARS2020 \citep{HsiCow07}, and virtual and physical competition events such as RoboCup Rescue \citep{BalCar07} and the Multi Autonomous Ground-robotic International Challenge (MAGIC 2010) \citep{ButDan12}. The DARPA Subterranean Challenge \citep{darpasubt}, herein referred to as \textit{SubT}, extends these previous developments with a focus on autonomy, made essential by the underground environment, which severely limits communications, and increasing degrees of difficulty in distance, urban obstacles such as stairs, and extreme and ambiguous traversability presented in natural caves. A single operator must control the fleet of robots to complete a mission. The goal is to find as many artefacts, e.g., survivors, backpacks, cell phones as possible within the limited time of one hour.

This paper describes the development of the Team CSIRO Data61\footnote{Comprising CSIRO, Emesent and Georgia Tech.} entry to SubT, its deployment in the Subterranean Integration Exercise (STIX), Tunnel and Urban Circuit events, and the domestic Cave event held in Northern Australia in lieu of the Cave Circuit which was cancelled due to the COVID-19 pandemic. 

Teams of robots have advantages in speed of executing a distributed task, and robustness to failure of any single agent \citep{ParRus16}. Heterogeneous teams additionally combine the strengths of different platforms, and allow more capable (and expensive) platforms to be reserved for the tasks that they can uniquely complete \citep{Par99}. When combined with shared, distributed maps, homogeneous sensing strengthens the ability for agents to continue to act independently and autonomously when communications to both the operator base station and to other agents are disrupted.

The CSIRO Data61 team has deployed a heterogeneous team of quadrotor UAVs (developed by Emesent), and a variety of tracked, quadruped and hexapod UGVs as described in Section~\ref{sec:systemdescription}, with each bringing their own strengths. For example, the terrain able to be traversed by the large tracked UGV in cave environments significantly exceeded the smaller UGVs, while the latter were able to pass through narrower doorways and stairwells in the Urban Circuit. The UAV's higher speed enables a rapid overview of an area, but have been found to be particularly effective when deployed from the large UGV in an area identified to have promising structures that are inaccessible from the ground. 

We have approached the problems posed by heterogeneous environments with a heterogeneous team of robots, but with homogeneous sensing capabilities. All robots in the team utilise a common sensor suite consisting of a spinning lidar, and cameras, all utilising CSIRO's Wildcat real-time multi-agent Simultaneous Localisation and Mapping (SLAM) solution, and 3D frontier-based exploration algorithm \citep{williams_2020}. The UGVs utilise a common local navigation pipeline optimised for challenging unstructured terrain and negative obstacles \citep{hines_virtual_2021}, while the UAVs use Emesent's commercially available navigation system. The component autonomy systems used on each agent are discussed in Section~\ref{sec:singleagent}.

Coordinating multiple agents, and providing situational awareness to the operator is underpinned by a shared global map, upon which all other shared information is referenced, described in Section \ref{sec:multiagent}. Maps are shared through a peer-to-peer communications network, with each agent independently solving for a global map, based on the currently received subset of shared SLAM frames generated by each agent during exploration. While there is no guarantee agents' global maps are the same (i.e., agents may have a different set of frames at a particular instant, and receive frames from different agents in different orders), frame overlap guaranteed from a common starting area allows each agent to incrementally construct its own global solution based on information from all agents, enabling an agent to interpret information (such as frontiers, artefact detections, or traversability data) referenced to any other agent's SLAM frame.  

A human operator guides the robot team, described in Section \ref{sec:humanrobotteam}. The system can be commanded using various levels of autonomy, ranging from single agent teleoperation, to a consensus-based Multi-Robot Task Allocation (MRTA) approach, which can coordinate exploration between agents with no input from the operator.  The human supervisor brings unique abilities in confirming artefact detections and classes, commanding single agents, such as deciding when to launch the UAV, and influencing MRTA using region-based priorities. 

The human robot team had increasing levels of autonomous operation in each successive DARPA event, discussed in Section \ref{sec:discussion}. Based on experiences at each event, effort was reduced in developing pervasive communication solutions and complex legged UGV systems, and instead focused on simpler, robust systems with more autonomy. At STIX, robots were commanded with manual waypoints. At the Tunnel Circuit, an autonomous exploration capability was fielded, but the operators tended to prefer manual waypoints which had more time in testing. The need for autonomy was noted, and at the Urban Circuit, the explore-sync behaviour \citep{williams_2020} was the primary mode of operation, accounting for the majority of detected artefacts and distance travelled. In the natural cave environment, our team explored the majority of the terrain that could reasonably be covered by a ground agent using coordinated MRTA. On rough terrain, the physical capabilities of the tracked UGV Platforms still exceeds the ability to distinguish between achievable and unachievable tasks. Ongoing work is being conducted to improve feedback within the layers of the navigation stack, and prioritising terrain difficulty when selecting tasks.

\subsection{Related Work}
\label{ss:related_work}
While research into elements of autonomous systems has progressed steadily, robotic competitions have provided significant impulses in demonstrating the state of the art for integrated systems. Before this trend emerged, a 
significant demonstration was made in the DARPA SDR program. \cite{HowPar06} describes a system in which a team of four lidar-equipped robots explored and mapped a building, and then guided 70 small robots (with limited computing and sensing) into place to perform the task of surveillance, effectively forming a self-deploying sensor network. Each mapping agent solved a single agent SLAM problem (without loop closures), while maps were merged at the base station.

\cite{HsiCow07} considered teams of heterogeneous UAVs and UGVs for building situational awareness, performed as a part of the DARPA MARS2020 program. The system was utilised in three distinct phases, where the fixed-wing UAVs first generated maps, of the region; secondly, the UGVs constructed a communications map of signal strength between pairs of locations; and finally the maps were exploited to search and localise human targets with the aid of a single human operator. Autonomy was based on MissionLab, which permits visualisation, testing and code generation for control of single robots and robot teams, allowing higher level behaviours to be composed (and reused) by combining lower level building blocks.

\cite{BalCar07} presents results of the RoboCup Rescue virtual competition, a simulation based competition for urban search and rescue. Teams were required to explore a building and detect survivors through simulated RFID tags (selected to avoid the burden of computer vision). Scores were assigned based on survivor detection, map quality and exploration completeness. The winning entry, described in \cite{KleZip06}, utilised local grid-based methods, and deployed RFID tags in the environment to aid localisation and for global coordination.

\cite{ButDan12} described work performed in MAGIC 2010. A team of nine UGVs were controlled by two operators to explore a large, unstructured environment, detect static and dynamic objects of interest, and follow neutralisation procedures requiring cooperation of multiple UGVs for discovered objects. The UGV fleet were equipped with GPS, IMU, lidar and cameras, while only particular platforms had the equipment required for simulated neutralisation. Global navigation, mapping and task and exploration planning were executed on the operator base station, while individual agents localised and navigated using their own maps. Operator controls included permitted/excluded exploration regions for each agent, inter-agent distance limits, and heading biases.

\cite{gregory2016application} worked to explore issues of team management and communication, which were identified as drivers of complexity in both MAGIC 2010 and the Robocup Rescue League. Motivated by humanitarian assistance and disaster relief, the goal was to evaluate damage to infrastructure (such as roads, forming a series static goals), and localise survivors through radio signals (forming dynamic goals, appearing as the signals are first detected). SLAM utilised lidar, 3D camera and GPS measurements separately on each agent, communicating resulting poses. The primary behaviour utilised was GotoRegion, allowing robustness to the precise waypoint location. A higher level GuardedNavigation behaviour additionally defined a safe region (with known good communications) to return to in the case of failed navigation. Two agents were utilised and, motivated by the goal of fully autonomous operation, success was measured as the frequency and duration of the accompanying safety operator.

\cite{Qin2018} considered UGV/UAV exploration of a 3D environment performed in two distinct phases, where the UGV first performed fast autonomous exploration based on a coarser 2.5D active SLAM using a 3D lidar, and the UAV subsequently performed fine 3D mapping using a rotating 2D lidar. The aerial and ground agents utilised different sensors and world representations, but similar approaches for viewpoint selection.

Recent relevant surveys include \cite{LiuNej13}, which considers control and perception for robotic search and rescue in urban environments; and \cite{RecTom18}, which focuses on post-disaster assessment using UAVs.

\subsection{Other SubT Teams}
\label{ss:other_teams}
There are multiple teams competing in the DARPA SubT competition. We briefly outline existing publications from other teams and contrast to our approach; further details of other teams' entries can be found in other papers in this special issue.  

Team Pluto's solution to the Tunnel Circuit is described in \cite{Miller20}. The robot fleet incorporated micro-aerial vehicles (MAVs) and Ghost Robotics Vision60 legged quadrupeds. Due to the limited communications, a focus was put on autonomous exploration, which executed for a fixed maximum duration before returning for further commands using a global topological graph. The operator was able to preplan sequences of turns at a priori unknown tunnel junctions to ensure diversity between missions executed by different agents. A distributed database was used to share data between all agents and the base station.

Team CoSTAR's winning solution from the Urban Circuit is described in \cite{BouFad20}, focusing on the Spot platform, supplementing Spot's inbuilt sensing with a sensor package that incorporates lidar, RealSense cameras, an IMU, gas and wifi detectors, and a thermal camera. The architecture utilises NeBula (networked belief aware perceptual autonomy); key aspects of this work are risk and uncertainty-aware local planning \cite{KimBou21}, yielding a highly capable platform that is able to conduct long-duration, missions beyond communications range. Comparably, our focus has been on establishing a common understanding on all agents, enabling them to coordinate autonomously, out of range of the operator base station.

Team Explorer's winning solution from Tunnel Circuit utilised a combination of large wheeled UGVs and custom UAVs. A modular autonomy pack incorporated sensors including lidar, IMU, RealSense and thermal camera, and associated compute. Submaps are shared via a ledger to avoid revisiting the same region.\footnote{https://www.youtube.com/watch?v=PVRG4Nc3hXE}

Team Cerberus's approach to exploration using aerial scouts and ANYmal legged robots is described in \cite{DanTra20}. The method combines a local planner, which utilises a rapidly exploring random graph, with a global planner that is built incrementally based on selected branches of the local graph, and is activated when a dead end is reached. Agents operate independently given a bounded volume of interest, and the planned paths are constantly checked to ensure that sufficient battery life remains to return to the base.

\cite{OhrMil20} describes aspects of Team Marble's approach to the tunnel event, achieving exploration using ground-based agents using a topological graph combined with reactive control for path centring and obstacle avoidance. 

Finally, \cite{HuaLu19} describes the innovative robust blimp used by team NCTU in the tunnel event. The system incorporates perception, localisation and autonomy into a lightweight computing platform.

\begin{figure}[!b]
    \centering
    \includegraphics[width=120mm]{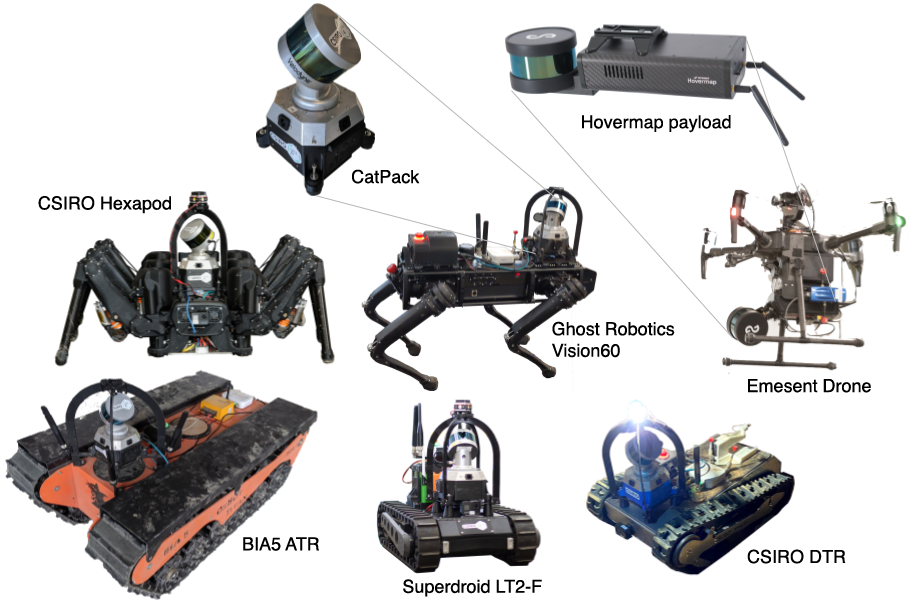}
    \caption{Team CSIRO Data61 deployed a set of heterogeneous ground and air robots, with homogeneous sensing payloads. \textit{Top Row}: CatPack and Hovermap modular sensing payloads, both using a rotating lidar system for the UGVs and UAVs. \textit{Middle Row}: CSIRO Hexapod, Ghost Robotics Vision60 quadruped, and a Emesent UAV with the Hovermap payload. \textit{Bottom Row}: BIA5  All Terrain Robot (ATR), SuperDroid LT2-F, and CSIRO Dynamic Tracked Robot (DTR).}
    \label{fig:robot_fleet}
\end{figure}

\subsection{Contributions}
\label{ss:contributions}
This paper describes the Team CSIRO development in SubT. Specifically, we make the following contributions:
\begin{itemize}
\item We describe the development and utilisation of our heterogeneous set of ground robots, and Emesent UAVs. Importantly, these are paired with \emph{homogeneous sensing packs} (respectively, CatPack and Hovermap), which utilise equivalent sensors, and common SLAM and 3D frontier logic. 
\item In Section \ref{sec:catpack} we describe the design of the CatPack perception module, which maximises robustness, and enables reuse of the entire perception and autonomy stack across all ground robots.
\item In Section \ref{ss:slam}, we describe the operation of our Wildcat SLAM system, which provides real-time operation and accuracy that \emph{won most accurate artefact detection} at the Urban Event (in relation to the gate, which is detected in one robot and reported to others using the shared SLAM solution). 
\item We describe our mechanism for peer-to-peer map sharing, with each agent solving for a global map without dependence on the centralised operator base station, for both ground and air platforms. This is enabled by homogeneous sensing, and was the only system deploying this capability at the Urban Event.\footnote{as per the Urban Technical Interchange Meeting, https://youtu.be/ymBrec7HY4A}
\item The map sharing establishes \textit{frames}, stable coordinate systems through which frontiers, artefacts and traversability data can be detected on one robot, and shared (again, peer-to-peer) with another. Again, this is enabled by homogeneous sensing and mapping.
\item In Section \ref{sss:multiagentglobal}, we propose global navigation using a topometric map, obtained by sequentially combining local navigation costing data and segmenting using superpixel methods, and referencing data to SLAM frames. Again, this is shared peer-to-peer among ground robots, enabling one agent to navigate to frontiers or artefacts detected by another.
\item In Section \ref{ss:mule}, we describe our disruption-tolerant, peer-to-peer communications solution, which enables cooperation of robots in the challenging subterranean communication environment. This is planned to be released as open source software in the near future.
\item In Section \ref{sec:humanrobotteam}, we outline our system for multi-agent task allocation, exploiting the homogeneous sensing and mapping capability, accommodating the highly continuously moving tasks of exploration, and assigning tasks according to platforms' heterogeneous capabilities and operator input on task priorities.
\item Finally, in Section \ref{sec:discussion}, we discuss system usage, operator interaction and platform choice based on deployment to STIX, Tunnel, Urban, Cave environments in the SubT challenge.  We discuss lessons learned in the various events, and their influence on subsequent stages of the development program.
\end{itemize}

\section{System Description}
\label{sec:systemdescription}

The DARPA SubT Challenge allowed for a fleet of robots to be controlled by a single operator, with little to no prior knowledge of the course. A wide variety of terrain was encountered in the Urban, Tunnel and Cave Circuits, including multiple levels connected with stairs and shafts, tunnels spanning multiple kilometres with constrained passageways, and complex irregular natural formations.

A heterogeneous fleet of robots was use to overcome the diverse set of challenge elements (Figure~\ref{fig:robot_fleet}), and included UAV and UGV systems. The composition of the robot fleet was changed for each event, based on performance testing and the expected challenges in the event. A modular sensing payload, the \textit{CatPack}, along with common power distribution boards and emergency stop systems, reduced the complexity of maintaining multiple systems. In the Urban an Cave circuits the UAVs were mounted on the large tracked platforms, and launched in the field. This enabled the UAV to be carried through narrow openings and conserved battery life. Table \ref{tab:ugv_capabilities} provides insight into platform capabilities, and some of the compromises in using each system.

The robot fleet used a mesh network to provide communications between each agent and the base station. The tracked UGVs additionally carried \textit{drop nodes}, deployable elements of the mesh network, to maintain connection as communications degrades in the underground environment. The higher payload capacity of the larger tracked robots meant it could carry up to eight drop nodes, as used in the Tunnel Circuit. 

The following sections describe the CatPack and each of the deployed heterogeneous robots in more detail.

\begin{table}[t]
\centering
\begin{tabular}{lcccccc}
\toprule
& \multicolumn{6}{c}{Platforms}\\
\cmidrule(lr){2-7}
 \multirow{2}{*}{Attribute}  & CSIRO & GR & BIA5 & SuperDroid & CSIRO & Emesent  \\ 
   & Hexapod & Vision60 &  ATR & LT2-F & DTR  & UAV  \\ 

\midrule
Weight (kg) & 53.7 & 48.4 & 90 & 25 & 25 & 10 \\
Width (m) & 1.08 & 0.55 & 0.78 & 0.46 & 0.51 & 1.07 \\
Length (m) & 1.0 & 0.93 & 1.4 & 0.58 & 0.8 & 1.07  \\
Speed (ms$^{-1}$) & 0.5 & 0.75 & 0.75 & 0.5 & 1.0 & 2.0 \\
Max Step Height (m) & 0.2 & 0.25\footnotemark & 0.25 & 0.12 & 0.19 & n/a \\
Max Slope (\degree) & x & x & 43 & 30 & 44 & n/a \\
Constrained Passage Width (m) & 1.5 & 1.0 & 1.0 & 0.75 & 0.75 & 2.0 \\
\bottomrule
\end{tabular}
\caption{Platform Capabilities: Weight includes batteries and sensor payloads; Width and Length is standing for the legged systems, and blade tip-to-tip for the quadrotor; Speed is that used in the planner, or the default gait for legged systems; Max Step Height was found by empirical testing; Max slope is with a full payload; Constrained Passage Width is the minimum width the vehicle was tested to reliably traverse autonomously.}
\label{tab:ugv_capabilities}
\end{table}

\subsection{CatPack}
\label{sec:catpack}

\begin{figure}[ht]
    \centering
    \begin{subfigure}{.25\linewidth}
    \centering
    \includegraphics[height=30mm]{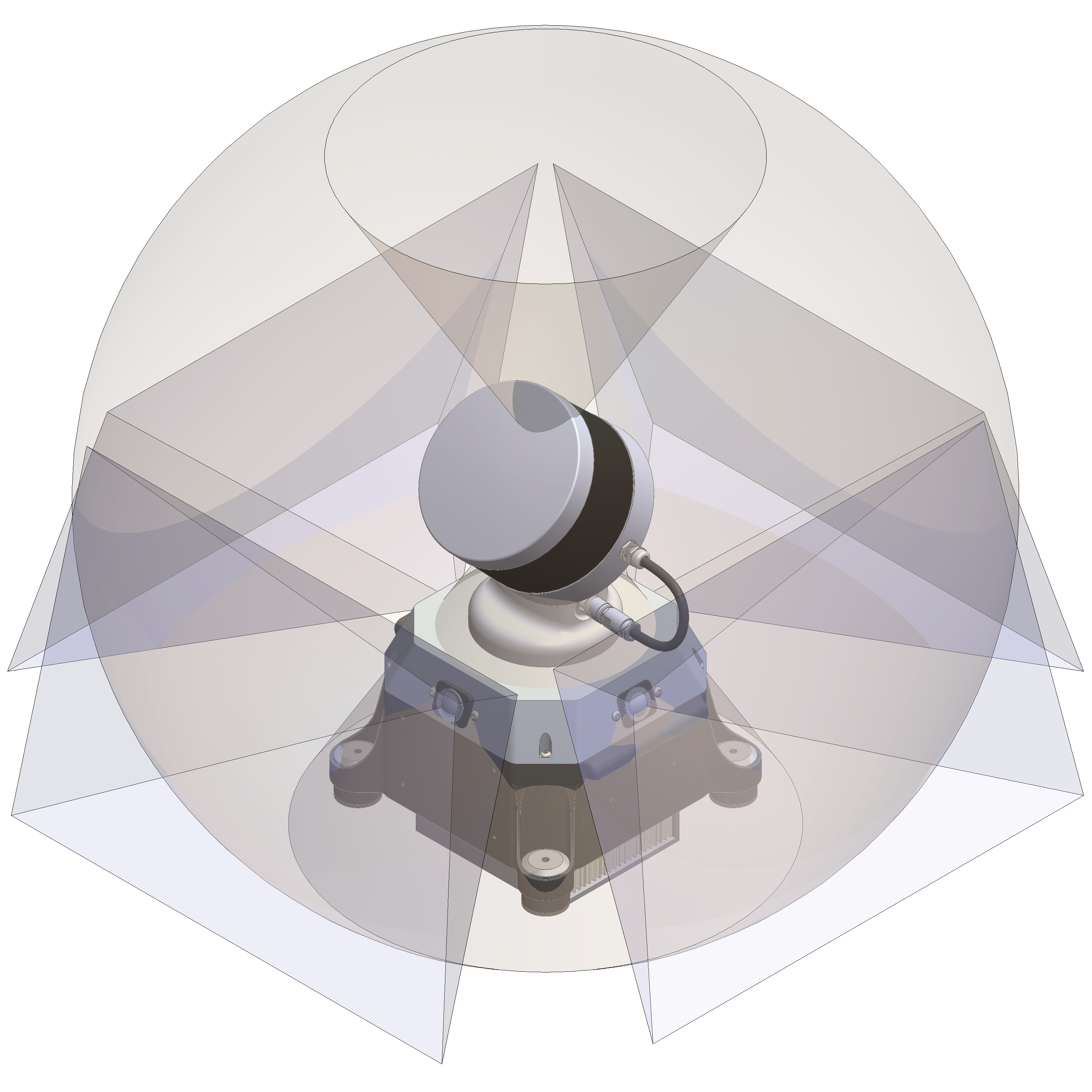}
    \caption{}
    \label{fig:pp2_iso_view_cam_LiDARa_fov}
    \end{subfigure}
    \begin{subfigure}{.25\linewidth}
    \centering
    \includegraphics[height=30mm]{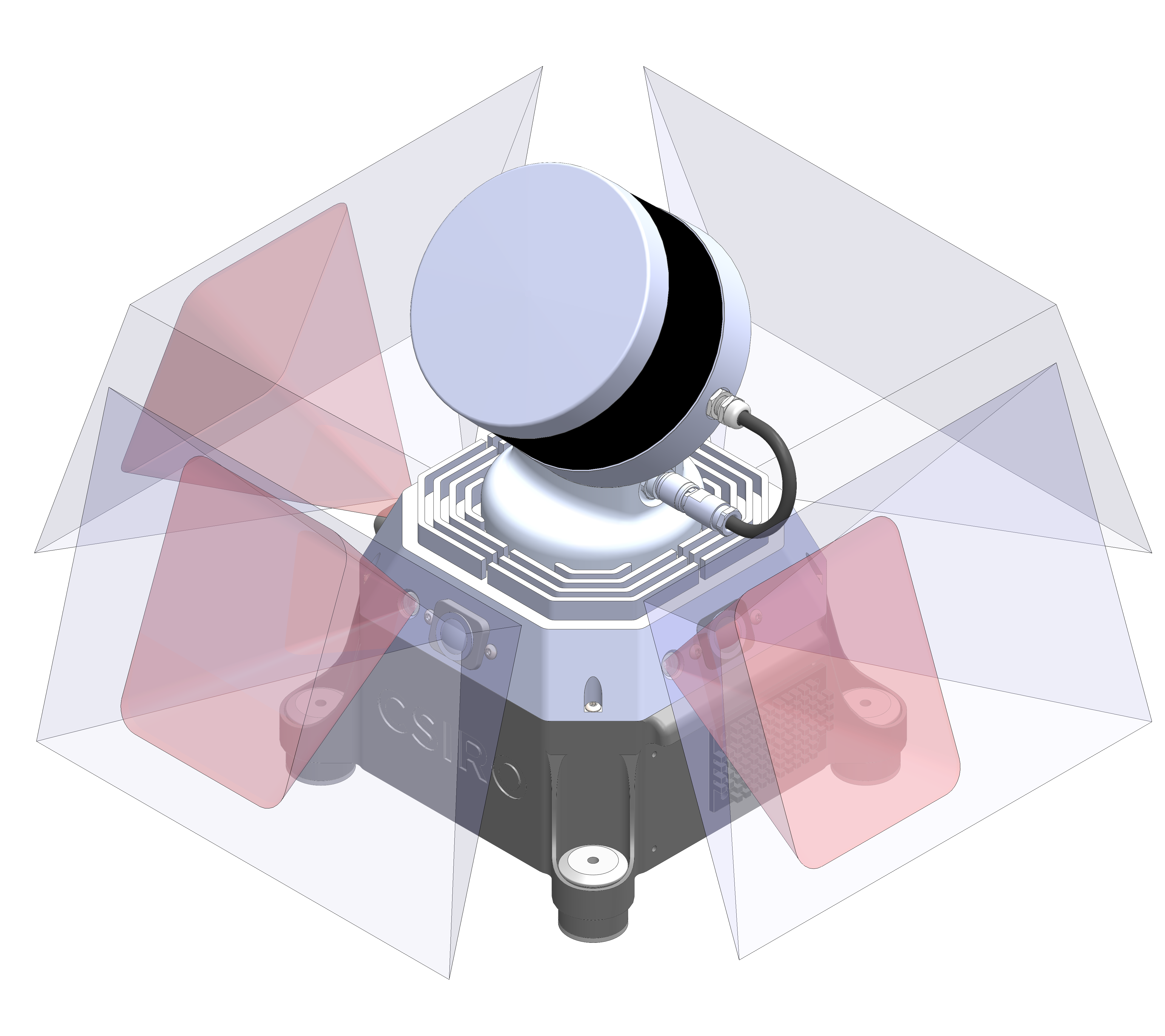}
    \caption{}
    \label{fig:thermal_camera_fov}
    \end{subfigure}
    \begin{subfigure}{.25\linewidth}
    \centering
    \includegraphics[height=30mm]{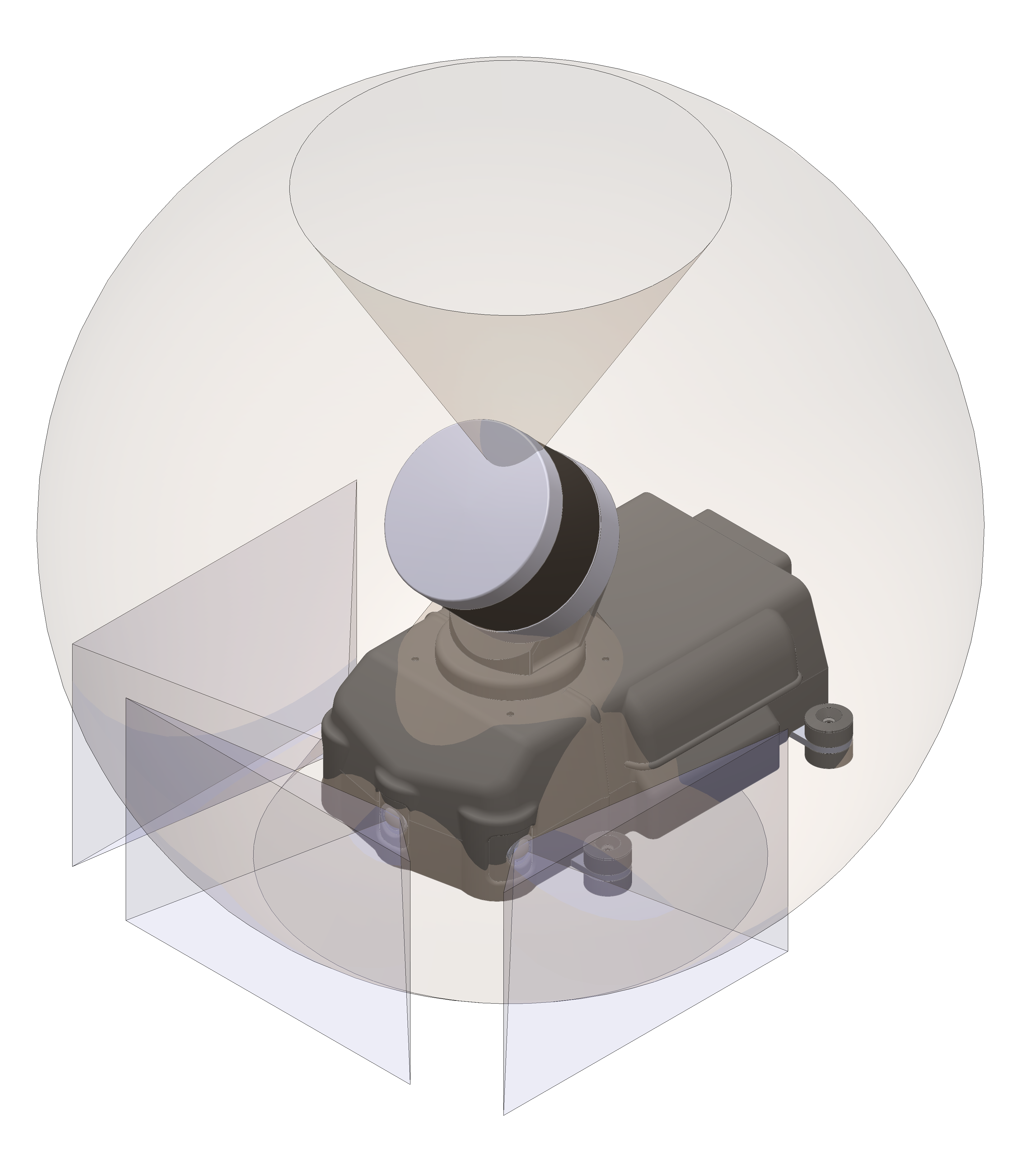}
    \caption{}
    \label{fig:stix_catpack_fov}
    \end{subfigure}
    \caption{The CatPack showing FoV of the lidar and RGB cameras for the newer configuration utilised in Tunnel, Urban and Cave events in (a), with the thermal camera FoV shown in red in (b) and the older configuration utilised in STIX with narrow FoV RGB cameras in (c).}
    \label{fig:catpack_fov}
\end{figure}

\begin{figure}[t]
    \centering
\includegraphics[width=100mm]{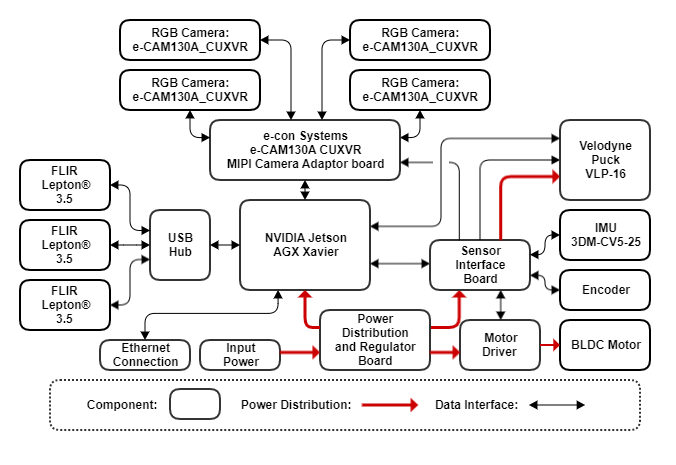}
\caption{High level system architecture of the CatPack used on all the UGV platforms. Power distribution is shown with red arrows, whereas data interfaces are shown with black arrows.}
    \label{fig:perception_pack_system_diagram}
\end{figure}

The ability to maintain multiple UGV platforms, and integrate new systems quickly, is enabled by a modular perception system called the CatPack. This was custom-developed in response to a gap in the capability of commercially available offerings at the time of design. This contains all the visual sensors used by the UGVs including multiple RGB and thermal cameras, a spinning lidar and associated compute. Figure~\ref{fig:catpack_fov} shows the field of view (FoV) of each of the sensors and the high level system architecture of the CatPack is illustrated in Figure~\ref{fig:perception_pack_system_diagram}.

\footnotetext{This is the vertical distance between the robot's feet and the ground while walking and was not tested.}

The cameras and lidar are calibrated to have pixel-wise alignment (Section \ref{ss:perception}), and co-locating them on the same sensor head reduces parallax and allow the CatPacks to retain calibration if shifted between platforms. The CatPacks use a spinning lidar configuration to provide dense depth measurements with an effective 120\degree~vertical field of view (FOV). The Velodyne VLP-16 lidar has a 30\degree~vertical FOV, and is mounted at 45\degree~off horizontal and spun about the CatPack's vertical axis. The LORD Microstrain 3DM-CV5-25 inertial measurement unit (IMU) is mounted to coincide with this rotation axis. These CatPacks are strategically mounted near the front of the UGVs to minimise occlusion of the lidar by the vehicle body, which allows returns from the ground very close to the front of the robot. On the smaller UGVs, this is limited by the lidar's minimum return distance of 0.6\,m. The high FOV enables the same sensor to be used for both SLAM (long range) and perceiving the near field for navigation and avoiding negative obstacles (Section \ref{ss:ugvnavigation}).

\begin{figure}[!b]
    \centering
\includegraphics[width=100mm]{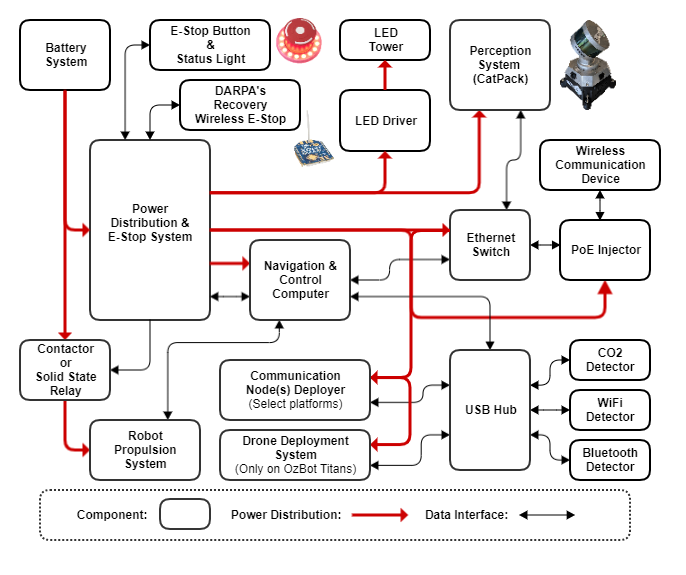}
\caption{The high level systems architecture used on all UGV platforms. Power distribution is shown with red arrows, whereas data interfaces are shown with black arrows.}
    \label{fig:ugv_system_diagram}
\end{figure}

The CatPacks use four 94.9\degree H $\times$ 71.2\degree V FOV RGB cameras (e-con Systems e-CAM130A CUXVR) 
to form a near continuous 360\degree~horizontal FOV. Matching the FOV of the cameras and the lidar (Figure~\ref{fig:catpack_fov}) simplifies coverage planning for the UGVs, meaning that dense coverage is achieved in lidar based exploration (Section \ref{ss:exploration}) and simplifies camera coverage for artefact detection (Section \ref{ss:perception}) as well. At the STIX event, the prototype packs only had three cameras (Figure~\ref{fig:stix_catpack_fov}), and artefacts were missed behind the vehicle as it moved past alcoves. The RGB cameras are interfaced via MIPI in order to exploit the Xavier's dedicated hardware interface, and to avoid overloading the USB bus. The FLIR Lepton 3.5 thermal cameras are lower resolution so this is less of a concern; additionally, the thermal cameras have not been utilised to date as they have not been found necessary to detect the artefacts.

\subsection{UGV Platforms}
\label{ss:ugvhardware}

The UGV systems deployed in the challenge were chosen based on a compromise of capability, size, complexity, payload, robustness and cost. Legged systems promise extreme capability with a small or adaptable vehicle footprint. Tracked (or wheeled) platforms are comparatively simple, robust, inexpensive and with a high payload capacity, but require a larger footprint to be able to traverse comparable step heights or discontinuities.

Each of the UGV platforms that are deployed use the CatPack (Section~\ref{sec:catpack}) as its primary sensor for perception, localisation and mapping. They also share the same high level system architecture that includes a common power distribution and e-stop system, wireless communications system and a navigation and control computer system as illustrated in Figure~\ref{fig:ugv_system_diagram}. The following sections provide more details about each of the UGV platforms while a summary of their capabilities was given earlier in Table~\ref{tab:ugv_capabilities}.

As discussed with each platform, the trajectory of platform selection and investment in custom development over time reflects identified gaps in the capability of available systems. Significantly, the emergence of the high TRL rugged tracked robots, and high TRL quadruped systems, caused reflection on the necessity of the custom hexapod development.

\subsubsection{SuperDroid LT2-F}
\label{sec:SuperDroid}
The SuperDroid LT2-F platform is a configurable off-the-shelf tracked robot kit. It is small enough to traverse small doorways and can be checked in as luggage on international flights, whereas the other larger robots required to be shipped. 
The system was fitted which our own electronics, shown in Figure~\ref{fig:ugv_system_diagram}, including an Intel NUC navigation and control computer running our autonomy stack.
Between the Tunnel and Urban Circuit events, two additional LT2-F systems were added to our fleet. The flipper arms were instrumented with customised absolute encoders, enabling the platform to be driven up and down stairs.

\begin{figure}[t]
    \centering
    \begin{subfigure}{.25\linewidth}
    \centering
    \includegraphics[height=35mm]{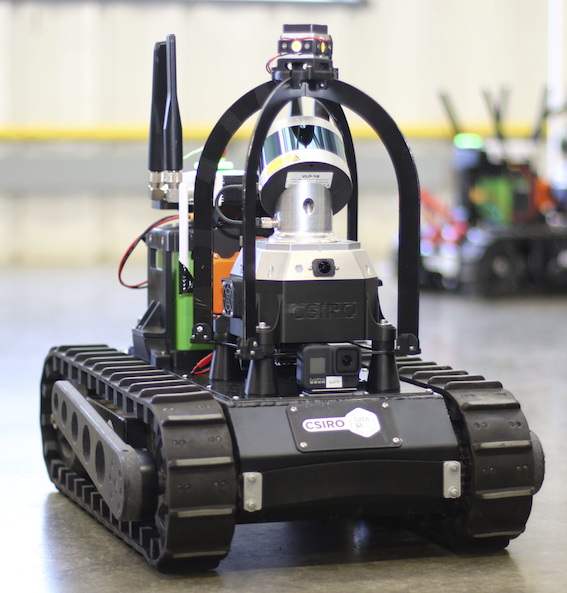}
    \caption{}
    \label{fig:SuperDroid_drop_node}
    \end{subfigure}
    \begin{subfigure}{.25\linewidth}
    \centering
    \includegraphics[height=35mm]{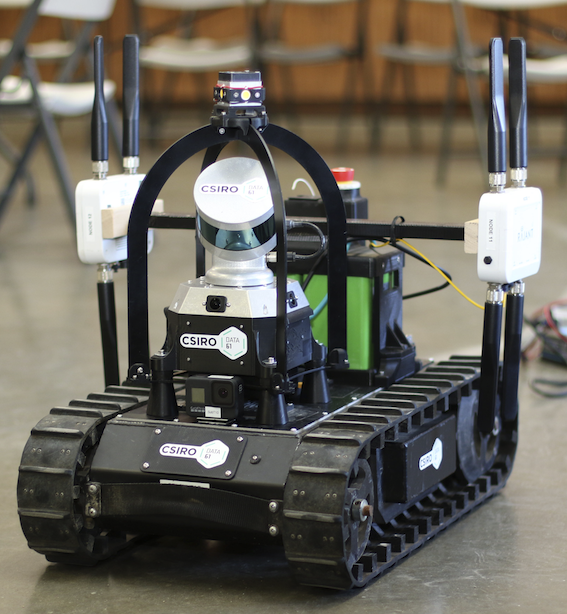}
    \caption{}
\end{subfigure}
    \begin{subfigure}{.25\linewidth}
    \centering
    \includegraphics[height=35mm]{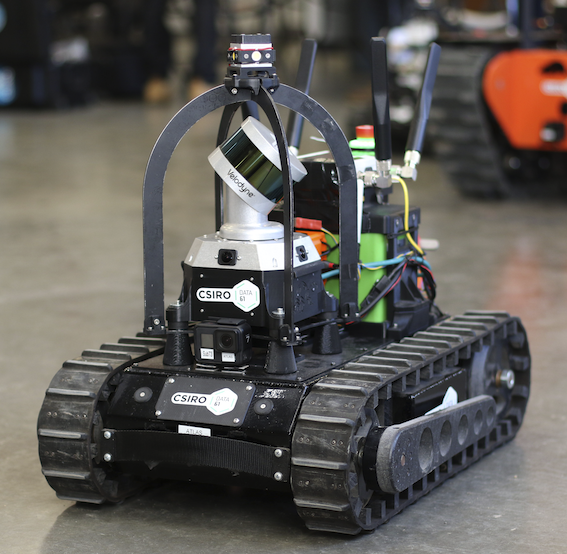}
    \caption{}
\end{subfigure}
    \caption{SuperDroid LT2-F platforms as used at the Urban Circuit event.}
    \label{fig:superdroids}
\end{figure}

The platform performed well in the Urban Circuit where the terrain was mostly concrete floor, and there was a need to traverse narrow stairwells. In rougher and granular terrain, the platform had difficulty due to its low ground clearance and regular derailing of the chain or de-tracking when small rocks got caught in drive chains or tracks. This significantly increased the maintenance requirement for this platform in such terrain. 
These limitations prompted us both to focus on procuring a larger tracked platform, the BIA5 All Terrain Robot, and to develop our own small tracked platform, the CSIRO Dynamic Tracked Robot. 

\subsubsection{BIA5 All Terrain Robot}
\label{sec:titan}

The experience at STIX identified that a larger platform with greater ground clearance and payload capacity was required. The Brisbane based Australian company BIA5 \citep{bia5} designs and manufactures the  All Terrain Robot (ATR) platform in partnership with Deakin University.\footnote{https://www.deakin.edu.au/research/research-news/articles/life-on-mars--may-help-nasa-find-out} A light weight 90\,kg version of the  BIA5 ATR was custom developed for us, enabling the robot to be easily recovered and lifted. The original version of this was initially developed for assisting first responders as a remote controlled platform, and fit through standard doors (764\,mm wide vs a ADA compliant door of 813\,mm), but weighed nearly 300\,kg. 

We received our first ATR (Version 1) in early June 2019 ahead of the Tunnel Qualification. This was then fitted with a similar electronics system used on the SuperDroid LT2-F, along with a CatPack. BIA5 developed a lighter weight version of the system (Version 2), which includes an  aluminium (instead of steel) chassis along with lighter gearbox and motors. The batteries used the lighter LiFePO$_4$ chemistry which also delivered 
additional energy. The navigation and control computer changed from the Intel NUC to a Cincoze DX-1100 ruggedised workstation following difficulties operating in hot environments.

The platform proved reliable and robust in both the Tunnel and Urban events. The exceptional ability of the platform to traverse rough terrain was thoroughly tested in the cave event, as discussed in Section \ref{ss:cave}. The size of the platform allowed us to carry and deploy our communications network at both the Tunnel and Urban event, and also permitted the UAV to be carried and deployed in situ at the Urban event. Both versions of the ATR have class-leading capability to climb stairs, a feature that we have not had the chance to utilise at competition to date due to the narrow stairwells at the Urban event. The size also precludes entering narrow doorways, thus smaller UGVs and the ATR provide complementary capabilities. The ATRs are now the most robust, capable, and tested robots in the fleet. 

\begin{figure}[t]
    \centering
    \begin{subfigure}{.32\linewidth}
    \includegraphics[height=35mm]{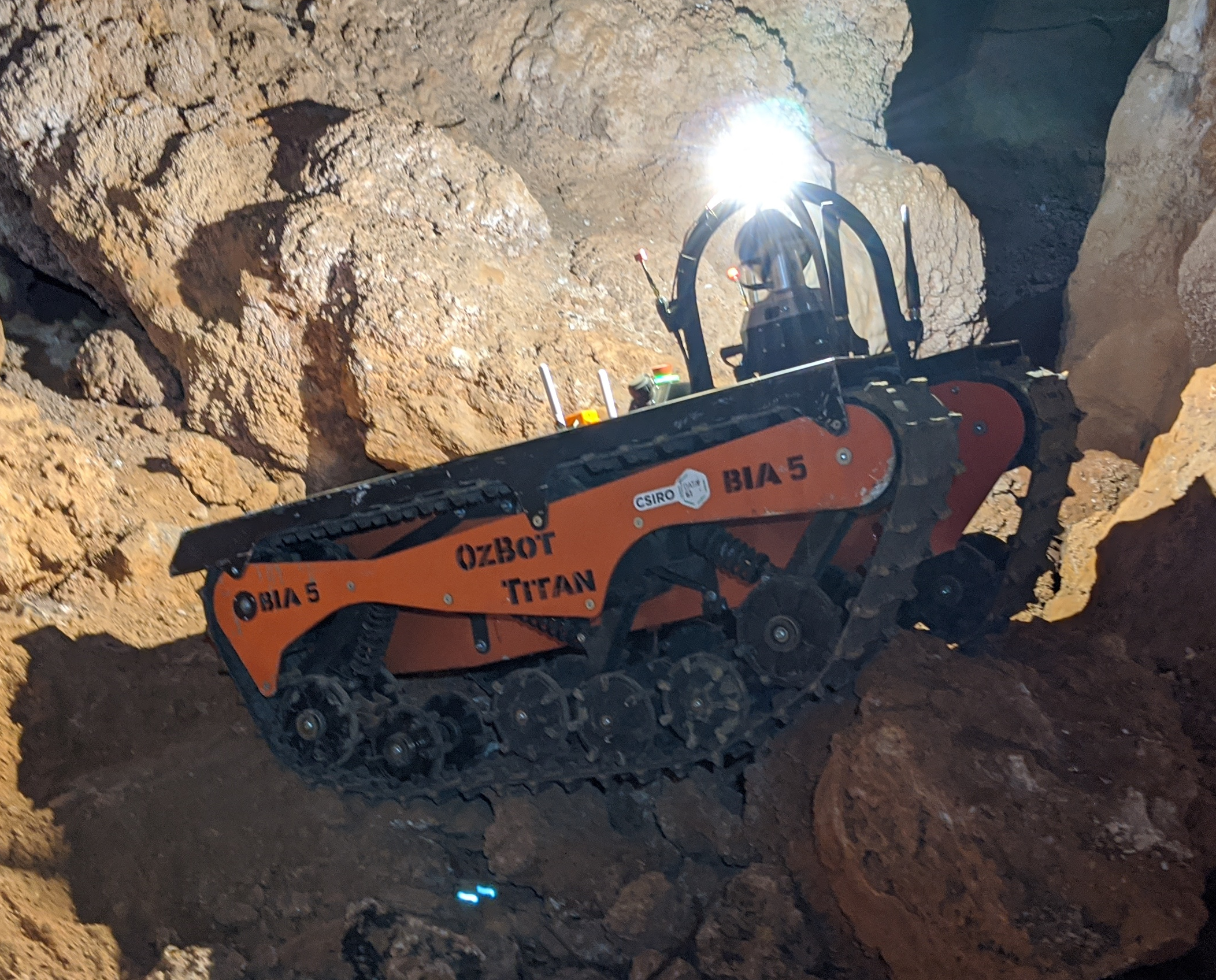}
    \caption{}
    \label{fig:titan_cave_rocks}
    \end{subfigure}
    \begin{subfigure}{.32\linewidth}
    \includegraphics[height=35mm]{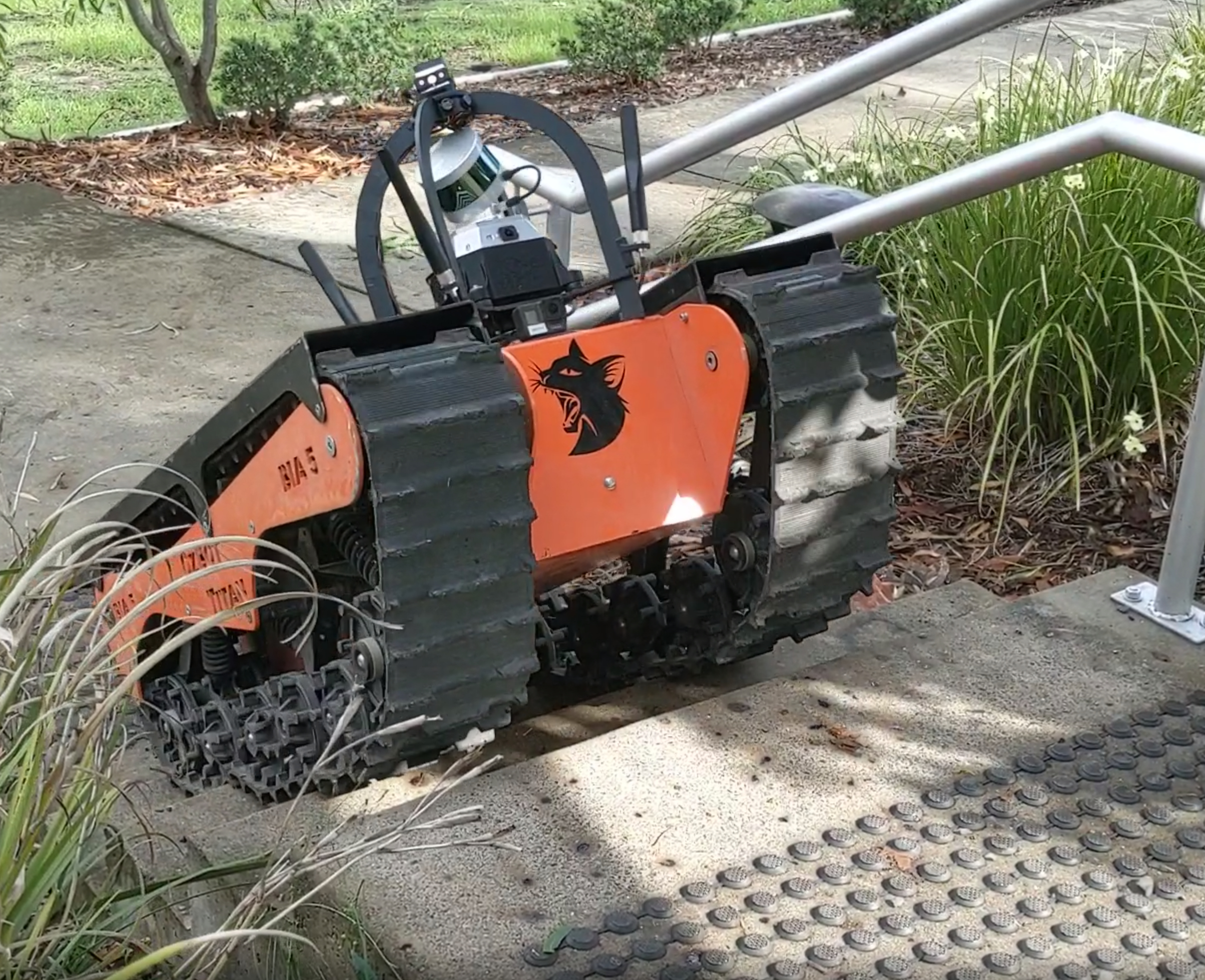}
    \caption{}
    \label{fig:titan_tunnel_node_dropped}
    \end{subfigure}
    \begin{subfigure}{.32\linewidth}
    \includegraphics[height=35mm]{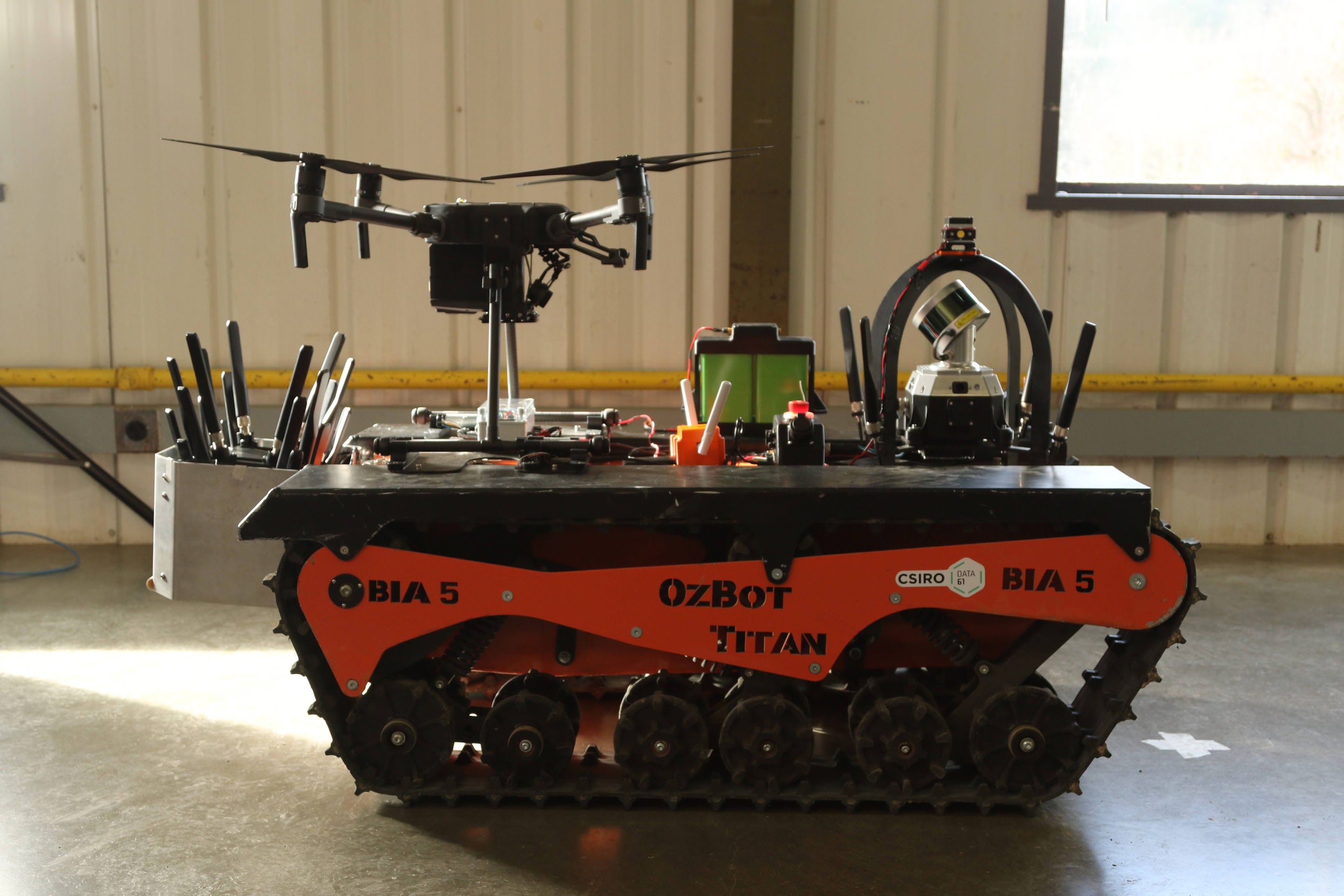}
    \caption{}
    \label{fig:titan_tunnel_drone}
    \end{subfigure}
    \caption{BIA5 ATRs (a) traversing challenging terrain during the Cave event, (b) traversing stairs, (c) carrying UAV deployment platform and communication node dropper utilised at Urban Circuit event.}
    \label{fig:ATR}
\end{figure}

\subsubsection{Ghost Robotics Vision60}
\label{sec:ghost}
The Ghost Robotics Vision 60 quadruped is a commercial-off-the-shelf (COTS) platform which is integrated into our fleet of heterogeneous robots \citep{ghostrobotics}. An early prototype (v3.2) was used at STIX and a later model (v3.7) that was rented from Ghost Robotics was deployed at the Tunnel event. As with other agents, perception, localisation and mapping is provided with a CatPack mounted on the robot. The UGV autonomy stack is implemented on `the barn', an enclosure with the same computation and electrical payload as in the SuperDroid LT2-F and version 1 of the  ATR Systems. Velocity and mode inputs are sent to the robot's API from the barn. The `Sit', `Stand' and `Walk' modes can be selected via the operator station.

While the platform's capabilities improved with new hardware and software iterations, due to reliability concerns with the electrical and power system, we decided to de-prioritise the use of this platform for the Urban and Cave events, focusing instead on the ATRs due to its reliability and traversal capabilities. However, with upgraded hardware, electrical and power systems, the v4.3 robot is showing promise with rough terrain, narrow gap and stair climbing capabilities. This version is shown with the CatPack and the barn mounted in Figure~\ref{fig:robot_fleet}. The improved low level firmware provides better `blind walking' capability on rough terrain without relying on explicit foot placement planning. The platform itself is only 0.55\,m wide, making it physically capable of walking through narrow openings. The blind stair mode uses high stepping to climb up stairs and a low body height with backward walking to go down stairs. With the success of team CoSTAR in the Urban event using quadruped robots, we are re-prioritising the use of the Vision60 quadruped platform for the final event through rigorous testing and evaluation of its locomotion capabilities.

\begin{figure}[ht]
    \centering
    \begin{subfigure}{.25\linewidth}
    \includegraphics[height=28mm]{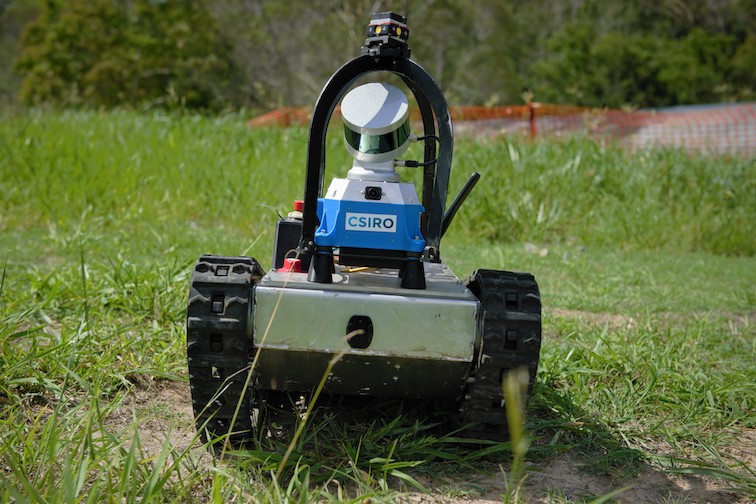}
    \caption{}
    \label{fig:pumpkin_front_profile}
    \end{subfigure}
    \begin{subfigure}{.25\linewidth}
    \includegraphics[height=28mm]{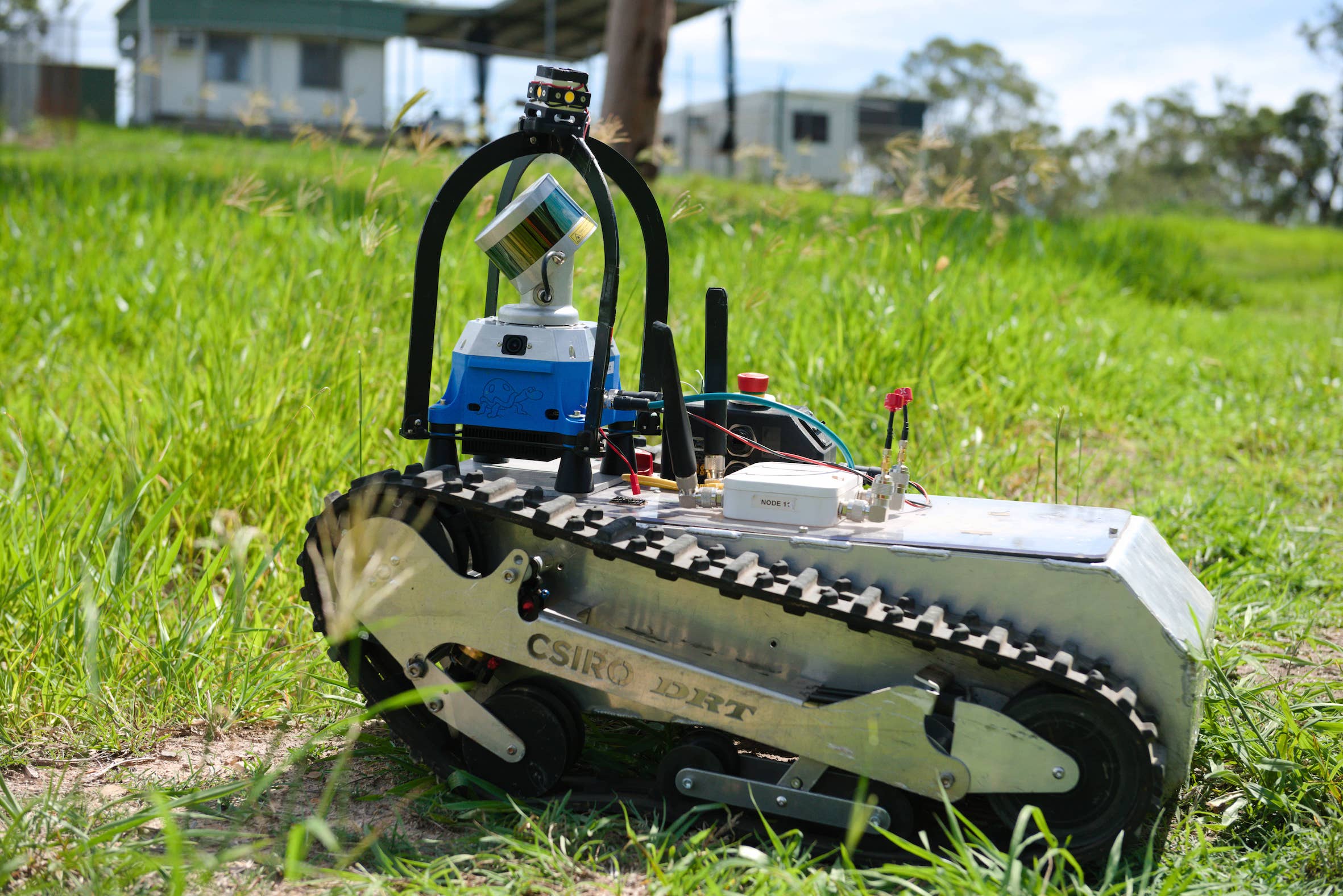}
    \caption{}
    \label{fig:pumpkin_side_profile}
    \end{subfigure}
    \begin{subfigure}{.45\linewidth}
    \includegraphics[height=25mm]{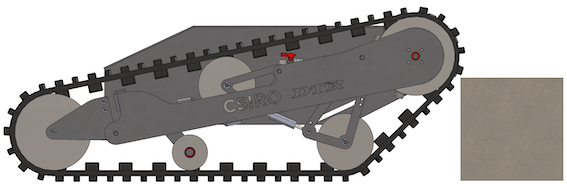}
    \caption{}
    \label{fig:pumpkin_step_height}
    \end{subfigure}
    \caption{CSIRO Dynamic Tracked Robot (DTR) shown in (a) and (b) was designed to overcome stairwells and high steps without the use of a flippers. A CAD drawing of the DTR is shown next to a  180\,mm step as reference in (c).}
    \label{fig:dtr}
\end{figure}

\subsubsection{CSIRO Hexapod}
\label{sec:bruce}
 Initial development focused on building a custom hexapod, aiming for a generalist legged robot with a large polygon of support using a splayed tripod gait. This was based on previous work done by our group developing prototype hexapod robots \citep{Elfes2017,bjelonic_weaver:_2018}. The resulting CSIRO Hexapod robot named Bruce \citep{steindl_2020} is a six legged prototype robotic platform designed and built by CSIRO to participate in the DARPA SubT Challenge. Similar to all other platforms in the fleet Bruce has the CatPack mounted for perception, localisation and mapping. It has two computers, one for the low level drivers and controllers and another similar to that of the barn on the Vision60 quadruped running the UGV autonomy stack. Its `Sit', `Stand' and `Walk' modes were commanded via the operator station. 

Bruce, shown in Figure~\ref{fig:robot_fleet} and later in Figure~\ref{fig:tunnel_circuit}, was intended to expand the capabilities of the SubT heterogeneous robotic fleet through the addition of fast statically stable locomotion over difficult rough terrain, the ability to climb over obstacles too difficult for traditional modes of locomotion and the ability to change the body configuration to fit through narrow openings. Given the short development time to bring up a brand new platform, the prototype did not achieve the ambitious goals within the expected timeline, but did demonstrate rough terrain traversal over grass, rocks and rubble at 0.3\,ms$^{-1}$, and flat-ground speeds of up to 0.5\,ms$^{-1}$. The overall weight of the platform and the performance limits of the actuators introduced a severe constraint on operational time where the actuators would reach the thermal cut-off with prolonged operation at the top speed after a few minutes. Investigations of customised heat sinking combined with additional ventilation were commenced to address this issue.

As with the Vision60 quadruped platform, development of Bruce was de-prioritised for the Urban and Cave events in favour of the ATR platforms that demonstrated exceptional rough terrain traversal capability. Due to the development effort and time required to bring this platform up to the desired reliability and capability levels, it was decided to halt work and divert the resources to address other priorities.

\subsubsection{CSIRO Dynamic Tracked Robot}
\label{sec:pumpkin}
The Dynamic Tracked Robot (DTR) was designed to bring the strengths of the ATR platform to a smaller footprint, allowing passage through narrow doorways, climbing of stairs without flippers, and easy transport for international travel as checked-in luggage, as well as improving ground clearance and robustness of the SuperDroid LT2-F. As shown in Figure~\ref{fig:dtr}, the front track geometry mimics that of the ATR, allowing the system to climb up steps, and use the same motion planning and behaviours as the larger ATRs. The tracks are also suspended akin to the ATR platform to absorb impacts from drops.

\subsection{UAV Hardware}
\label{ss:uavhardware}

The UAV system consists of a DJI M210 UAV, fitted with an Emesent Hovermap payload, custom gimballed camera system and Rajant radio. The Hovermap payload is a commercial payload\footnote{https://www.emesent.io/Hovermap/} that runs Emesent's autonomy stack with additional functionality to facilitate integration with the rest of the CSIRO Data61 Team's system, and to provide perception capabilities. The Hovermap payload and the UAVs that were deployed are shown in Figure~\ref{fig:robot_fleet} and later in Figures~\ref{fig:stix}, \ref{fig:tunnel_circuit} and \ref{fig:urban_circuit}.

The Hovermap payload utilises a similar hardware configuration to the CatPack, utilising the same Velodyne VLP-16 lidar and SLAM software. Compute is provided by an Intel NUC single board computer. The payload utilises the DJI Onboard SDK to interface with the UAV, and provides attitude and thrust setpoints.

In contrast to the UGV's CatPack, the UAV utilises a single camera system on a three-axis gimbal for detection and localisation of artefacts. Gimballed cameras are common in UAV inspection applications as the dynamic movement of the UAV easily leads to motion blur in imagery from fixed cameras. In the subterranean environment, a lack of lighting is an additional motivator as this requires longer exposure times which exacerbates motion blur. 

For the Tunnel and STIX event, we used a DJI Zenmuse 4S gimbal system with an illumination system fixed to the aircraft frame. This constrained the gimbal range of movement to the forward 90\degree of the aircraft as illumination was only provided in this direction. This constraint prevented behaviours such as looking down side passages during aircraft navigation. During the Urban and Cave events, this gimbal was replaced with a custom implementation that mounted a spotlight coaxially with the camera system allowing light in the direction of pointing, freeing the camera pose from the UAV body frame. This second system utilised FLIR BlackflyS cameras and a 16\,mm lens, which limits FOV but gave an effective artefact detection range of around 10\,m.

\subsection{Communication Systems}
\label{ss:commsys}

The concept and implementation for the communication system changed drastically during development, based of the results of the circuit events, and by taking inspiration from other teams.

\begin{figure}[ht]
    \centering
    \begin{subfigure}{.21\linewidth}
    \centering
    \includegraphics[height=45mm]{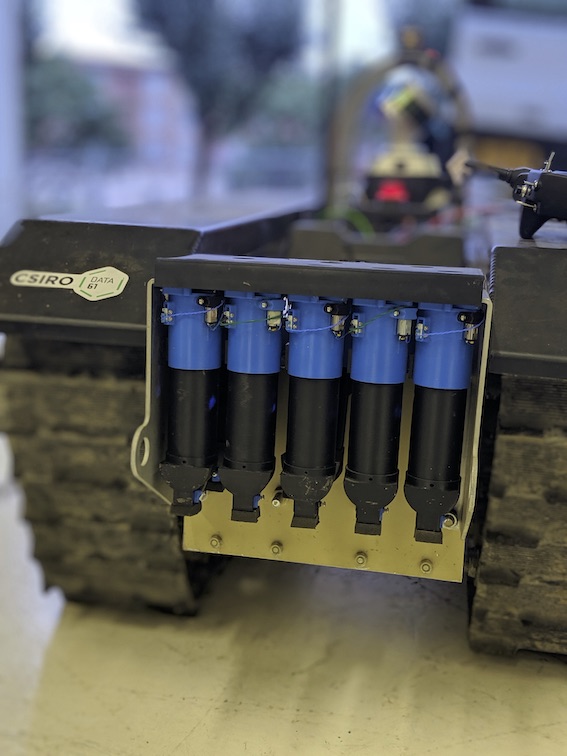}
    \caption{}
    \label{fig:TunnelDropNode_1}
    \end{subfigure}
    \begin{subfigure}{.21\linewidth}
    \centering
    \includegraphics[height=45mm]{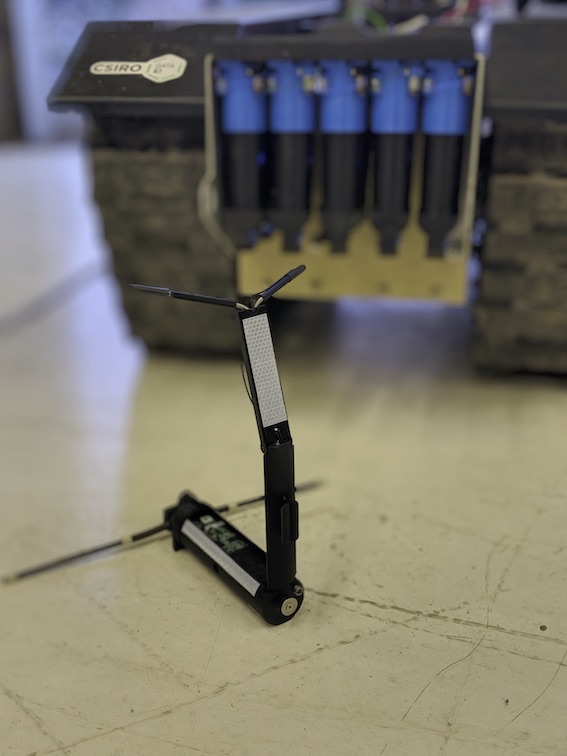}
    \caption{}
    \label{fig:TunnelDropNode_2}
    \end{subfigure}
    \begin{subfigure}{.21\linewidth}
    \centering
    \includegraphics[height=45mm]{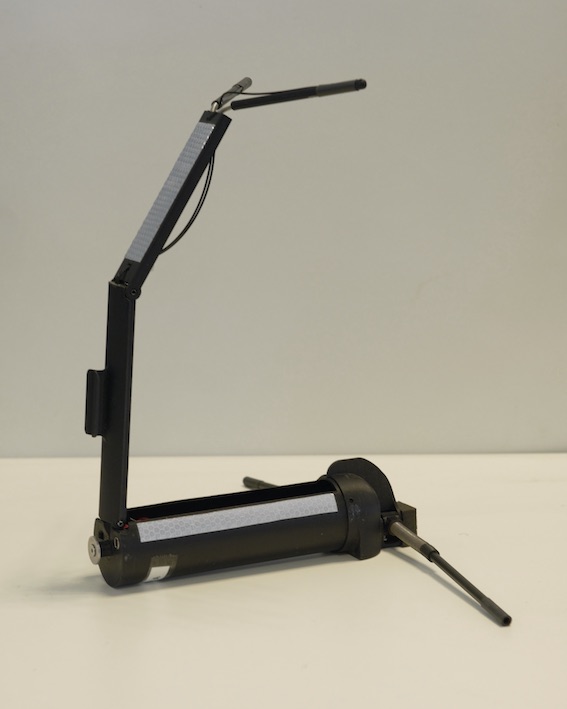}
    \caption{}
    \label{fig:TunnelDropNode_3}
    \end{subfigure}
    \caption{Ubiquiti drop nodes, as they were carried in the node dropper at the back of the ATR (a), a self erected node after being dropped (b) and a another view of the deployed drop node that consists of the Ubiquiti radio, a LiPo battery, a servo motor and the circuitry to perform delayed self righting and erecting the antenna after being dropped (c).}
    \label{fig:ubiquiti_drop_nodes}
\end{figure}

The original concept of operations had robots deploying a chain of small lightweight drop nodes (Figure~\ref{fig:ubiquiti_drop_nodes}), creating a communication backbone in a \emph{tree topology}, with most information flowing from each agent back to the base. The primary focus was on creating a long multi-hop communication backbone, following the path of each robot, with robots staying in contact with this backbone to relay information, trickling data at a steady low bandwidth. An initial implementation of this network used Ubiquiti UniFi Mesh Access Points (UAP-AC-M), communicating over dual-band WiFi (2.4GHz and 5GHz) and repackaged into self erecting, self powered drop nodes. While the drop mechanism, deployment and self erection worked as designed, this initial communication system had many limitations, as it used all in-band communications, with bandwidth halving at each hop. In addition, meshing was unreliable, even when a fixed topology was configured, and caused significant issues in the STIX and Tunnel event.

In parallel to using the Ubiquiti-based network, a custom communication node was developed, using three ESP-32 modules at each node. Each ESP-32 module on the device operated in a different band, with alternating frequencies connecting to each neighbour along the chain, reducing bandwidth loss from in-band interference. A functioning demonstration of this system was created, but ESP-32 modules transmitting overloaded the other ESP-32 modules receiving on the same node, causing degradation.             

Instead of continuing to develop a pervasive but fixed communication network, we instead switched to increasing the autonomy of each agent, allowing them to operate beyond communication range and return to report data. Inspired by team PLUTO at the Tunnel event, who used a mesh network of Rajant radios, accumulating data on each agent and ferrying the data back through some combination of connectivity and deliberate return of agents to the base \citep{Miller20}, we adopted Rajant radios. Specifically, we utilised a combination of dual-band (2.4GHz and 5GHz) ES-1 nodes for ground robots and drop nodes, and single-band (5GHz) DX-2 nodes for the UAVs. Unlike PLUTO, we continued to drop nodes, aiming to keep persistent communications back to the operator wherever possible.    

This decision to switch communication paradigms was also driven by the desire for robots to share map data with other agents (Section \ref{sec:multiagent}). Agents now store all data onboard in a database called \emph{Mule} (Section \ref{ss:mule}), sharing it with nearby agents and the base station when connected. If an agent, or group of agents, is disconnected from the base station and they accumulate significant information, they will return to communication range either independently or coordinating through task allocation (Section \ref{ss:taskallocation}). 

\begin{figure}[ht]
    \centering
    \begin{subfigure}{.27\linewidth}
    \centering
    \includegraphics[height=33mm]{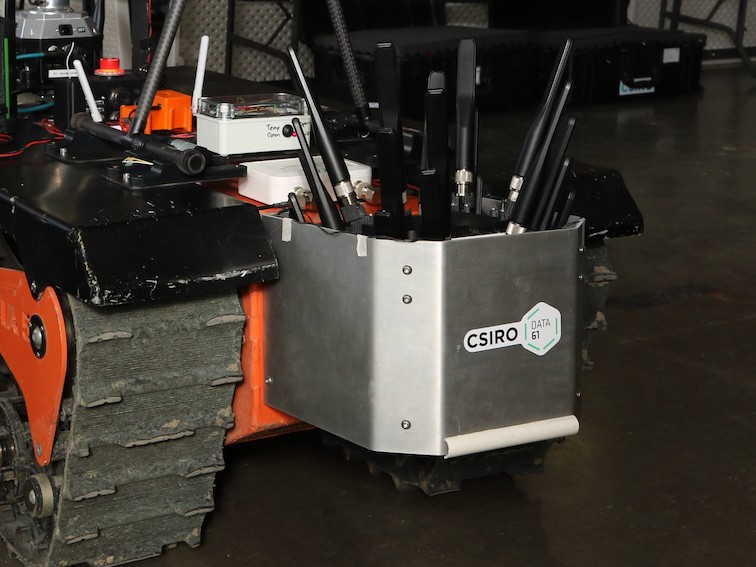}
    \caption{}
\end{subfigure}
    \begin{subfigure}{.15\linewidth}
    \centering
    \includegraphics[height=33mm]{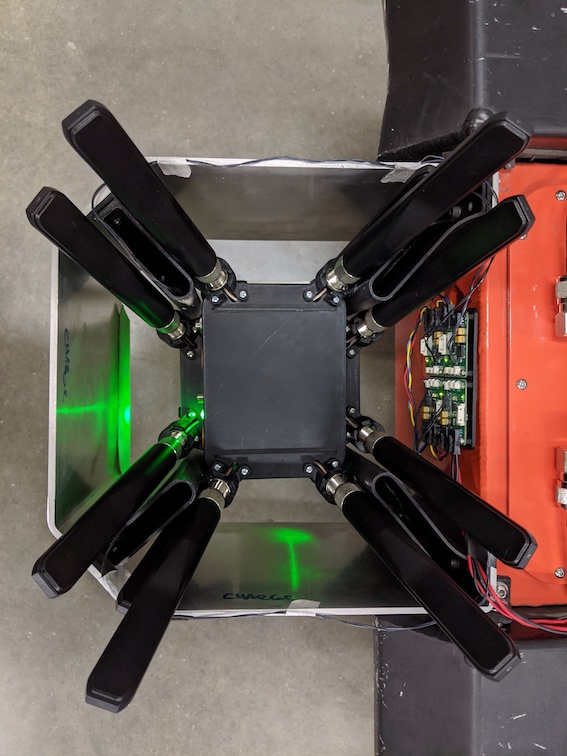}
    \caption{}
\end{subfigure}
    \begin{subfigure}{.3\linewidth}
    \centering
    \includegraphics[height=33mm]{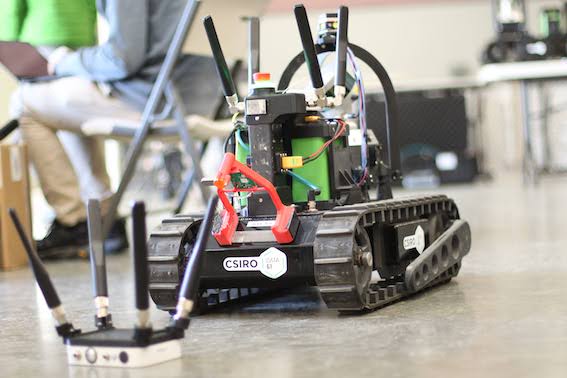}
    \caption{}
\end{subfigure}
    \begin{subfigure}{.18\linewidth}
    \centering
    \includegraphics[height=33mm]{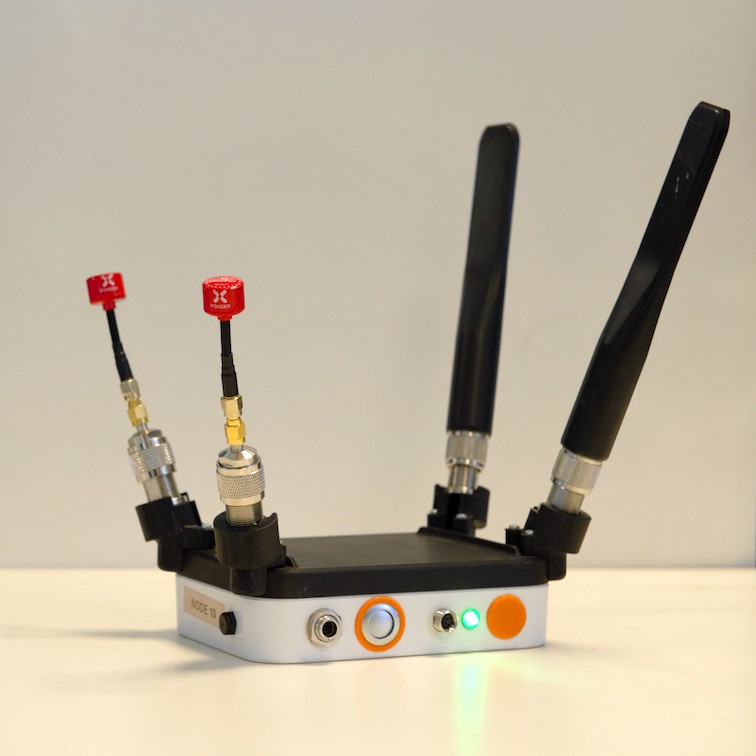}
    \caption{}
\end{subfigure}

    \caption{Drop Nodes from the Urban Circuit: Rajant ES-1s were modified to contain a battery and hold the antenna upright. (a) and (b) show a five node dropper `stack' attached to the back of the ATR. (c) shows a single node deployed from a SuperDroid robot. (d) shows a single Rajant node.}
    \label{fig:rajant_droppers}
\end{figure}

\begin{figure}[ht]
    \centering
    \begin{subfigure}{.3\linewidth}
    \centering
    \includegraphics[height=33mm]{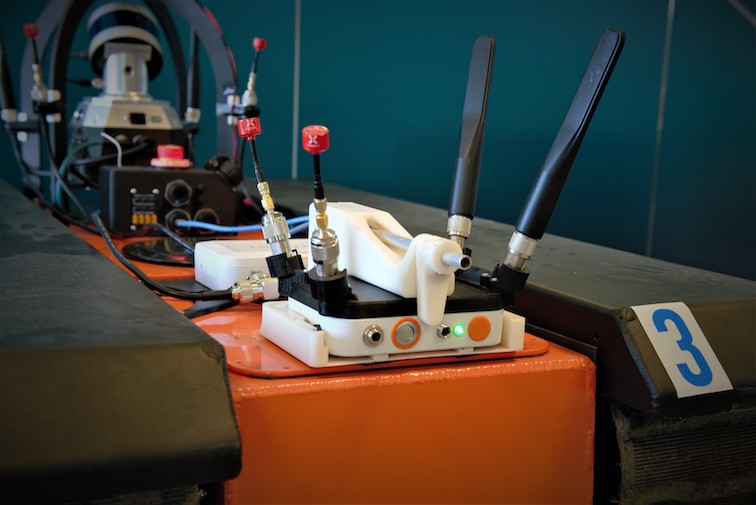}
    \caption{}
    \label{fig:shooter}
    \end{subfigure}
    \begin{subfigure}{.2\linewidth}
    \centering
    \includegraphics[height=33mm]{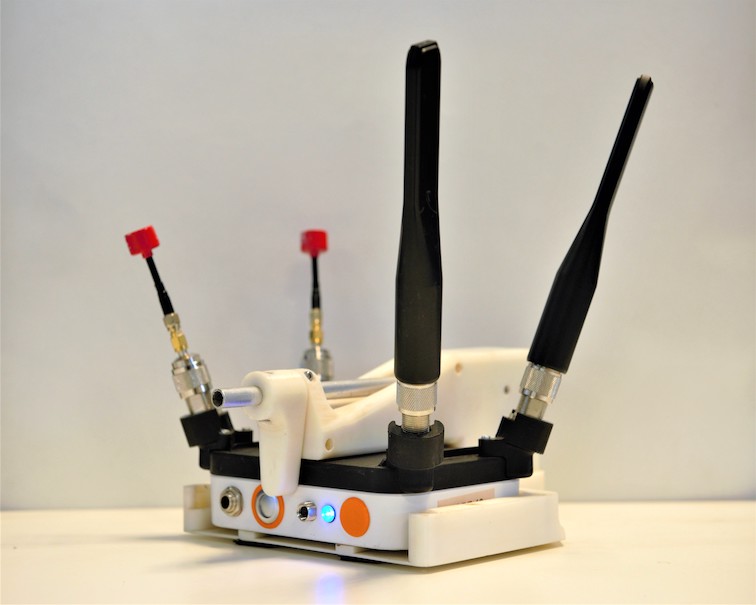}
    \caption{}
    \label{fig:shooter_profile}
    \end{subfigure}
    \caption{Node droppers from the cave circuit, with (a) showing the new dropper attached to the ATR.}
    \label{fig:shooters}
\end{figure}

At the Urban event, the ATRs each had a dropper containing five nodes, while the SuperDroids had a single node dropper (Figure~\ref{fig:rajant_droppers}). After analysing the performance at Urban circuit, it was found that only a few nodes were actually dropped, and most data traversed though the nodes on the robots themselves, or when they returned to base. Thus, for the cave event each UGV had a single drop node, and the ES-1 hardwired on the robot itself. A new deployment mechanism was also designed so that the node dropper no longer hung off the back of the robot, which decreased the overall footprint of the robot (Figure 
\ref{fig:shooters}). This modification allowed greater manoeuvrability in tight spaces, and reduced the likelihood of the dropper colliding with obstacles during autonomous operation. 

\section{Single Agent Capabilities}
\label{sec:singleagent}

To score points in SubT, artefacts must be detected, correctly classified, and localised to within five metres of their surveyed location, where the origin is defined at the starting gate to the course. This in turn requires agents to be able to autonomously explore a large, complex environment, traverse challenging terrain, construct an accurate map for both navigation and accurate localisation of artefacts, and communicate results to other agents and the base station. Each robot in our fleet is equipped with the capability to perform each of these tasks, exploring, mapping, and locating artefacts. In this section we introduce the systems that are present on each platform that allow us to perform these feats. 

\subsection{Perception}
\label{ss:perception}

The robots were required to autonomously detected five artefact classes at each circuit event. Three artefact classes, survivors (a thermal mannequin in high visibility clothing), cell phones (emitting Bluetooth and WiFi signals), and backpacks were common at each event. Additional course specific items included fire extinguishers, drills, air-vents, high \coo\ gas concentrations, rope, and helmets. Detecting \coo\ required the use of a specific gas sensor, whereas all other artefacts could be detected visually. Both the UGV and UAV platform primarily detected artefacts using onboard RGB cameras, using lidar data projected into the camera frame to localise the artefact with respect to the a local map.  The UGV platforms had additional Bluetooth and WiFi signal signal analysers to detect cell phones, and gas sensors to detect high \coo\ concentrations. 

All artefact detections are relayed back to the operator, and appear on a specialised GUI for the operator to review, confirm and report to the DARPA scoring server. Artefacts are provided to the operator within a sortable list, alongside the artefact location in the global map, plus its associated image. 

\subsubsection{Camera Based Detection}
An annotated RGB dataset was collected of the DARPA artefacts using a combination of images acquired from cellphone cameras, hand-held CatPacks, and images from the robots during runs. This dataset has been continuously extended and now contains more than 13,000 images in multiple environments including urban construction sites, natural caves, and underground areas, with artefacts appearing in poorly illuminated environments with a wide variety of poses, with and without occlusion. The dataset contains labelled instances of fire extinguishers (262), cell phones (477), survivors (793), vents (1242), helmets (2484), ropes (2413), backpacks (987), and drills (1250). Images were captured in 9 different environments, averaging 1480 labelled artefacts per environment. Images containing multiple artefacts are also present. Images of the environment containing none of the artefacts (i.e. negative examples) were also gathered (3412 images) in order to increase the robustness of the model in each environment.

The UAV and UGV platforms use detectors trained on this data set, augmented with images from the UPenn Dataset by \cite{shivakumar2019pst900}, MSCOCO by \cite{MSCOCO2014} and ImageNet by \cite{ImageNet2009}.

The total number of annotated images of each artefact was guided by the performance of the artefact detector during practice competition events. In cases where the detector performed badly (e.g., the artefact was in the field of view of agent for 150 frames but was detected only five times), a subset of these frames would be annotated and a new detector model would be trained. This manual inspection process was repeated until the artefacts were reliably detected during full system testing. This approach was adopted as it gives the best indication of actual detector performance in a new environment.

The only factor we consider when deploying the artefact detector on the agent is the number of artefact detections received by the operator. The operator has a finite amount of time to process detections and it is critical that they are not overwhelmed. The desired rate is achieved by removing detections with a posterior probability below an artefact specific threshold. The thresholds are calculated a priori using datasets acquired in environments which were anticipated to be similar to competition events. The datasets were approximately 45-60 minutes in duration and were captured by an agent operating as it would during a competition event.

All ground vehicles use the NVIDIA Jetson Xavier GPU in the CatPack for image processing, batching together the images from the four cameras before inference. DeNet \citep{Tychsen_Smith_2017,Tychsen_Smith_2018} is used for artefact detection, ported to the NVIDIA TensorRT framework \citep{TensorRT_2020}, and processes images at 12\,Hz (3\,Hz per single camera) using a single CPU core in conjunction with the GPU. 

In contrast to the UGV, the UAV utilises a single camera system on a three-axis gimbal for detection of artefacts. Artefact detection was performed using Mobilenet V2 SSD 300 by \cite{Sandler18} from the Tensorflow Object Detection Zoo by \cite{Huang16}. This artefact detector was optimised to run onboard the Intel CPU at 3Hz using Intel's OpenVino Toolkit. The model was trained with additional data containing images with false positives generated from parts of the aircraft in the frame. 

\subsubsubsection{\textbf{UGV Artefact Localisation and Tracking}}

Prior to the Urban Circuit, all detected artefacts were localised in the map (using the lidar data) and then reported directly to the operator. We experienced an unmanageable number of (false) detections when operating in novel environments, and the vast majority of these were re-detections of a previously detected artefact. For the Urban circuit, we introduced an artefact location tracker which ran locally on each UGV agent.

Localisation is performed by selecting 3D lidar points which project inside the 2D bounding box of the detected artefact (the intrinsic and extrinsic parameters for each camera are calculated a priori). The 3D location of the artefact is determined as a weighted average of the selected points where the weights favour points closest to the camera. Artefact detections greater than 20 metres away from the robot are ignored due to the size of the artefact in the image\footnote{A plane with dimensions $0.5$m$\times 0.5$m positioned parallel to the image plane at a distance of $20$m will occupy $33 \times 33$ pixels in the image.} and due to the limited range of the lights on board the agent, recalling that the environments are poorly illuminated in general.

Localised artefacts are tracked by computing the posterior of a new artefact at the incoming localised artefact location. A Gaussian distribution with isotropic covariance was used to model the position of known artefacts and a uniform distribution was used to model the probability of a new artefact. A uniform distribution was used to model the prior probabilities of each tracker hypothesis. The volume of the uniform distribution was selected such that a new track was created if the artefact position was more than 1m away from a known artefact. Tracks are dropped when an artefact has not been re-detected for more than thirty seconds. Artefact detections are linked to the trajectory of the agent making the detection, which can then later be corrected using the multi-agent SLAM solution. This pipeline provided artefact localisation accuracy at the Urban event, where the nine artefacts reported from autonomous UGV camera detections had a 0.46\,m  RMS error compared to the DARPA ground truth provided post-event, and included the most accurately reported artefact from any team with a 0.22\,m error.

\subsubsubsection{\textbf{UAV Artefact localisation and Gimbal Pointing Autonomy}}

One of the motivations for using a gimbal camera system on the UAV was to decouple perception from the flight autonomy, such that a single stabilised camera could cover multiple view points. This had advantages as tight spaces that were un-flyable were still potential artefact locations, and could look down side passages that were not at flight autonomy goal locations.

The gimbal pointing autonomy analyses the space around the aircraft for the expected utility for seeing something new. The expected utility of visible space around the aircraft is conditioned on the camera sensor model and prior utility of that space. The point target maximising the expected utility evaluated from the gimbal's current position is tracked by the gimbal using a 3D pose controller. This autonomy enables the investigation of side passages as the aircraft passes them as well as spaces that are initially occluded, but become visible as the aircraft moves past the intervening obstacle. The expected utility is updated each exposure and maintained globally over the course of the flight. 

The UAV maintains a polar range map of occupied voxels around the UAV based on the perception map used for navigation. Bounding boxes from the above artefact detection pipeline are transformed into a spherical range map, representing the nearest artefacts around the aircraft. The ranges of the pixels contained in the spherical bounding box are binned on a 1m resolution and the range to the artefact is computed as the average of points in the nearest occupied bin. This range is then projected using the centroid of the bounding box in the original detection using camera intrinsics and extrinsics to provide a location in the odometry frame. Artefact detections are not tracked on the UAV, and were less accurate than the UGV system at the Urban event with a 2.34\,m RMS error for reported artefacts.

\subsubsection{\textbf{Additional UGV Sensing}}

The UGV platforms included a Sensirion SCD30 sensor module for detecting \coo\ gas. This sensor was affected by changes in atmospheric pressure, altitude and temperature and thus simple thresholding methods to detect the presence of \coo\ gas gave intermittent false negatives. To avoid constant re-calibration of the sensor based on these environmental factors, the presence of the gas artefact was instead detected by sensing spikes in the levels of \coo\ within a room. 

UGV platforms also included a UD100 Bluetooth adapter and Alfa Long Range 802.11n AWUS036NHA WiFi adapter for detecting the cell phone artefact. The RSSI from both the Bluetooth discoverable and WiFi hotspot packets was used to determine the proximity to the artefact. The position of the agent at the time of reception was reported to the operator, who then inferred the likely artefact position based on the history of reports, and  contextual information. \coo\ reports were handled similarly.

These additional sensing modalities provided five out of 15 artefact reports scored at the Urban event, including three cell phone detections missed by the RGB cameras.

\subsection{Odometry}
\label{ss:odometry}
Localisation and mapping is based on CSIRO's Wildcat SLAM solution. Here we describe how odometry is generated from lidar and IMU data, while in Section \ref{sss:slamframes} we provide an outline of the multi-agent mapping process.

The odometry solution is based on \cite{BosZlo12}. The trajectory is optimised in a five second sliding window, parameterised as a series of trajectory pose samples at times spaced by 0.1~sec. Iterative optimisation steps produce corrections that are applied to the corresponding pose to provide improved trajectory estimates.

The optimisation seeks the trajectory which minimises errors, incorporating disagreements with inertial measurements, and lidar matching errors. Optimisation variables including both the trajectory and the correspondence of lidar surface elements (surfels). Like iterative closest point (ICP), we alternate between determining correspondence of lidar surfels, and optimising the trajectory conditioned on these correspondences. Trajectory optimisation steps are solved through iterative re-weighted least squares, where Cauchy weights provide robustness to outliers.

Lidar points are extracted and voxelised in time intervals of 0.5\,s and cubes of 40\,cm side, representing the surfel for each voxel as the mean and covariance of points in each voxel, both of which depend on the trajectory at the corresponding time. Multi-resolution surfels are formed by nesting hierarchically up to a resolution of 3.2\,m, and an additional offset grid is formed at the finest level offset by half a voxel in each dimension. 

The error term of the optimisation for the correspondence of surfels measures uses the difference between the surfel means, projected onto the eigenvector corresponding the smallest eigenvalue of the sum of the covariance matrices. This eigenvector represents an estimate of the normal vector for the combination of the two surfels, effectively providing plane-to-plane correspondence error. Error terms are included for corresponding surfels, where correspondence is determined using a tree-based $k$-nn lookup considering position, normal vector and time.

As lidar points leave the five-second sliding window, they are added to consolidated multi-resolution surfels for each six-second frame (generated every five seconds, containing points from the interval 5-11\,s ago). As discussed further in Section \ref{sss:slamframes}, these frames form stable local coordinate systems which can be used to locate artefacts, and reference exploration and navigation data over time and between agents.

\subsection{Exploration}
\label{ss:exploration}
Efficient exploration is achieved by maintaining awareness of the area that has been mapped so far, and pushing towards the boundary. Frontier methods were first proposed in \cite{Yam97}, and have been widely used representing the information state as a global 2D occupancy map. Complications in the SubT context include the need to incorporate SLAM loop closures (rendering a single global map representation ineffective), and the fundamental 3D characteristics of the regions to be explored (rendering a 2D occupancy map insufficient).

Our approach is detailed in \cite{williams_2020}, and is described here for completeness. The method exploits the efficient point cloud visibility approach of \cite{KatTal15} to maintain an awareness of the boundary between known and unknown space. The point cloud is supplemented by points on a sphere at a nominal maximum range, so that maximum range points that are determined to be visible represent frontiers. As well as determining point visibility, the quickhull operation in \cite{KatTal15} also produces a triangulated mesh of visible points. Long edges in this mesh represent discontinuities which could be openings, another form of frontier. The full watertight mesh of visible points is referred to as a viewpoint. Segmented frontiers are stored, representing the boundary between free space and unknown space.

\begin{figure}
    \centering
    \includegraphics[width=\columnwidth]{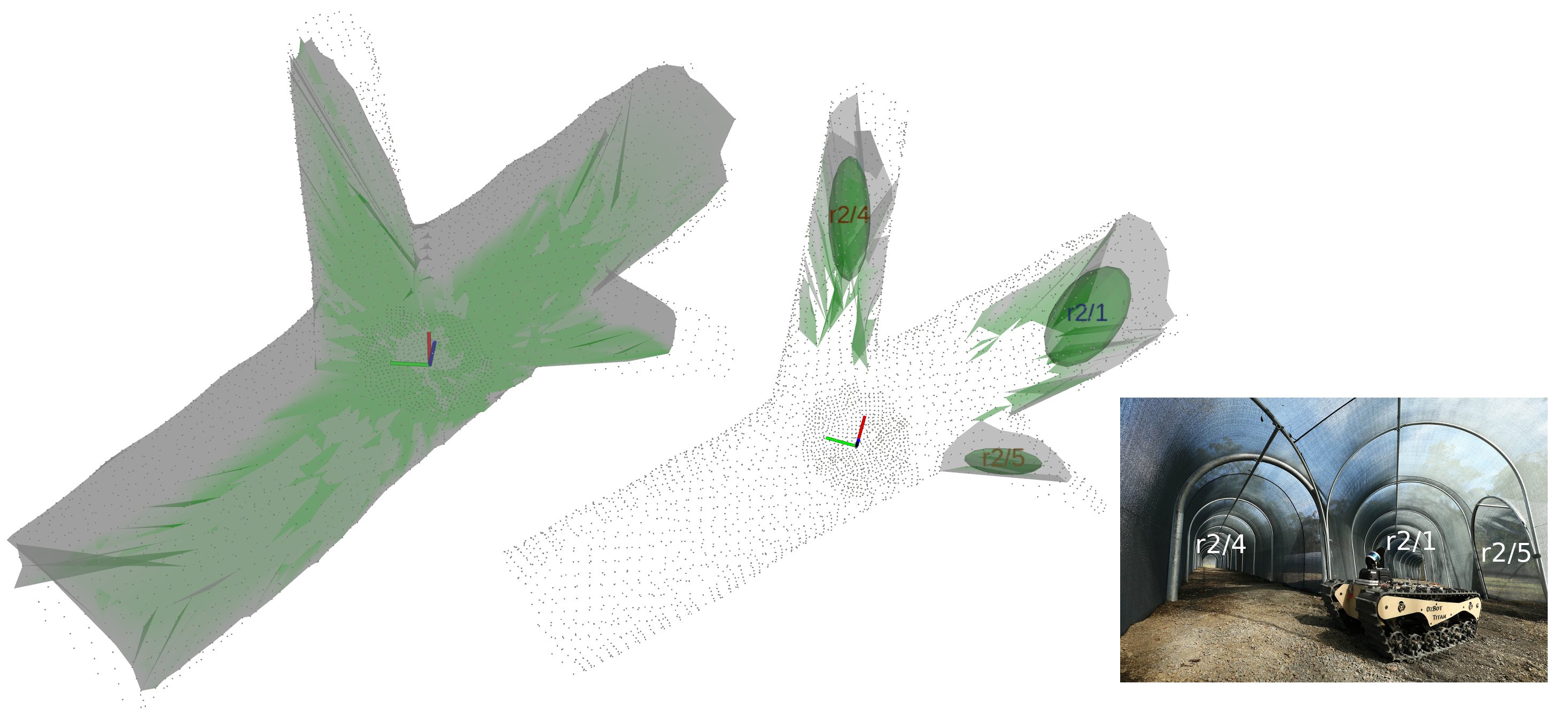}
    \caption{
        Example of frontier processing in CSIRO test tunnel. Left: watertight mesh of visible space from observer location. Middle: frontiers identified from agent position, having entered from left. Right: corresponding image with frontiers labelled.
    }
    \label{fig:frontier}
\end{figure}

Frontier processing is repeated after a metre of agent motion. New frontiers are tested for visibility against old viewpoints, and old frontiers are tested against the new viewpoint. Any vertices that are visible (i.e., inside the watertight mesh) in other viewpoints are no longer frontiers. Frontiers and viewpoints are stored in the coordinate system of the most recent Wildcat frame, so that they can be compared to data received later when the agent departs and later returns to an area. An example of the frontier processing is shown in Figure~\ref{fig:frontier}. 

A nascent explore capability was deployed in the Tunnel Circuit, where segmented frontier centroids were used as goals. In the Urban Circuit multi-agent, autonomous explore became the primary mode of operation, and traversability was estimated based on the mesh generated in producing frontiers, allowing candidate observer locations to be identified, and selected by estimating the volume behind the frontier that would be observed from different candidate observer locations. For cave testing, the additional effort that had been invested into traversability analysis in the local and global navigation stacks was leveraged to propose candidate observer locations.

Frontiers (but not viewpoints) are shared between agents, and rewards are discounted according to the proximity that agents have been to the proposed observer locations. This provides a low-communications coordination method, although it does result in somewhat reduced performance compared to that achieved communicating complete viewpoint data. Selection of frontiers utilises the task allocation framework described in Section \ref{ss:taskallocation}. 

Enabled by the homogeneous sensing and mapping capabilities, the same frontier solution is utilised on all UGV and UAV vehicles. Differences in frontier selection are described in \cite{williams_2020}.

\subsection{UGV Local Navigation}
\label{ss:ugvnavigation}
\begin{figure}[ht]
    \centering
\includegraphics[width=150mm]{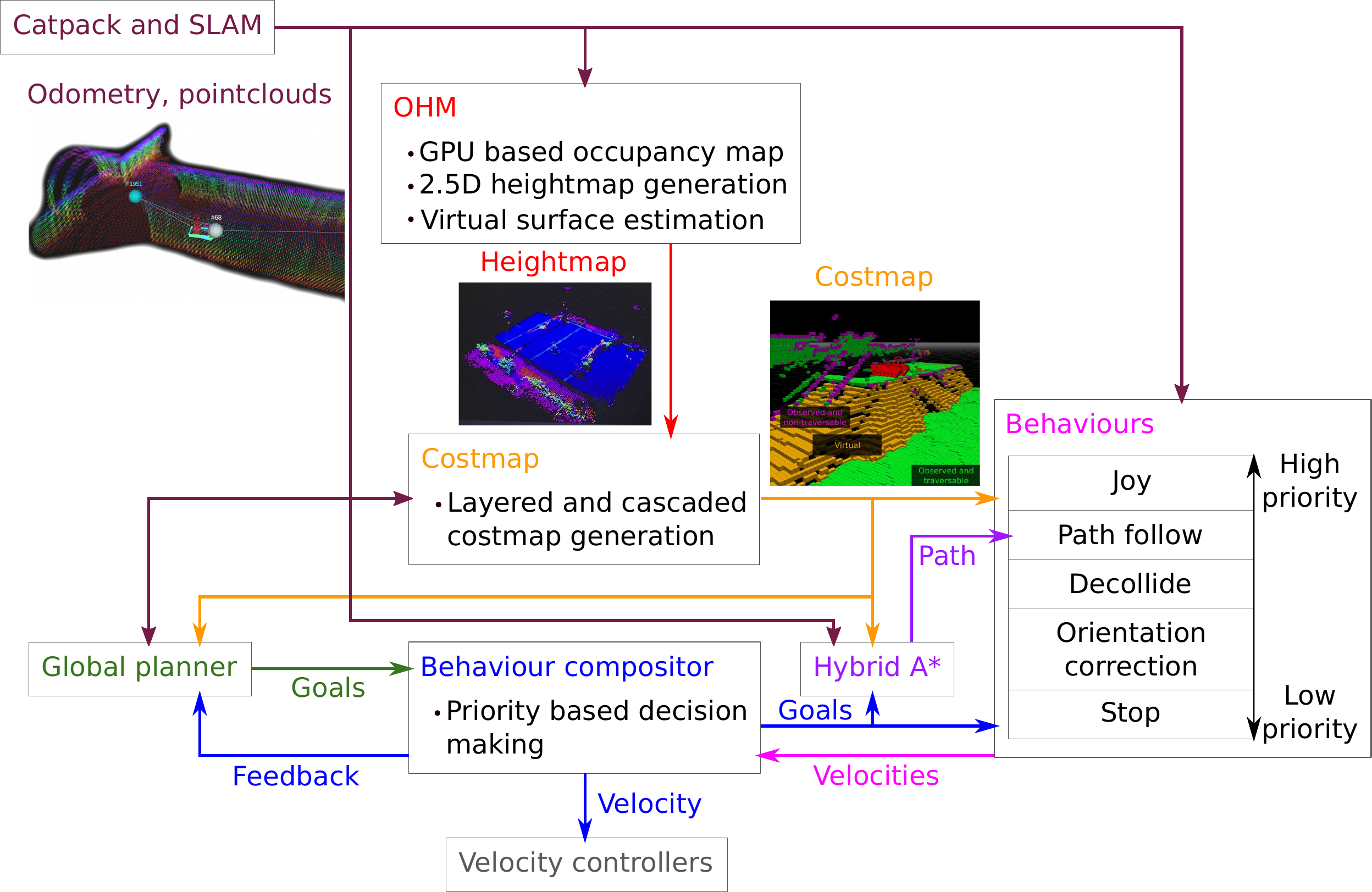}
    \caption{
        UGV local navigation components and shared information.
}
    \label{fig:local-navi}
\end{figure}

In this section, we present the local navigation approach used on UGVs as in \cite{hines_virtual_2021}. Vehicle odometry and robot relative point clouds from the front-end SLAM solution are used to plan and execute movements in order to reach waypoints provided by the global planner. This involves trajectory planning and the generation of velocity commands for the lower level speed controller. An overview of the local navigation system is shown in \Cref{fig:local-navi} and key components are summarised below.

The major role of local navigation is enabling UGVs to traverse challenging, unstructured subterranean terrain (e.g., mud, rubble, puddle, or concrete) through a cohesive arrangement of surrounding mapping, deliberative planning and reactive behaviour modules. To achieve this goal, all submodules are required to be tightly connected and to have access to high-level situation awareness including terrain slope, visibility and vehicle orientation enabling robots to recognise, plan and react around unobserved areas. In turn, it is expected for UGVs to efficiently recognise and handle negative obstacles (e.g., cliffs, ditches, holes), slopes, steps, overhangs and narrow passageways while minimising collisions.

\begin{figure}[ht]
    \centering
    \includegraphics[width=\columnwidth]{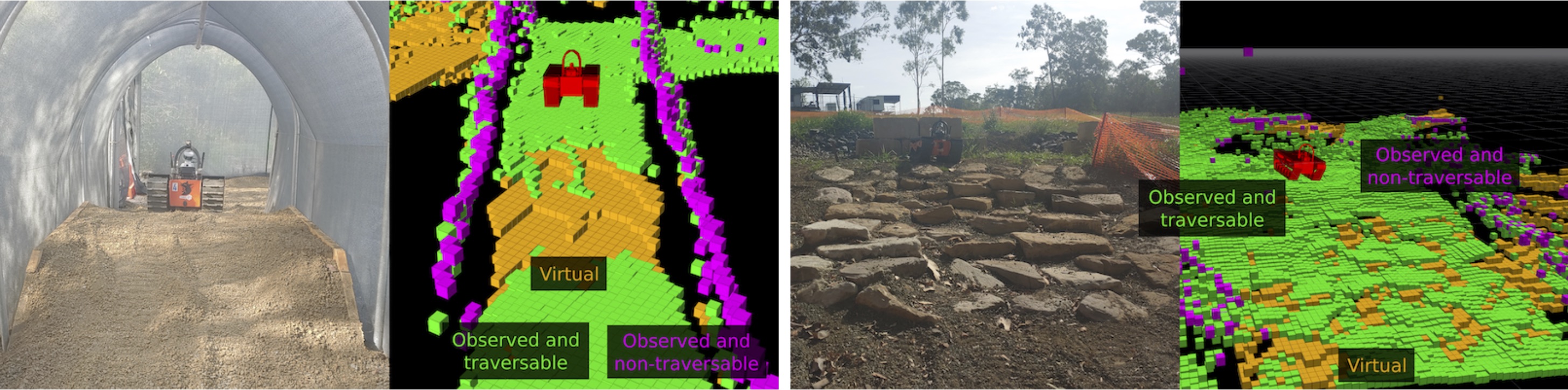}
    \caption{
        Examples of 2.5D cost maps generated for a bump in a tunnel on the left and a slope unevenly embedded with stone steps on the right. Green cells have been directly observed and are possibly traversable; pink cells have been directly observed and are definitely not traversable; and orange cells have not been observed but are best case height estimates called virtual surfaces.
    }
    \label{fig:vs}
\end{figure}

Occupancy Homogeneous Map (OHM, \cite{ohm}) is a component that is critical for safe and secure local navigation. It provides a high-fidelity heightmap which allows subsequent modules to estimate virtual surfaces and detect obstacles. OHM maintains a 3D occupancy map by integrating the SLAM-corrected point clouds produced by Wildcat's odometry processing. OHM also generates a height map that specifies a ground height per cell of a 2D grid around the robot. This height map includes virtual surfaces which are a best case height estimate for surfaces that are obscured. \Cref{fig:local-navi} shows OHM's place in the system relative to adjacent components and \Cref{fig:vs} contains examples of cost maps generated from OHM height maps with the virtual surfaces labelled.

\Cref{fig:vs} also includes labels for fatal (definitely non-traversable) cells. These are found using a method that considers the change in height between nearby cells in a way that is resilient to noisy or rough terrain and small holes in the ground. These cells are a subset of the cells that the robot should never plan to intersect with its footprint.

Short range trajectories are planned using hybrid A*. The cost function that hybrid A* uses considers the heightmap and costmap cells that intersect with the robot's footprint. Configurations that cause intersections with fatal cells in the costmap are costed fatally. The heightmap cells are used to estimate the pitch and roll components of the robot's orientation. The orientation estimate is used to increase the cost of difficult terrain and fatally cost terrain that is too steep for the robot to safely traverse.

Robot velocity commands are generated by a set of behaviours. Each behaviour is intended to handle a specific situation or task and independently calculates whether it is currently useful (admissible) and what the robot should do (a velocity command). The behaviours are ordered by priority and the velocity commands from the highest priority admissible behaviour are given to the robot's lower level controller. Using a set of independent behaviours makes it easier to make changes to one behaviour or introduce new capabilities with minimal impact on other behaviours. Some of the behaviours used are: a behaviour to attempt to follow the path from hybrid A* while avoiding obstacles; a behaviour to move away from nearby obstacles when hybrid A* is not able to plan; a behaviour to attempt to prevent tipping over if dangerous pitches or rolls are experienced.

\subsection {UAV navigation}
\label{ss:uavnavigation}
The UAV utilises the navigation functionality commercially offered by Emesent as Autonomy Level 2, providing both local and global navigation solutions \cite{emesent}. The UAV features a manager node to coordinate local and global planning, and implement core behavioural primitives. The  manager receives higher level tasks from the operator, such as move to these waypoints, explore, etc. It coordinates activities to achieve these tasks, and interfaces with the lower level autonomy. This modular architecture has simplified the process of continual improvement to our higher level autonomy functionality.

To ease operator load, the UAV supports four major objectives: exploration, 3D waypoints, 2D waypoints, and planar waypoints. 3D waypoints specify an exact position in space where the UAV must move, useful to get the UAV into narrow openings when other alternatives are available (i.e., moving into a shed). 2D waypoints specify an X and Y location, but leaves height free, which is a useful general purpose commands. Planar waypoints simply require that the UAV reach any point on a user-specified plane, but does not specify where. It is often selected as a vertical plane, in order to provide a direction of travel without the need for a precise goal. This is useful for sending the UAV to a general location (i.e., not just basic exploration) in a space whose approximate layout is very roughly known, e.g., go 100\,m down a tunnel and then turn left at the branch. It can also serve as another form of more directed exploration.

Frontiers for the purpose of exploration are generated using the same solution as implemented for other robots, as described in Section \ref{ss:exploration}, again leveraging the homogeneous perception of the fleet. Frontiers of sufficient size are fed to the local navigation of the UAV as objectives at approximately 0.5 Hz, biasing for large, close frontiers in the same direction as the UAV is currently exploring.  The full criteria is available in \cite{williams_2020}. One weakness is that the UAV currently uses the direct metric distance to frontiers, rather than using the global map to reason about the presence of walls or similar obstructions.  In practice, this has not proved a significant drawback given the limited duration of UAV operations.  Exploration distances at DARPA competitions have been limited by the dimensions of the environment, with the UAV successfully exploring more than 300\,m from the start point and returning autonomously in tests in mines.

The UAV can reliably operate in a 3\,m tunnel at speeds of 2\,ms$^{-1}$. The UAV operated at 1\,ms$^{-1}$ both to optimise artefact detection, and because that is the speed at which the UAV has demonstrated the ability to perceive and thus avoid a vertically dangling 1\,mm wire. In an area known to feature wires no smaller than 4\,mm, the UAV can operate at 2\,ms$^{-1}$ and has demonstrated the ability to travel more than 600\,m underground and then return autonomously to the launch point in a single flight (total length $>$ 1.2\,km).

\section{Multi-Agent Mapping and Data Sharing}
\label{sec:multiagent}
The Subterranean Challenge requires the ability to coordinate robotic exploration in difficult three dimensional environments, often with degraded communication. Particularly during the Urban challenge event, we found that both single agents and small groups of agents were isolated from the operator and base station for periods up to ten minutes. To enable coordination, robots shared information with each other using a system called \emph{Mule} (Section \ref{ss:mule}), periodically broadcasting their data manifests and then requesting missing information from nearby (in the communication sense) agents. The operator's base station is considered an agent, which both receives information from the robotic agents and disseminates information to robots (i.e., when a new robot boots up, it receives information cached on the nearby base station). Agents share multiple types of data encapsulating mapping data, intent, artefact information and status.  

Robot coordination, and all shared information is referenced to Wildcat SLAM frames described in the next section. Waypoints, task goals, artefact locations and information generated from other agents can be interpreted, conditioned on receiving the respective frame.

\subsection{Multi-Agent SLAM}
\label{ss:slam}
The Wildcat Multi-Agent SLAM system is designed to compute global maps on each agent in a fully decentralised way, using all currently available information at each agent (i.e., a partial set of the union collected by all agents). Here we sketch its operation; further details will be provided in a forthcoming publication.

Wildcat works by integrating multi-sensor (lidar, IMU) data over a small time window (as described in Section \ref{ss:odometry}, and computing a SLAM \emph{frame} (Section \ref{sss:slamframes}) which contains features such as surfels and IMU estimates. Frames are shared between agents and each agent creates an \emph{Atlas Graph} (Section \ref{sss:posegraph}), solving for the global alignment of all frames. 

SLAM frames are the core building blocks on top of which all other information is referenced. For instance, a detected artefact's location is broadcast with respect to a corresponding SLAM frame. For other agents to interpret the location of this artefact they then require the corresponding SLAM frame as well. 

Since the ordering of frames received from different agents will not be identical on each agent, there is no formal guarantee of metric or topological consistency between the global maps computed by each agent. However, when the agents commence from a common starting area, the robustness is such that the solutions are sufficiently similar for all practical purposes. Furthermore, downstream processing (e.g., global navigation) is designed to be robust to these variations, keeping data in local frames so that it can be reinterpreted on other agents in light of the local SLAM solution.

The Wildcat SLAM solution produced a solution with error $<$ 0.05\% over distance distance travelled during the Tunnel circuit as discussed in Section \ref{ss:tunnel}. To the best of our knowledge, Wildcat was the only multi-agent SLAM solution run in a distributed manner for the most recent Urban circuit. Building on this framework has allowed us to have robots autonomously coordinate with each other when disconnected from the base station, and also played a key role in winning the \emph{most accurate artefact detection} discussed in Section \ref{sec:discussion}.

\subsubsection{SLAM Frames}
\label{sss:slamframes}

Wildcat generates and shares SLAM frames periodically over a small time time window (i.e., every five seconds a six-second frame is generated). The continuous trajectory that is generated by the odometry solution (as described in Section \ref{ss:odometry}) is used to project the multi-sensor data (i.e., lidar and IMU) into the fixed-world frame. This results in a spatially and temporally consistent set of information representing a portion of the odometry solution. As sending raw sensor data would result in prohibitive bandwidth requirements, the frames instead contain SLAM features that are derived from the raw sensor data.

\begin{figure}
    \centering
\includegraphics[width=\columnwidth]{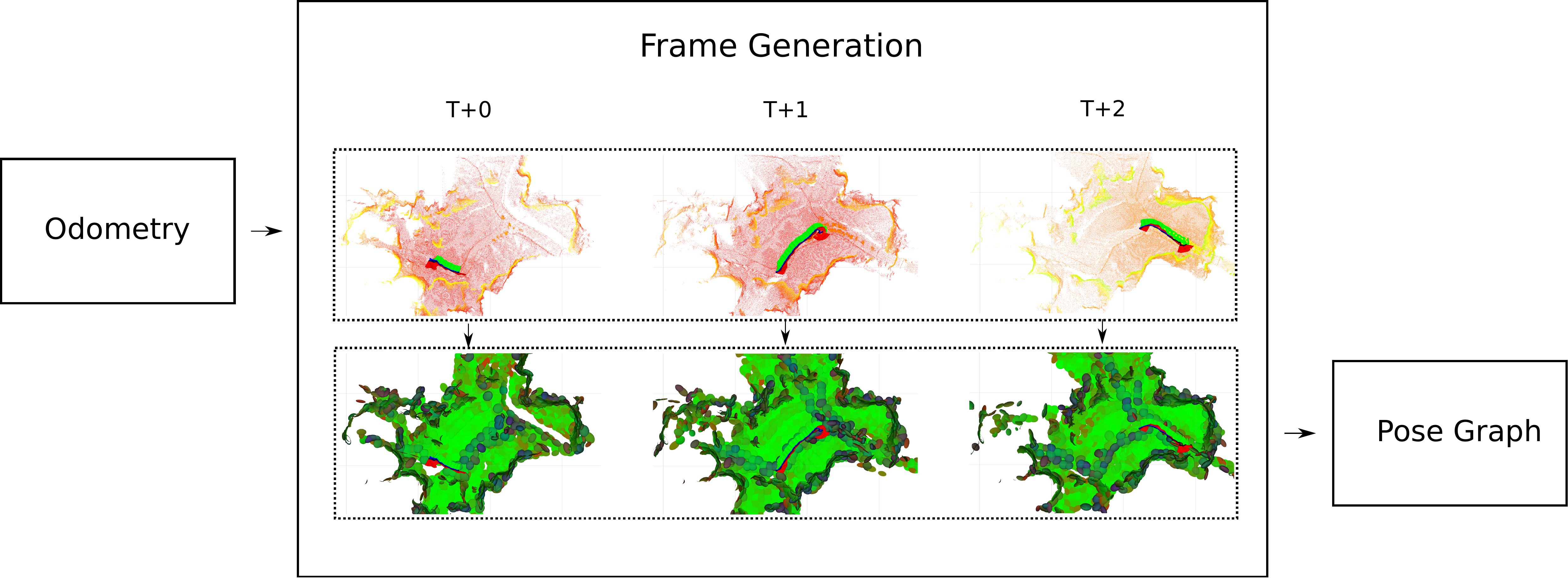}
    \caption{
        Three sequential portions of the odometry output and the associated features generated for the corresponding frames.
    }
    \label{fig:slam_frames}
\end{figure}

A SLAM frame contains the features required for the pose graph optimisation, described in Section \ref{sss:posegraph}. As illustrated in Figure~\ref{fig:slam_frames}, this includes surfels and an IMU integration estimate. Surfels are generated as described in Section \ref{ss:odometry} with the exception that each voxel does not have a temporal limit or bounding, resulting in a maximum of one surfel for each voxel. The IMU integration estimate is the mean of all IMU linear acceleration errors with respect to the odometry solution, producing an estimated gravity vector. The usage of surfels and an IMU integration estimate, rather than raw sensor data, also allows for the bandwidth requirements to be reduced when sharing the frames with other agents.

The size of a serialised SLAM frame used for sharing varies and the primary contributor for the size are surfels. The variation represents the observed geometric volume introduced by the voxelisation process used for the surfel generation. With a maximum lidar range of 100m the size can reach as much as 500\,KBytes in typical large/open environments. Since frames are generated every five seconds the effective bandwidth requirement is one-fifth of the indicated size. In the SubT Urban Circuit, the average size of a single frame over a representative agent's one hour mission was 168.8\,KBytes (yielding a data rate of 270.1\,KBit/sec), while in our cave testing, the average size was 102.6\,KBytes (yielding a data rate of 164.2\,KBit/sec).

\subsubsection{Atlas Pose Graph Optimisation}
\label{sss:posegraph}

\begin{figure}
    \centering
\includegraphics[width=\columnwidth]{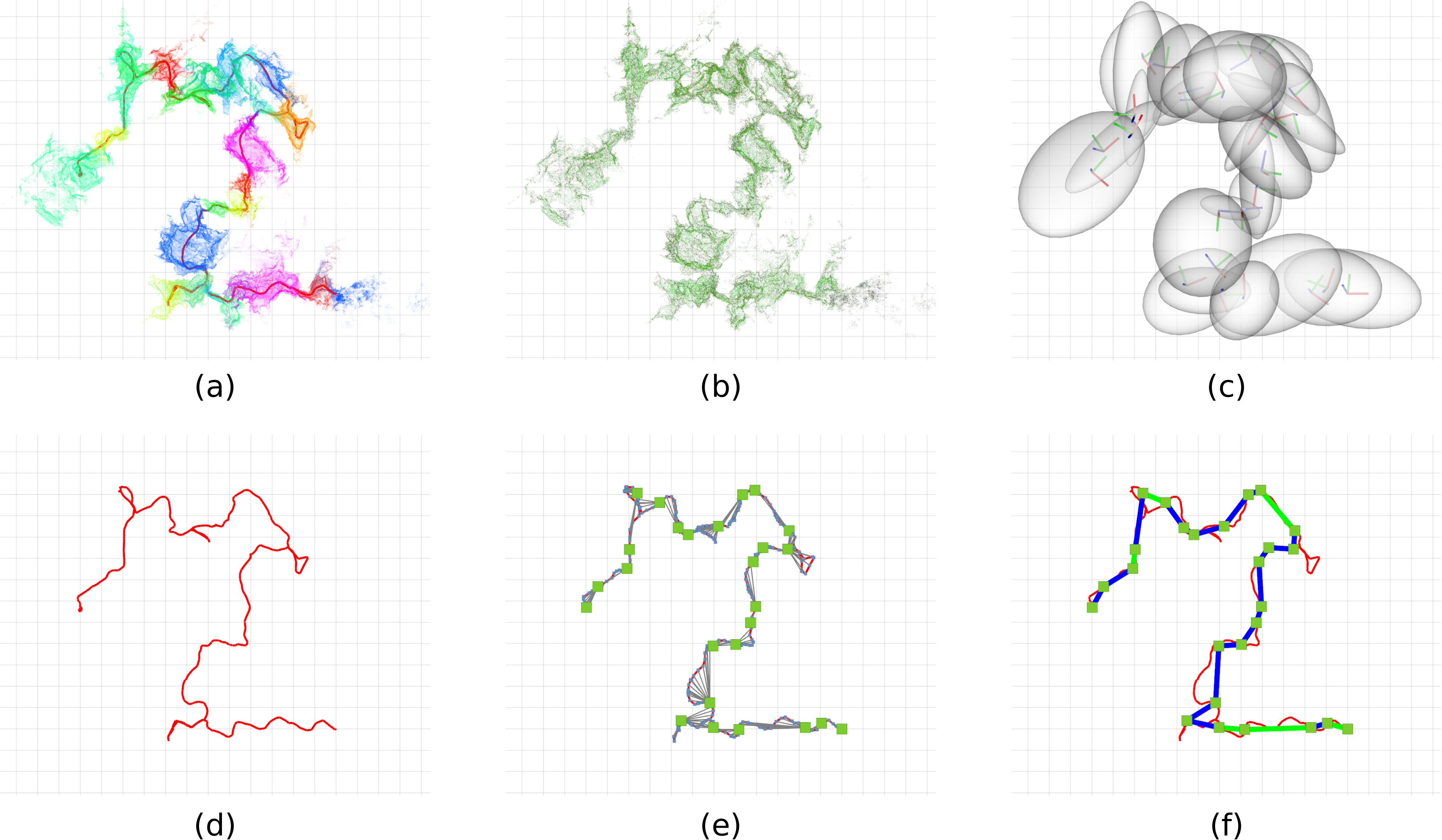}
    \caption{
        Various Pose Graph data representations for a single agent scan of a cave system, grid-size 10m.
        (a) Odometry trajectory (red) with point clouds coloured randomly according to the associated optimisation node
        (b) All surfel features present in the graph
        (c) All optimisation nodes and associated bounding zones representing the data distribution
        (d) Odometry trajectory (red)
        (e) Odometry trajectory (red), optimisation nodes (green), redundant nodes (blue), redundant nodes parent connections (grey)
        (f) Odometry trajectory (red), optimisation nodes (green), odometry edge constraints (blue), surfel match edge constraints (green).
    }
    \label{fig:pose_graph}
\end{figure}

As SLAM frames are generated and transmitted by agents, they are collected and optimised as a Hierarchical Pose Graph problem in a method similar to \cite{BosNew03}, as illustrated in Figure~\ref{fig:pose_graph}. Each SLAM frame represents a potentially optimisable pose, with edges between frames forming constraints on the final solution. Edges between frames are formed from various sources such odometry estimates, pair-wise frame geometry alignments, and loop-closure events. To enable better scaling for real-time multi-agent scenarios, redundant frames/edges are marginalised (shown in Figure~\ref{fig:pose_graph}e) which results in a graph structure whose complexity grows with distance travelled rather than time taken (shown in Figure~\ref{fig:pose_graph}f). Non-redundant frames are referred to as \emph{root frames}.

Edges within the graph are predominantly formed by odometry and pair-wise frame geometry alignment. Odometry edges formed between successive frames are guaranteed to exist, which ensures graph connectivity within an agent. In cases where frames overlap geometrically, a pose offset is generated using feature matching between the frames. For neighbouring frames ICP methods are used, while for distant frames pose-invariant methods are used. As the graph is optimised, outlier edges are detected and removed by analysing the residuals of the optimisation process. For edges where it is not possible to detect outliers, such as loop-closure or place-recognition events, hypotheses are used.

Hypotheses represent a potential edge between frames that requires more information to be validated or rejected. As an agent explores an environment the hypothesis is repeatedly checked for new overlapping frames which can be used. The amount of overlap required is determined by the amount of uncertainty which generated the hypothesis; i.e., larger loops require more overlap. Once a sufficient amount of overlap is detected the hypothesis is processed, where it is then added to the graph as additional constraints, or rejected for further processing. The validation step is performed using a derivation of clique maximisation for pairwise measurements as explored in \cite{ManDom18}.

During the DARPA events the agents were initially merged into the graph through the use of hypothesis hints. These hints act as seed edges within the graph and act in the same manner has a loop-closure or place-recognition edge. This method was also used when the UAV agents are deployed by the UGV agent carrying it. In this scenario the UGV provides a hint relative to itself which is used by the optimiser on the each agent to merge the graphs. The hint has an attached uncertainty value of 2.5\,m and 15$^\circ$ representing translation and rotation respectively. This uncertainty is often sufficient to initiate an immediate accept of the hypotheses, but in scenarios where it is not (due to featureless environments etc.) the agents must traverse the same area until it is.

Currently all agents broadcast all frames as the bandwidth can accommodate it. While this ensures that all agents have access to the complete set of data, it does not scale well as the bandwidth required increases linearly with each added agent. Additionally, as all frames are broadcast, an agent that is stationary will publish redundant information. Future work involves investigating delta-style communications or partial map-transfers.

\subsection{Multi-Agent UGV Global Navigation}
\label{sss:multiagentglobal}
Global navigation has evolved significantly as the challenge has progressed. For the Tunnel Circuit, global navigation relied on the SLAM Atlas graph, described in Section \ref{sss:slamframes}. Navigation to a distant waypoint occurred via intermediate nodes in the SLAM graph, which was restricted to the local agent's data (at the Tunnel Circuit, each agent only held its local SLAM map). In the more complex environments of the Urban Circuit, the global map was made up of SLAM frame poses (i.e., poses in which an agent has previously been), but connections between them were based on traversability analysis performed as a part of the frontier placement. Since the global, multi-agent map was now being solved in each agent, this allowed a shared global map, in which an agent could navigate to a pose previously held by any other agent.

Navigation in a cave presents significant challenges in determining what terrain is and is not traversable. The lethality determination described in Section \ref{ss:ugvnavigation} was leveraged to address this challenge, developing and maintaining a graph of the traversable terrain that has been mapped by all UGVs.

Thus we perform global navigation on a sparse topometric graph that is constructed from cost map data generated by all agents, and the Wildcat pose-graph solution. As cost maps are generated on an agent, they are associated with the most recent Wildcat frame. Cost maps associated with the same Wildcat frame are bundled together and transmitted to all other agents for integration into their topometric graph. The topometric graph maintains a collection of 2D submaps; one for each Wildcat root frame in the current pose-graph solution. Integration of a new cost map bundle involves merging cost map data into the corresponding submap. This approach handles 3D environments (e.g., overhangs) by relying on accurate gravity-alignment of cost maps and maintaining multiple submaps for Wildcat root frames where distinct heights have been observed for the same position projected on to the gravity-aligned plane. Superpixel methods are then applied to each submap to generate a subgraph suitable for navigation.

Inter-subgraph edges are computed via submap intersections for neighbouring Wildcat root frames. The complete topometric graph is thus composed of the vertices and edges of all subgraphs, plus the inter-subgraph edges. As the Wildcat pose-graph solution changes over time, submaps may be merged together and inter-subgraph edges may need to be recomputed in order to minimise computation costs and remain globally accurate.

\begin{figure}[!t]
    \centering
    \def \trajectory {{0,2.5},{0.5,2.5},{1,2},{1.5,1.5},{1.5,1},{1.5,0.5},{2,0},{2.5,0}}
	\def \segmentation {{0,5},{0,16},{0,23},{0,24},{0,35},{1,5},{1,16},{1,24},{1,25},{1,35},{2,5},{2,16},{2,25},{2,26},{2,35},{3,5},{3,16},{3,26},{3,35},{4,4},{4,5},{4,6},{4,7},{4,8},{4,16},{4,26},{4,27},{4,35},{4,36},{5,3},{5,4},{5,8},{5,9},{5,10},{5,11},{5,16},{5,27},{5,36},{6,0},{6,2},{6,3},{6,11},{6,16},{6,27},{6,28},{6,29},{6,30},{6,31},{6,35},{6,36},{6,37},{6,38},{6,39},{7,0},{7,1},{7,2},{7,11},{7,13},{7,14},{7,15},{7,16},{7,17},{7,27},{7,31},{7,32},{7,34},{7,35},{8,2},{8,11},{8,12},{8,13},{8,17},{8,18},{8,27},{8,32},{8,33},{8,34},{9,2},{9,11},{9,12},{9,18},{9,19},{9,27},{9,34},{9,35},{10,2},{10,12},{10,19},{10,26},{10,27},{10,35},{10,36},{11,2},{11,12},{11,19},{11,26},{11,36},{12,2},{12,11},{12,12},{12,19},{12,25},{12,26},{12,36},{13,2},{13,11},{13,19},{13,25},{13,36},{14,2},{14,3},{14,4},{14,5},{14,6},{14,9},{14,10},{14,11},{14,19},{14,20},{14,25},{14,36},{15,3},{15,4},{15,6},{15,7},{15,8},{15,9},{15,10},{15,19},{15,20},{15,21},{15,22},{15,23},{15,24},{15,25},{15,26},{15,36},{16,4},{16,10},{16,19},{16,26},{16,27},{16,36},{17,4},{17,10},{17,19},{17,27},{17,28},{17,29},{17,35},{17,36},{18,4},{18,10},{18,19},{18,29},{18,30},{18,35},{19,4},{19,10},{19,11},{19,19},{19,30},{19,35},{20,4},{20,11},{20,12},{20,13},{20,14},{20,15},{20,16},{20,17},{20,18},{20,19},{20,20},{20,30},{20,34},{20,35},{21,4},{21,11},{21,20},{21,30},{21,33},{21,34},{21,35},{21,36},{22,2},{22,3},{22,4},{22,10},{22,11},{22,20},{22,21},{22,30},{22,33},{22,36},{22,37},{22,38},{23,0},{23,1},{23,2},{23,4},{23,9},{23,10},{23,21},{23,30},{23,32},{23,33},{23,38},{23,39},{24,4},{24,5},{24,9},{24,21},{24,22},{24,30},{24,31},{24,32},{25,5},{25,6},{25,7},{25,8},{25,9},{25,10},{25,22},{25,30},{25,31},{26,7},{26,8},{26,10},{26,11},{26,22},{26,30},{27,6},{27,11},{27,12},{27,21},{27,22},{27,27},{27,28},{27,29},{27,30},{28,5},{28,6},{28,12},{28,13},{28,21},{28,22},{28,23},{28,24},{28,25},{28,26},{28,27},{28,29},{28,30},{29,3},{29,4},{29,5},{29,13},{29,14},{29,20},{29,21},{29,30},{30,2},{30,3},{30,14},{30,15},{30,19},{30,20},{30,30},{30,31},{31,2},{31,14},{31,15},{31,16},{31,17},{31,18},{31,19},{31,20},{31,31},{31,32},{32,2},{32,3},{32,14},{32,16},{32,17},{32,20},{32,32},{32,39},{33,3},{33,13},{33,14},{33,20},{33,21},{33,32},{33,36},{33,37},{33,38},{33,39},{34,0},{34,1},{34,2},{34,3},{34,4},{34,12},{34,13},{34,21},{34,32},{34,35},{34,36},{35,0},{35,4},{35,11},{35,12},{35,21},{35,32},{35,33},{35,34},{35,35},{36,4},{36,5},{36,6},{36,10},{36,11},{36,21},{36,22},{36,32},{36,33},{37,6},{37,7},{37,8},{37,9},{37,10},{37,22},{37,32},{38,8},{38,9},{38,21},{38,22},{38,32},{39,9},{39,10},{39,21},{39,32}}
    \def \centroids {{1,11},{2,1},{2,31},{2,38},{6,22},{9,6},{13,31},{13,38},{14,15},{17,1},{20,7},{21,25},{25,16},{27,1},{28,36},{31,8},{34,26},{37,16},{37,37},{38,3}}
    \def \adjacencies {{2/6},{13/17},{3/8},{13/16},{3/7},{9/11},{14/20},{7/12},{6/10},{15/17},{17/19},{5/7},{4/8},{2/1},{8/15},{10/14},{12/13},{5/9},{6/9},{9/12},{6/11},{10/11},{12/17},{16/18},{11/13},{7/15},{1/6},{12/15},{3/4},{9/13},{1/9},{1/5},{2/10},{15/19},{17/18},{5/12},{5/3},{13/18},{16/20},{11/14},{14/16},{7/8}}
    \def \intersectioni {14,16,20}
    \def \intersectionj {4,8,3,7}
    \begin{subfigure}{.44\linewidth}
        \begin{tikzpicture}[scale=.5,every node/.style={minimum size=1cm},on grid]
            \begin{scope}[yshift=-4cm,every node/.append style={yslant=0.5,xslant=-1},yslant=0.5,xslant=-1] \draw [black, dashed] (0,0) rectangle (4,4);
                \fill [white, fill opacity=1] (0,0) rectangle (4,4);
                \foreach [count=\i] \p in \centroids {
                    \coordinate (c\i) at ($0.1*(\p) + (0.05,0.05)$) {};
                    \fill [black] ($0.1*(\p) + (0.05,0.05)$) circle (0.1);
                }
                \foreach \i/\j in \adjacencies {
                    \draw [black] (c\i) -- (c\j);
                }
                \coordinate (subgraphcorner) at ($(4,0) + (-0.05,0.0)$);
            \end{scope}
            \begin{scope}[yshift=-2cm,every node/.append style={yslant=0.5,xslant=-1},yslant=0.5,xslant=-1] \draw [black, very thick] (0,0) rectangle (4,4);
                \fill [white, fill opacity=1] (0,0) rectangle (4,4);
                \foreach \p in \segmentation {
                    \fill [gray] ($0.1*(\p) $) rectangle +(0.1,0.1);
                }
                \foreach \p in \centroids {
                    \fill [black] ($0.1*(\p)$) rectangle +(0.1,0.1);
                }
                \coordinate (segmentationcorner) at ($(4,0) + (-0.05,0.0)$);
            \end{scope}
            \begin{scope}[yshift=0,every node/.append style={yslant=0.5,xslant=-1},yslant=0.5,xslant=-1] \draw [black, thick] (0,0) rectangle (4,4);
                \fill [white, fill opacity=1] (0,0) rectangle (4,4);
                \foreach \p in \trajectory {
                	\draw [black] (\p) rectangle +(1.5,1.5);
                    \fill [black, fill opacity=0.3] (\p) rectangle +(1.5,1.5);
}
                \foreach [count=\i] \p in \trajectory {
                    \coordinate (t\i) at ($(\p) + (0.75,0.75) + (2,2)$) {};
                    \coordinate (trajectoryend) at (t\i) {};
                    \fill [white] ($(\p) + (0.75,0.75)$) circle (0.05);
                }
                \draw [red,line width=1.5pt, preaction={draw=black, line width=2pt, cap=round}] \foreach \i [remember=\i as \lasti (initially 1)] in {2,...,8}{(t\lasti) -- (t\i)};
                \foreach [count=\i] \p in \trajectory {
                    \draw [dotted, black, line width=0.5pt] (t\i) -- ($(\p) + (0.75,0.75)$);
                }
                \coordinate (costmapcorner) at ($(4,0) + (-0.05,0.0)$);
            \end{scope}
            \begin{scope}[yshift=4cm,every node/.append style={yslant=0.5,xslant=-1},yslant=0.5,xslant=-1] \coordinate (root) at (2,2) {};
                \draw [red, line width=1pt, preaction={draw=black, line width=2pt, cap=round}] (root) -- +(0,-1);
                \draw [green, line width=1pt, preaction={draw=black, line width=2pt, cap=round}] (root) -- +(-1,0);
                \draw [blue, line width=1pt, preaction={draw=black, line width=2pt, cap=round}] (root) -- +(1,1);
                \foreach [count=\i] \p in \trajectory {
                    \draw [dotted, black, line width=0.5pt] (t\i) -- (root);
                }
            \end{scope}
            \coordinate (rootelbow) at ($(root) + (0.25,0.25)$) {};
            \draw[<-] (rootelbow) to[out=45, in=180] +(2,0) node[right] (rootlabel) {Frame};
            \draw[<-] (trajectoryend) to[out=90, in=180] +(2,0) node[right] (trajectorylabel) {Robot trajectory};
            \draw[<-] (costmapcorner) to[out=90, in=180] +(2,0) node[right] (costmaplabel) {Submap (merged costmaps)};
            \draw[<-] (segmentationcorner) to[out=90, in=180] +(2,0) node[right] (segmentationlabel) {Segmentation};
            \draw[<-] (subgraphcorner) to[out=90, in=180] +(2,0) node[right] (subgraphlabel) {Subgraph};
        \end{tikzpicture}
    \caption{}
    \end{subfigure}
    \hfill
    \begin{subfigure}{.44\linewidth}
        \begin{tikzpicture}[scale=.5,every node/.style={minimum size=1cm},on grid]
            \begin{scope}[yshift=0, every node/.append style={yslant=0.5,xslant=-1},yslant=0.5,xslant=-1] \draw [black, dashed] (0,0) rectangle (4,4);
                \fill [white, fill opacity=1] (0,0) rectangle (4,4);
                \foreach [count=\i] \p in \centroids {
                    \coordinate (ci\i) at ($0.1*(\p) + (0.05,0.05)$) {};
                }
                \foreach \i/\j in \adjacencies {
                    \draw [green] (ci\i) -- (ci\j);
                }
            \end{scope}
            \begin{scope}[yshift=0, xshift=5cm, every node/.append style={yslant=0.5,xslant=-1},yslant=0.5,xslant=-1] \draw [red, dashed] (0,0) rectangle (4,4);
                \foreach [count=\i] \p in \centroids {
                    \coordinate (cj\i) at ($0.1*(\p) + (0.05,0.05)$) {};
                }
                \foreach \i/\j in \adjacencies {
                    \draw [red] (cj\i) -- (cj\j);
                }
            \end{scope}
            \begin{scope}[yshift=0, xshift=2.5cm, every node/.append style={yslant=0.5,xslant=-1},yslant=0.5,xslant=-1] \fill [white] (1.25,1.25) rectangle ($(2,2) + (0.75,0.75)$);
                \coordinate (intersectioncentre) at (2,2) {};
            \end{scope}
            \begin{scope}[yshift=0, xshift=0cm, every node/.append style={yslant=0.5,xslant=-1},yslant=0.5,xslant=-1] \foreach \p in \centroids {
                    \fill [green] ($0.1*(\p) + (0.05,0.05)$) circle (0.1);
                }
                \coordinate (rooti) at (2,2) {};
                \draw [red, line width=1pt, preaction={draw=black, line width=2pt, cap=round}] (rooti) -- +(0,-1);
                \draw [green, line width=1pt, preaction={draw=black, line width=2pt, cap=round}] (rooti) -- +(-1,0);
                \draw [blue, line width=1pt, preaction={draw=black, line width=2pt, cap=round}] (rooti) -- +(1,1);
            \end{scope}
            \begin{scope}[yshift=0, xshift=5cm, every node/.append style={yslant=0.5,xslant=-1},yslant=0.5,xslant=-1] \foreach \p in \centroids {
                    \fill [red] ($0.1*(\p) + (0.05,0.05)$) circle (0.1);
                }
                \coordinate (rootj) at (2,2) {};
                \draw [red, line width=1pt, preaction={draw=black, line width=2pt, cap=round}] (rootj) -- +(0,-1);
                \draw [green, line width=1pt, preaction={draw=black, line width=2pt, cap=round}] (rootj) -- +(-1,0);
                \draw [blue, line width=1pt, preaction={draw=black, line width=2pt, cap=round}] (rootj) -- +(1,1);
            \end{scope}
            \draw[yellow, line width=1pt, preaction={draw=black, line width=2pt, cap=round}] (rooti) to[out=45,in=135] (rootj);
            \draw [thick] ($(rooti)!0.5!(rootj) + (0,1)$) to[out=45,in=-45] +(0,1.5) node[above] (connectionlabel) {Connection between frames};
            \draw [thick] (intersectioncentre) to[out=0,in=90] +(0,-2) node[below] (intersectionlabel) {Intersection used for inter-subgraph edge finding};
            \foreach \i in \intersectioni {
                \foreach \j in \intersectionj {
                    \draw[gray] (ci\i) -- (cj\j);
                }
            }
        \end{tikzpicture}
    \caption{}
    \end{subfigure}
    \caption{Topometric map generation involves (a) derivation of subgraphs from the superpixel segmentation of submaps associated with each root frame, and (b) finding traversable edges between vertices of different subgraphs by inspecting the intersection of submaps from connected frames}
    \label{fig:topomap}
\end{figure}
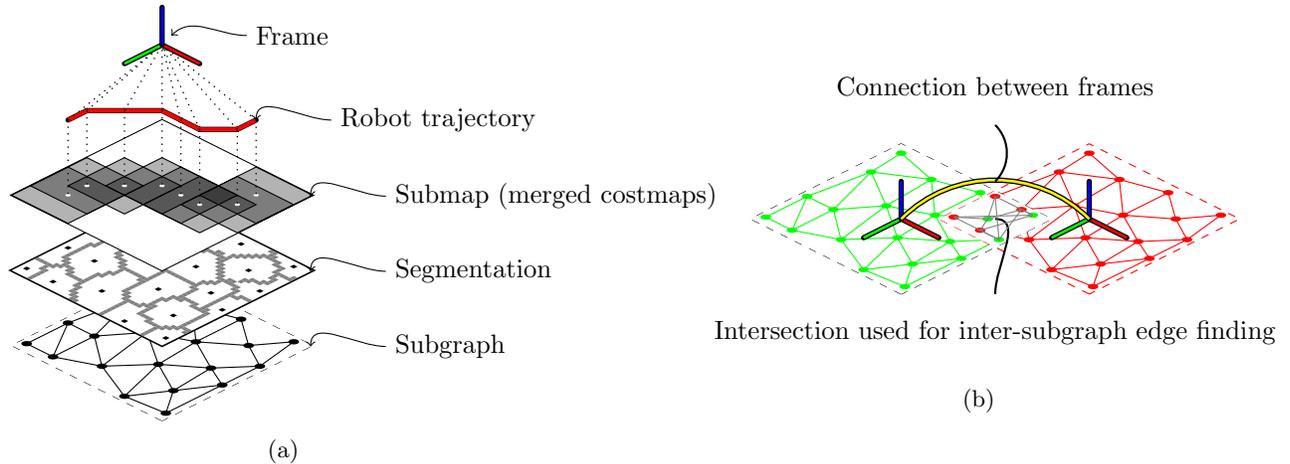

\subsection{Data Sharing}
\label{ss:mule}
Data-sharing over the Rajant mesh network required a communications strategy that could robustly handle the highly dynamic and unreliable network conditions encountered in SubT. Since our software stack was primarily ROS based and a single unified ROS system is not well suited to distribution across an unreliable network (\cite{Tiderko16}), we opted to run an independent ROS system on each agent with a multi-master communication layer to handle data-sharing amongst agents.

At the STIX and Tunnel events, we used an off-the-shelf commercial software package and the open-source rosbridge\_suite ROS package to handle multi-master communication. The commercial software was marketed primarily for cloud-based monitoring/control of robotic fleets and required a `gateway' process (running at the base station) to broker any connections between `client' processes (running on each agent). This limited communication to times when there was simultaneous connectivity between the base station and the agents trying to communicate, and encumbered autonomous coordination at range. Agents that spent long periods without connectivity to the base station risked losing critical messages such as artefact detections due to a lack of persistent storage for the data sharing layer. Subsequently, we chose to abandon these packages and develop a replacement that more specifically suited our needs.

Our replacement solution, Mule, is a ROS package that provides multi-master, disruption-tolerant, transparent bridging of ROS topics. Mule processes dynamically discover one another on the network via UDP multicast, and establish peer-to-peer (P2P) connections via TCP unicast. Messages associated with ROS topics configured for bridging are multiplexed onto the P2P connections according to quality-of-service (QoS) options and link quality metrics provided by the local Rajant node. Mule can bridge a particular topic with either a `volatile' or `persistent' QoS. 

A `volatile' QoS is suitable for ROS topics that do not require guaranteed delivery of every message (e.g. status, teleop commands, and camera images). Mule uses the ZeroMQ library and its PUB/SUB sockets to transport these messages to interested receivers in a best-effort fashion. 

Conversely, a `persistent' QoS is suitable for ROS topics that do require guaranteed delivery (e.g. Wildcat frames and artefact detections). An on-disk database is used to store messages from `persistent' ROS topics, so that they may be retrieved for delivery to a peer at some later opportunity (in a manner similar to \cite{Miller20}). Guaranteed delivery of these messages is thus achieved by the synchronisation of databases amongst Mule processes. Each Mule periodically assembles a `manifest' compactly describing which messages are currently stored in their database, and broadcasts this `manifest' to its peers. Peers can then use this information to request the transfer of certain messages in order to achieve synchronisation. This pull-based approach is robust in the face of unreliable network conditions and allows for inter-agent data-sharing without requiring end-to-end paths.

Additionally, Mule publishes information to the local ROS system to enable autonomous decision making around data-sharing. This includes synchronisation status for each peer, making it possible for agents to decide if they should return to base to deliver mission-critical data, as was done at the Urban and Cave events.

\section{Human Robot Team Decision Making}
\label{sec:humanrobotteam}
In the SubT challenge, a single operator must control a team of robots. The difficulty for the operator is exacerbated when dealing with heterogeneous platforms with different capabilities. Incorporating human knowledge with robot autonomy to create an effective human-multi-robot team is therefore a key problem to solve. 

As the challenge has progressed, our solution has matured towards greater autonomy. The design employed in STIX and Tunnel Circuit utilised FSA-based control, predominantly executing waypoints or location based goals created by the operator. The high level FSA control design continued in Urban, but was mostly used through autonomous explore-sync missions, with the operator providing a starting location from which to explore, and subsequent execution continued for extended periods with no operator intervention.

In Cave testing, we utilised a general task allocation framework which enables autonomous task coordination among robots, but allows the operator to inject human guidance to influence the team’s behaviours. The end-goal is to have the operator to focus on high-level decisions to accomplish missions at the team level instead of individual behaviour at the robot level.

\subsection{Finite State Automaton Design}
\label{ss:fsa}
Low-level autonomy is implemented via the dynamic construction and execution of nested finite state automatons (FSAs). A set of primitive behaviours interact with other systems/nodes, e.g., a goto behaviour interfaces with the navigation systems to drive the agent to a goal location. More complex behaviours can be built as FSAs of these primitive behaviours, e.g., a drop-comms-node behaviour sequences a goto behaviour for the goal location, a behaviour to execute the node-drop, and a behaviour which moves away from the node to avoid accidental collisions.

The operator can specify linear sequences of simple parameterised behaviours via `command lists'. The executive node parses an incoming command list, constructs an FSA, and executes it. Execution involves `ticking' the FSA at 1\,Hz, limiting the speed at which the FSA can transition between states. Individual behaviours may perform computation or interact with the system at any rate, but are only entered or exited on a `tick'.

The primary behaviours needed in the SubT challenge have been: goto, explore, sync, drop-comms-node and launch-UAV. The goto behaviour takes a desired goal pose and engages the global planner to move the agent towards that goal. The explore behaviour reads frontier information and engages the global planner to pursue the selected frontiers. The sync behaviour reads Mule information and engages the global planner to move the agent back towards the base station until their Mule databases have synchronised. The drop-comms-node behaviour engages the comms-node-launcher and global-planner in order to drop a comms-node and move the agent to a safe distance. The launch-UAV behaviour engages the launch-platform to disengage latches holding the UAV to the agent, so that it may then begin take-off.

\subsection{Multiple-Robot Task Allocation}
\label{ss:taskallocation}
The multi-agent coordination problem is solved using a market-based task allocation approach. Agents bid on tasks by estimating the expected reward if they execute it. For simple cases and certain auction mechanism, the optimal allocation of a set of tasks amongst a team can be found (\cite{bertsekas1990auction}). However, the general case is NP-hard (\cite{gerkey2004formal}), so often in robotics literature simpler mechanisms are used that find good sub-optimal solutions while requiring less computation and communication, e.g., \cite{zlot2006market, ulam2007integrated, otte2017multi}. 

One of the major challenges in SubT is the dynamic nature. The environment and tasks are not known ahead of time. Task allocation must continuously update as new tasks are discovered/created and old ones are completed. Agents often lose communication with each other. Even team composition changes over time as platform failures are not uncommon in the difficult environments.

The core tasks defined and used so far are explore, drop-comms-node, and sync-data (for agents who are out of comms range). Each task has a reward, which is discounted by the expected time it would take an agent to complete the task. Agents bid on tasks by building bundles, an ordered list of tasks. When considering the bid for a new task, the difference in total reward between the new bundle and the original bundle is the bid placed the task. This approach allows an agent to consider what tasks are efficient to execute together, generally due to close positions. However, it avoids the full combinatorial calculation to find the optimal bundle, only to have the state of tasks change and the process repeated.

Tuning of rewards for use in task allocation is a challenging problem. To simplify, a hybrid system of reward and priority was introduced, where a higher priority acts as an infinitely higher reward within the task allocation system, which provides a simple way for the designer or operator to ensure certain tasks are executed first (or last) without estimating reward adjustments. The initial version of task allocation utilised at the cave event used fixed rewards for frontiers, but varied the priority between three values based on the size of the frontiers, with thresholds selected based on simulated and real data. This resulted in the behaviour to initially follow the main trunk of the environment, and subsequently fill in details. Further work has been conducted since based on extensive robot testing in a variety of environments, focusing both on improvements to frontier representation, and more directly on reward tuning.

\begin{figure}[ht]
    \centering
    \includegraphics[width=\columnwidth]{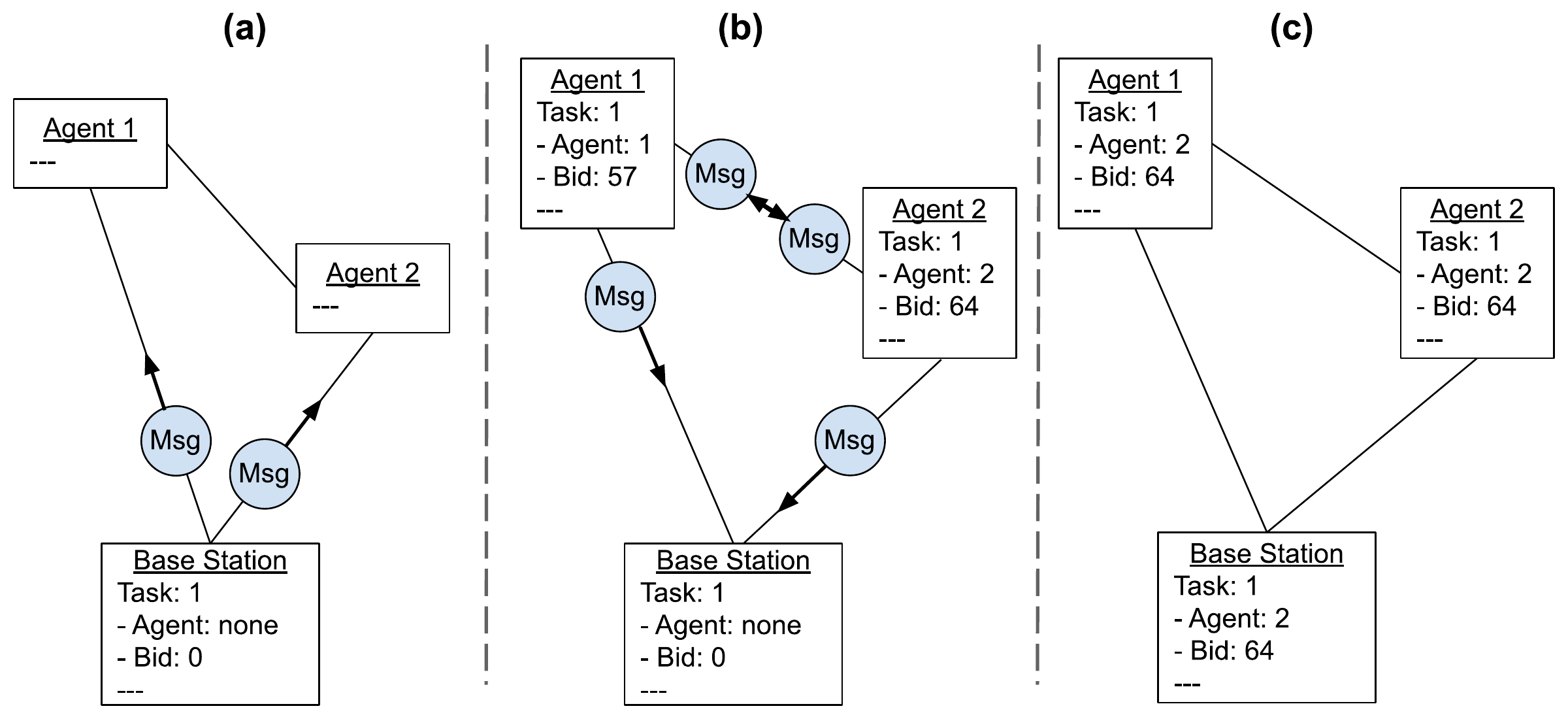}
    \caption{Simple consensus example. (a) Operator station sends a drop-node task. (b) each agent bids and broadcast their bid across the network. (c) Agents and base station form consensus on who will execute task.}
    \label{fig:consensus}
\end{figure}

Auctions use a consensus protocol, based on the work from \cite{choi2009consensus}, but extended to deal with continuously updating the task set. Available tasks are broadcasted to all agents, and any agent bidding on a task broadcasts its ownership and bid on the selected task to all other agents. If multiple agents bid, every agent will receive the bids and use the same rules to determine who wins. A simple example of this process is shown in Figure  \ref{fig:consensus}. These auctions are completely decentralised, and agents can generate and distribute tasks amongst any sub-group that shares communication.

Determining the winner of an auction is simple (the highest bid wins) but unreliable communication can lead to agents with more complicated conflicts. Two agents may execute the same task, only one agent may know a task has been completed, an agent may be unaware that another agent released a task it was executing, etc. The key is to carefully develop the rules such that when agents regain communication and share their task sets with each other, they will form a consensus about the state of all tasks. In practice, many difficulties are mitigated by the fact that agents that are competing for tasks are generally close, and thus are able to communicate.

\begin{figure}[ht]
    \centering
    \includegraphics[width=\columnwidth]{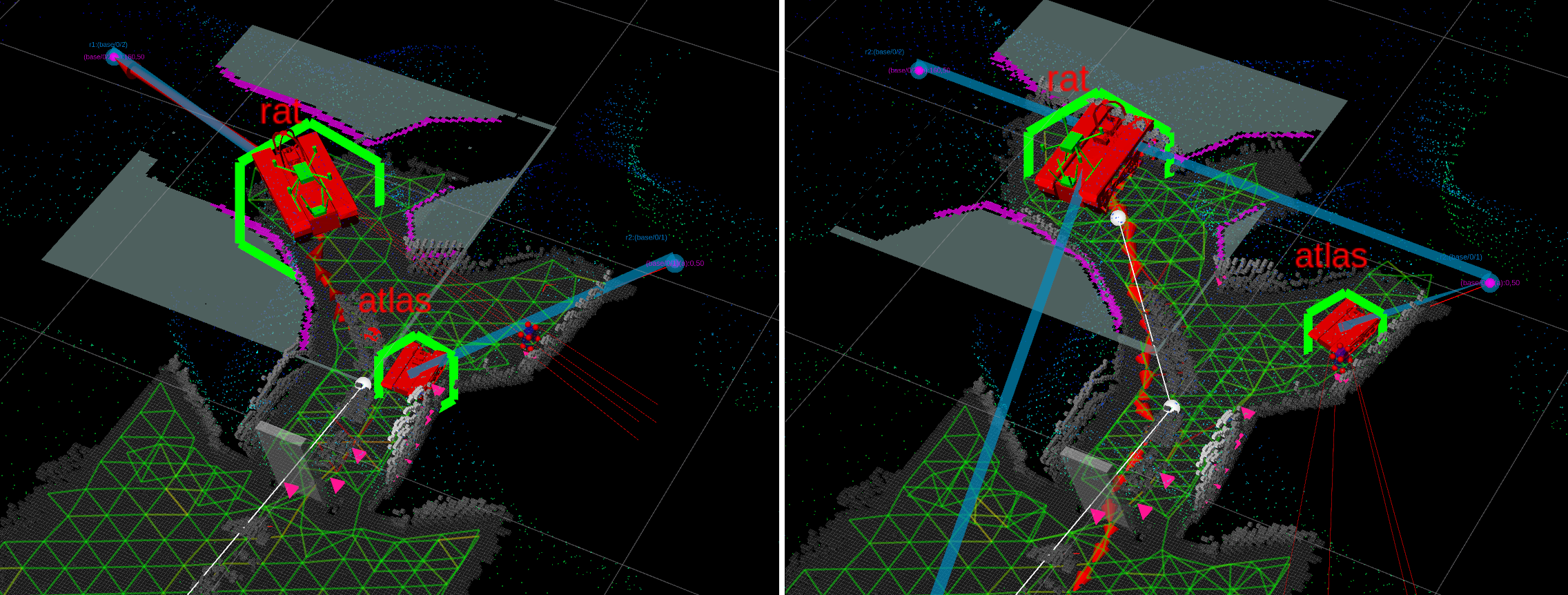}
    \caption{(Left) Two agents distribute tasks to investigate separate areas between themselves. The light-blue lines represent the 'task-path', the path the agent expects to take as it completes its bundle of tasks. (Right) One agent is tasked to return home, the other agent immediately picks up the released task to execute itself.}
    \label{fig:ta_example}
\end{figure}

Figure~\ref{fig:ta_example} shows an example of the potential coordination. When one agent is tasked by the operator to return home, another agent immediately takes the tasks it released over. This task allocation approach is enabled by the homogeneous sensing of the platforms, which permits tasks and navigation data from one agent to be shared in local coordinate frames which are solved by Wildcat on other agents. Subsequently, other agents can determine their ability to execute the task, and submit bids as required. 

\subsection{Incorporating Operator Guidance}
\label{ss:taoperatorinput}
Incorporating operator feedback into a complex multi-agent system such as this is an ongoing challenge. The current implementation allows an operator to place manual tasks, allowing the operator to point out tasks to be completed without requiring manual selection of an agent. A priority-region tool has also been developed, allowing the operator to create 3D regions that modify the reward and priority of tasks within them.

\begin{figure}[ht]
    \centering
    \includegraphics[width=\columnwidth]{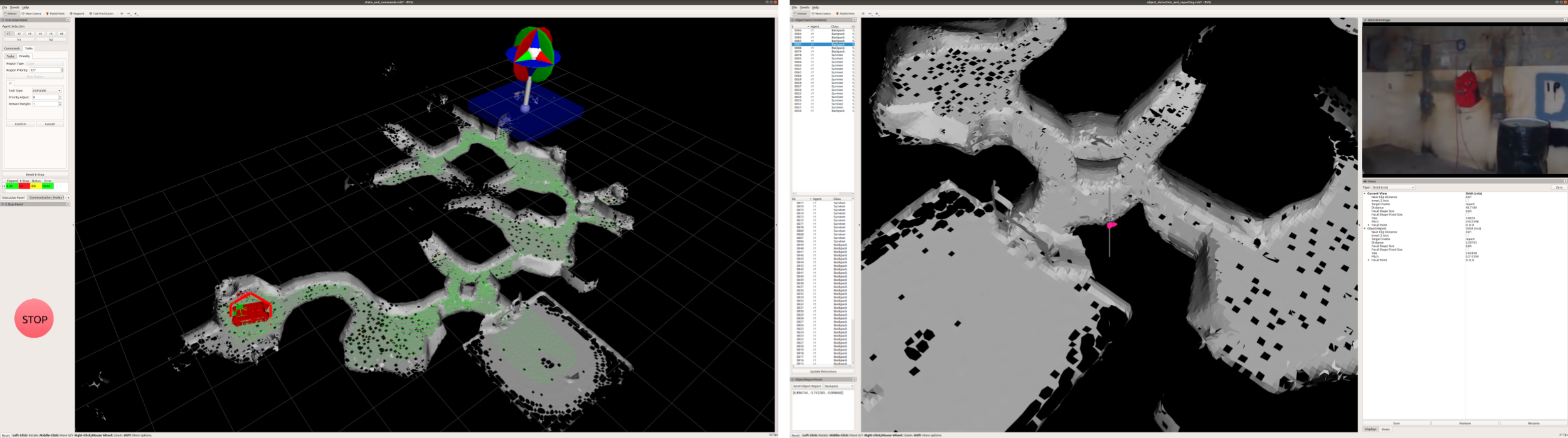}
    \caption{(Left) Command panel of operator GUI with example of operator command input in form of task prioritisation region in blue. (Right) Object panel of operator GUI with a detection selection from detection list, visualised as a location on the map and the detection image for operator verification and potential report submission.}
    \label{fig:gui_example}
\end{figure}

Priority regions allow the operator to prioritise areas they believe are important, or effectively block off regions they believe are unproductive or dangerous for platforms. These regions can be set to apply to specific task types or agents, also allowing the operator to ensure a specific agent executes certain tasks or works in a certain area. This is a flexible and powerful tool for an operator to influence task allocation, but one that can be difficult to use. Continued work is looking at ways to simplify and streamline the common actions operators wish to perform, such as direct assigning of tasks to specific agents, without needing to create regions.

The operator interface is shown in Figure~\ref{fig:gui_example}, specifically illustrating the interface for generating an arbitrary task prioritisation region. The right-hand figure shows the interface used to review artefact detections.

\section{Discussion}
\label{sec:discussion}

The first two phases of the DARPA Subterranean Challenge spanned two years of development, and evaluated the systems performance through four field trials. The first SubT Integration Exercise (STIX) took place six months after project inception at Colorado’s Edgar Experimental Mine in April 2019. This event was planned to be followed by three circuit events, where each competing team was ranked by the number of artefacts found. The Tunnel circuit was held at the Bruceton Experimental Mine, U.S. Bureau of Mines, in August 2019, and the Urban Circuit was held at Satsop Business Park in Elma, Washington in a partially-complete, abandoned, nuclear power plant in February 2020. The Cave Circuit, planned for August 2020, was cancelled due to global travel restrictions from COVID-19. The CSIRO Data61 Team instead travelled to the Carpenteria Cave systems, a naturally formed limestone cave in Chillagoe, Queensland, and progress was evaluated by conducting field tests that followed the format of the official circuit events. 

These formal periodic evaluations provide a rigorous set of tests to analyse the advantages and deficiencies of the developed approaches. As system development occurred throughout the phases, circuit testing results fed back into prioritising work and changing the approaches taken. The CSIRO Data61 Team entered the program with a mature SLAM solution \emph{Wildcat}, and UAV autonomy technology, which spun out from CSIRO into Emesent during Phase I. The modular UGV \emph{CatPack} was developed to complement Emesent's Hovermap payload used on the UAVs, and allowed greater heterogeneity in UGV platforms. UGV platforms where changed for each event, as compared to other teams such as Explorer and CTU-CRAS, who reused their highly effective platforms from early events. The deployed communications system, UGV fleet, and multi-agent autonomy and operational concepts changed significantly for each event, pivoting based on testing results. Overwhelmingly the importance of robotic autonomy and a coordinated robotic fleet was demonstrated, and prioritised over continuing to develop better communication systems and relying on a single operator for all higher level guidance. 

A nascent system was deployed at STIX, with the UAV and UGVs running independently, with separate Ubiquiti point-to-point radios, and Emesent/DJI controllers. All systems could only be commanded via waypoints and there were no higher levels of autonomy. The difficulty of the encountered terrain during testing and subsequent site inspections demonstrated the need for larger, robust UGVs.   

\begin{figure}[t]
    \centering
    \begin{subfigure}{.30\columnwidth}
    \includegraphics[height=40mm]{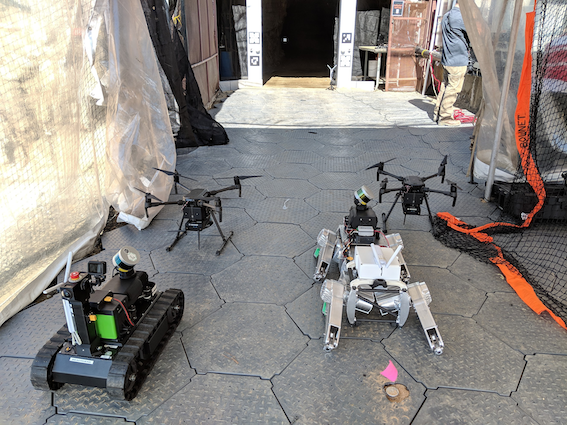}
    \caption{}
    \label{fig:stix_robots_and_team}
    \end{subfigure}
    \begin{subfigure}{.35\columnwidth}
    \includegraphics[height=40mm]{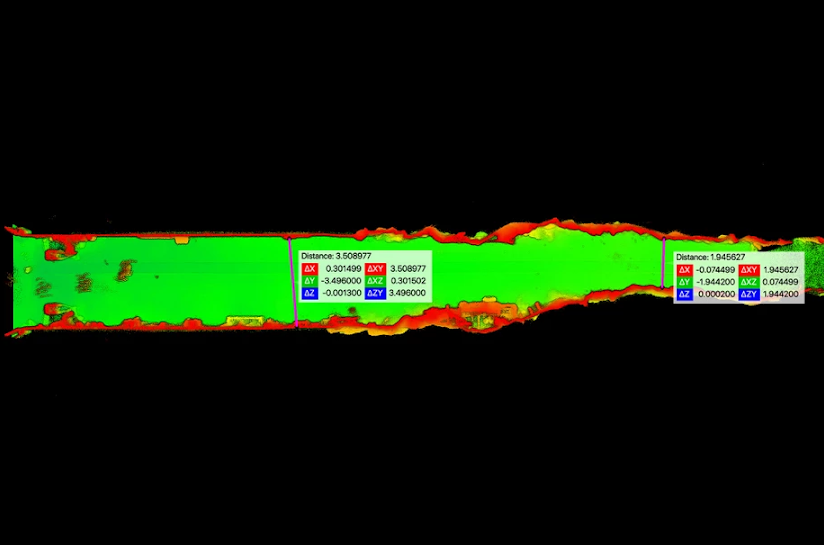}
    \caption{}
    \label{fig:stix_drone_constriction}
    \end{subfigure}
    \begin{subfigure}{.31\columnwidth}
    \includegraphics[height=40mm]{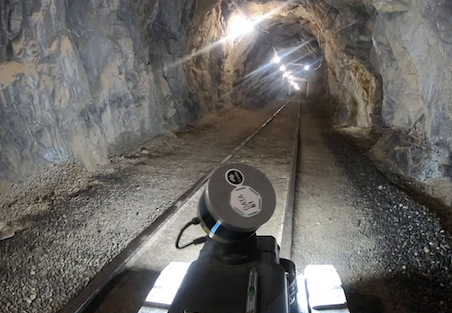}
    \caption{}
    \label{fig:stix_rails_UGV}
    \end{subfigure}
    \caption{Robot team utilised at STIX at starting gate (a), consisting of two Emesent Hovermap UAVs, a SuperDroid LT2-F and a Ghost Robotics Vision60 quadruped prototype (v3.2) automated with modular CatPacks. The Miami Tunnel at STIX constricted to under two metres width within 30\,m of the entrance (b),preventing further exploration with the UAV system. The rails in the Miami Tunnel caused the SuperDroid LT2-F platform to get stuck after traversing 120\,m due to its low ground clearance (c).}
    \label{fig:stix}
\end{figure}

Performance in the Tunnel Circuit was restricted by the reliance on an unreliable mesh communication system. The operator base station computed a joint global map from all available agent data allowing unified commanding in one map, where each robot was capable of simple exploration, waypoint following, and a return state. The success of autonomy versus manual waypoints led to prioritisation of further autonomy development, shutting down development of a CSIRO-communications system, and purchasing a dedicated Rajant mesh system for the subsequent circuits. An initial long ($\sim$250\,m) non-branching tunnel topology prompted the marsupial approach to UGVs carrying UAVs, increasing the number of large UGVs in the fleet. The lack of reliability in both legged systems and the excellent performance of the ATR platform focused UGV development entirely onto tracked platforms. 
 
By the Urban Circuit, each agent computed an on-board global map from shared data, which enabled implicit coordination between agents prioritising regions away from other agents. The explore-sync behaviour in \cite{williams_2020} was the primary mode of operation, and the most successful agents operated beyond communication range for extended durations. The marsupial UAV system allowed for targeted exploration through shafts, and the small UGVs were used to traverse stairwells. Several points were missed by not-traversing small doorways and barrier based switchbacks during exploration, leading to a focus on making global and local traversability consistent and shared. The clumping of exploring agents led to adoption of multi-robot task allocation. 

The expectation for the cave circuit further focused development on terrain costing and the ability to rationalise over unobserved ground. The robots primarily operated in the multi-robot task allocation mode, with the operator falling back to assist with stuck robots or prioritising tasks.

\subsection{STIX}
\label{ss:stix}
The team at STIX consisted of two Emesent Hovermap UAVs, one SuperDroid LT2-F, and a Ghost Robotics Vision60, as illustrated in Figure~\ref{fig:stix_robots_and_team}. In different runs, the SuperDroid and UAV each traversed 120\,m, while the Ghost's best result was 40\,m. 

It is believed that the Ghost slipped on rails in the tunnel, while the same type of obstacle caused the SuperDroid to become beached in one run. Two key advantages of the SuperDroid were its ease of repair, and ability to be carried as checked luggage, whereas the Ghost was not field repairable by our team, and needed to be disassembled into multiple cases to meet the checked luggage weight requirements. The rough terrain and issues with beaching drove the decision to acquire the large BIA5 ATR platform for subsequent events.
 
An automated artefact detection capability was deployed on the ground agents with some success. The limited field of view prevented some further detections, and was supplemented with additional cameras in subsequent events. Artefact detection was also subsequently deployed on the UAVs.

The UAV performed well, navigating 120\,m in the Army tunnel. Speed was limited by the dust experienced by the platform. A constriction 30\,m inside the Miami tunnel (shown in Figure~\ref{fig:stix_drone_constriction}) prevented further progress. This limitation drove the decision to develop the marsupial launch capability that was deployed in subsequent events.

The operator loading caused by both manual commanding of robots, lack of automation in startup, and communications difficulties, triggered subsequent work on higher levels of autonomy.

\subsection{Tunnel Circuit}
\label{ss:tunnel}
\begin{figure}[!b]
    \centering
    \includegraphics[width=150mm]{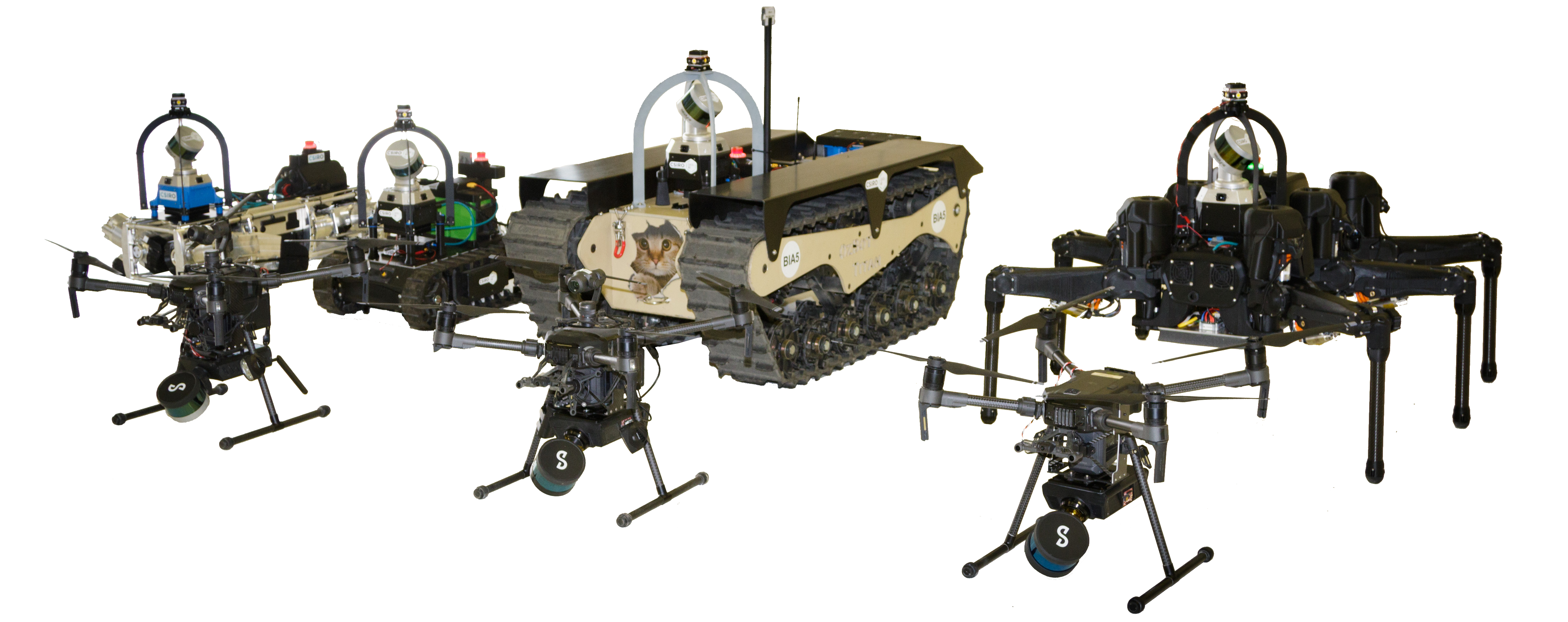}
    \caption{Tunnel Circuit robot fleet consisting of three Emesent Hovermap UAVs (front) which were launched from the starting gate, the Ghost Robotics Vision60 v3.7 (top left), SuperDroid LT2-F, BIA5 ATR (top middle) and the CSIRO Hexapod (top right).}
    \label{fig:tunnel_circuit}
\end{figure}
At the Tunnel Circuit, a diverse team was deployed consisting of the Ghost Robotics Vision60 quadruped, two SuperDroid LT2-Fs, three Emersent Hovermap UAVs, as well as the new BIA5 ATR and the internally developed CSIRO Hexapod as shown in Figure \ref{fig:tunnel_circuit}. The modular perception pack design was particularly valuable in deploying this heterogeneous team, and allowed us to develop independently and accommodate hardware received shortly before the event. A multi-hop mesh communications network was deployed by the lead robot, and used alongside a drop/delay tolerant backend.

\begin{figure}[t]
    \centering
\includegraphics[width=100mm]{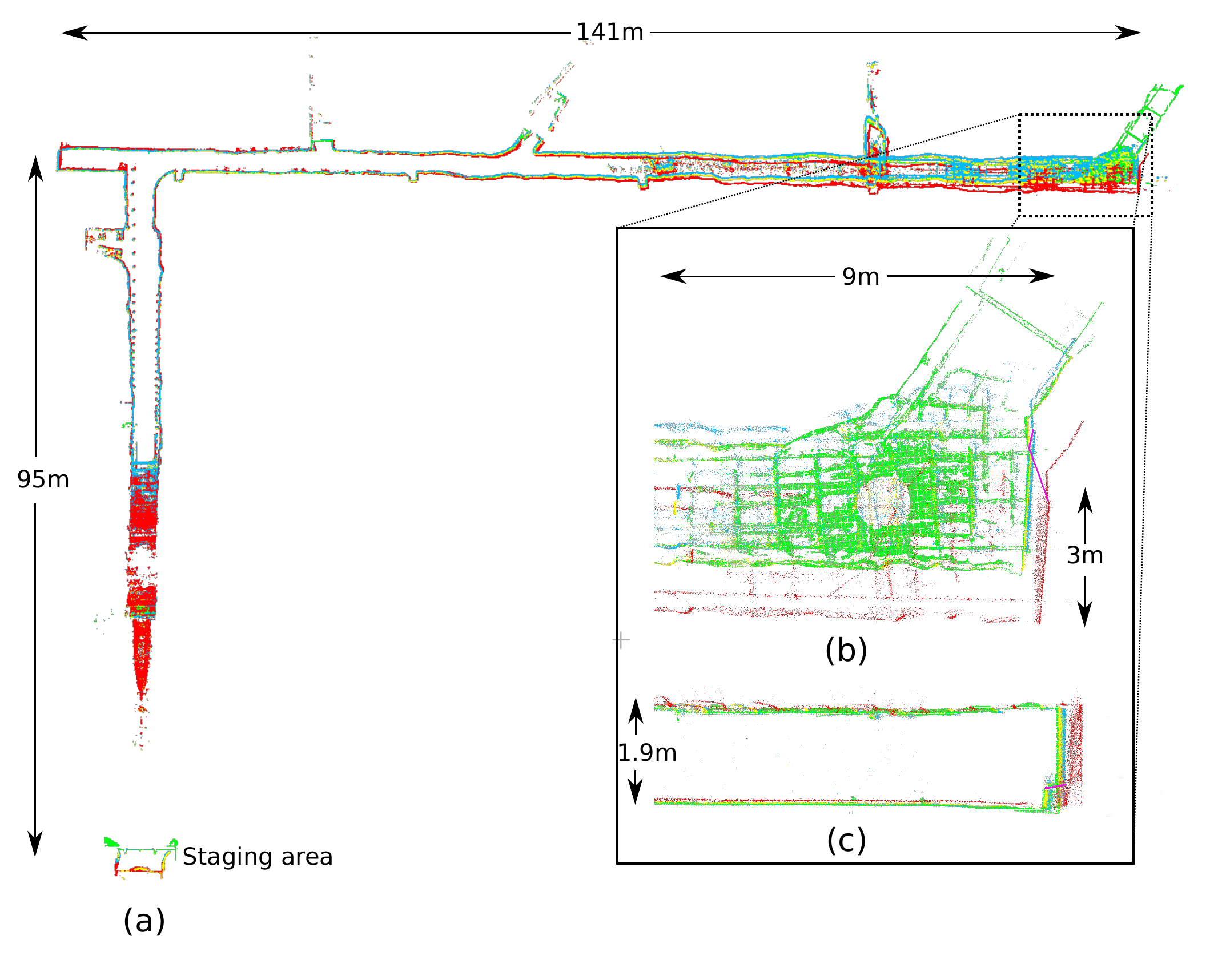}
    \caption{
        The point cloud map from a single UGV for the Safety Research Mine course with a total transit of approximately 215\,m overlaid with the SLAM solutions and reference map. Approximate cross-section measurements are supplied for reference.
        Red: Odometry, Blue: Atlas (real-time), Yellow: Global (offline), Green: Reference point cloud provided by DARPA.
        (a) Top orthographic projection showing the entire map
        (b) Top orthographic projection showing the end of the map
        (c) Side orthographic projection showing the end of the map.
    }
    \label{fig:tunnel_circuit_slam}
\end{figure}

At the Tunnel Circuit, a single agent autonomous exploration capability was fielded, but the operators tended to prefer manual waypoints due to time constraints in training. At this point in development, each agent communicated with the base station, but not with peers, and the base station contained the only multi-agent SLAM solution. 

Large amounts of operator time were consumed in troubleshooting of communications issues, which were exacerbated by robots traversing over and damaging communications nodes. The manual mode of operation was highly dependent on good communications, so this significantly limited performance. Subsequently, development of multi-agent autonomy beyond communication ranges was prioritised.

Artefacts were detected and scored by both the ATR UGV and the Hovermap UAV. This again demonstrated the multi-agent SLAM performance; the gate defining the reference coordinate system was detected on a single agent, and artefacts detected on other agents were transformed using the multi-agent SLAM solution on the base station. The additional FoV provided by the multiple cameras on the revised UGV pack design (Figure~\ref{fig:catpack_fov}) provided improved coverage and performed very well. The fixed alignment of the UAV spotlights to the frontal arc of the UAV limited the gimballed camera's usable FOV. This, combined with the UAV's goto waypoint autonomy and a poorly performing object detection network on the first two runs, led to missed artefacts.

The artefact detector running on the UGVs sent back a large number of false positives. Analysis of these detections showed some of the detections were of the robot itself e.g., communications antennae. These detections were subsequently removed with the introduction of a simple image mask. Other false positives were of artefacts not present in our training data e.g., the anchor plates of rods inserted in to the earth to reinforce the mine. These were mitigated by selecting thresholds for the detector which resulted in a manageable number of detections for the operator.

A total of 11 artefacts out 80 artefacts were correctly detected and located during the tunnel event. Three of the artefacts were detected by both the UGVs and UAVs. Three artefacts were detected by the UGVs only and two by the UAVs. The remaining three artefacts were scored by the operator by inspecting the map. The UGVs and UAVs detected two additional artefacts but the operator failed to report the detections due to the large number of false positives.  

Issues were experienced with local navigation with low-hanging obstacles such as lights. This was found to be due to an inflated vehicle height used in configuration data, which had not been identified as cases had not been encountered in prior testing. The number of agents utilised was more than had been encountered in prior testing, and this revealed a range of issues associated with restarting agents, etc. Many of these issues were the consequence of the aggressive schedule for bringing on new platforms; this led to the decision to focus development on the ATR and SuperDroid UGV platforms for the Urban Circuit.

The accuracy of the map was evaluated by comparing the point clouds produced by one of the UGVs in the Safety Research Mine to the registered, surveyed point cloud provided by DARPA. The analysis was performed by first registering UGV and reference point clouds for the staging area (as shown in Figure~\ref{fig:tunnel_circuit_slam}), and then evaluating the RMS error for corresponding points in the most distant region of the trajectory (215~m from the staging area). The analysis revealed error rates of 0.63\%, 0.22\%, and 0.025\% over distance travelled for the odometry, atlas (real-time), and global (offline) state estimation components of the solution respectively. This method creates an error distribution that accumulates with distance travelled from the deployment area and reflects the same model used for scoring points. This particular course contained several large corridors which proved to be difficult for SLAM as the geometric features were sparse. While it was observed that other teams employed additional sources of external positioning to aid SLAM (such as a totalstation or the secondary gate reference targets in the tunnel), no such aiding was used by the Wildcat SLAM solution. The only accuracy issues observed with our SLAM occurred during hard landings of the UAVs. At this stage, the operator did not have the ability to remove an agent or part of an agent's trajectory from the solution, so this did cause some difficulties.

\subsection{Urban Circuit}
\label{ss:urban}

\begin{figure}[!b]
    \centering
    \includegraphics[width=150mm]{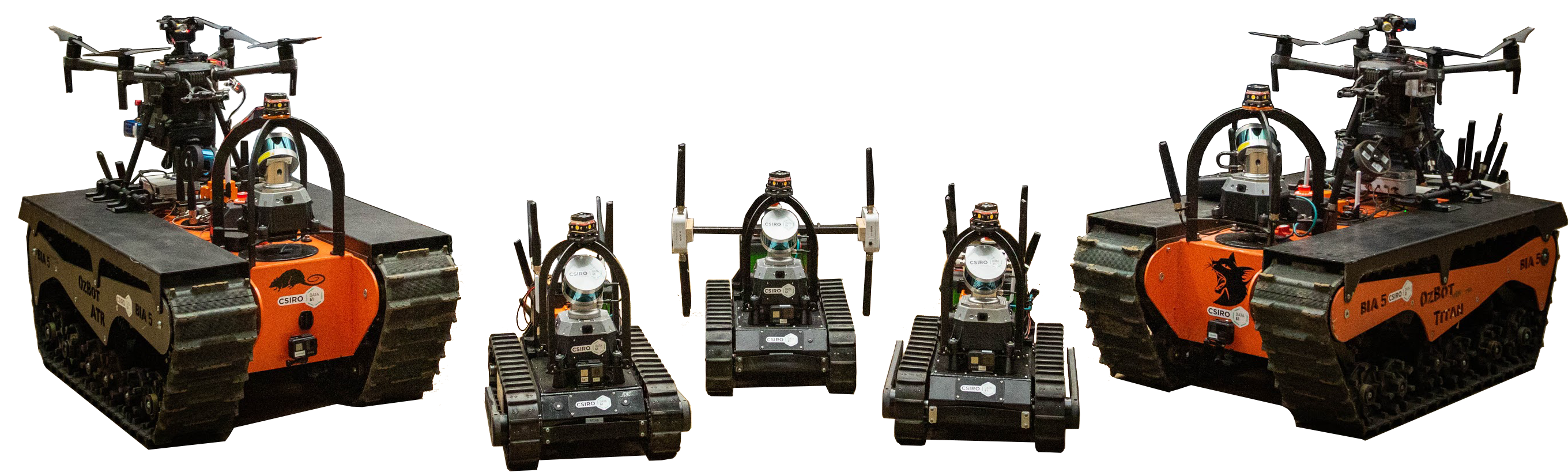}
    \caption{The Urban Circuit robot fleet consisted of two BIA5 ATRs each carrying an Emesent Hovermap UAV, and 3 Superdoid LT2-F platforms.}
    \label{fig:urban_circuit}
\end{figure}

\begin{figure}[!b]
    \centering
    \begin{subfigure}{\columnwidth}
    \centering
    \includegraphics[width=\columnwidth]{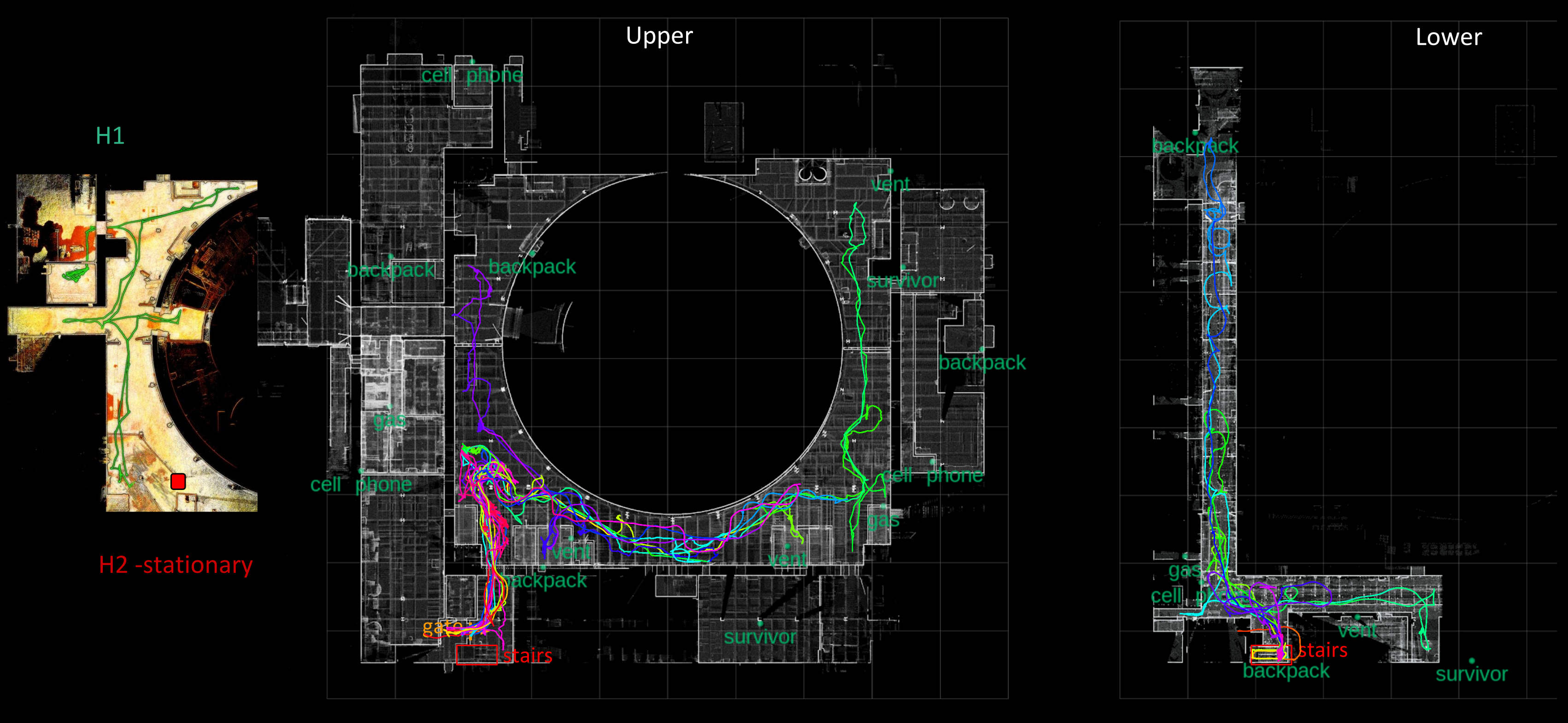}
    \caption{}
    \label{fig:urban_alpha_2}
    \end{subfigure}
    \begin{subfigure}{\columnwidth}
    \centering
    \label{fig:urban_beta_2}
    \includegraphics[width=\columnwidth]{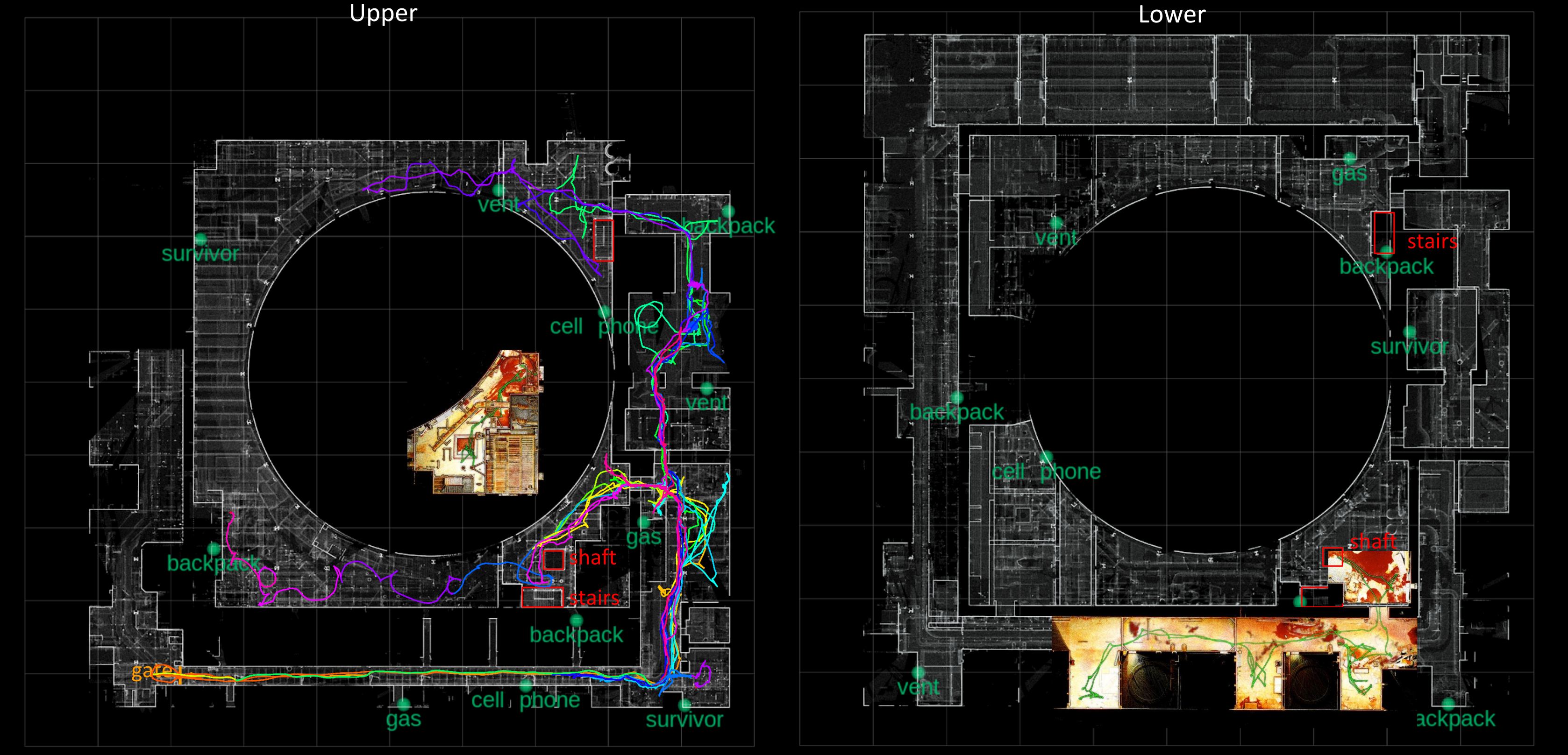}
    \caption{}
    \end{subfigure}
    \caption{Course map and robot trajectories for Urban Circuit event (a) alpha course and (b) beta course. The grey map sections were ground truth data provided by DARPA, with grid cells representing 10m. The coloured map sections, overlaid or adjacent to grey map, were generated by the UAVs. In the beta course, the UAV flew between the upper and lower levels of the course. The multi-colored trajectories overlaid on the grey map are from the multiple UGV systems.}
    \label{fig:urban_trajectories}
\end{figure}

At the Urban Circuit, the supported platforms were consolidated, so that the team consisted of two Hovermap UAVs, two large ATR UGVs and three smaller SuperDroid UGVs, as shown in Figure~\ref{fig:urban_circuit}. Communications switched to the Rajant-based solution (Section \ref{ss:commsys}), peer-to-peer sharing was incorporated (Section \ref{ss:mule}) and multi-agent, decentralised SLAM was activated, so that each agent solved for a global multi-agent map. This enabled navigation on the shared global map, as described in Section \ref{sss:multiagentglobal}. A marsupial launch capability was deployed, in which a UAV was carried by each ATR UGV and launched in the vicinity of a promising region that was unable to be accessed by the ground agents. This proved to be a critical capability as the Beta course had constriction early in the course which would have prevented the UAV flying in from the staging area.

Testing in a realistic site proved invaluable; we performed this in a parking garage under construction, containing many obstacles from construction supplies and debris. This provided representative scale and complexity, particularly with relatively wide open spaces. A key feature that it did not represent was narrow doorways and small rooms, which turned out to be a weakness of our approach. Conversely, Team Explorer trained in an abandoned hospital with many small rooms but without wider open spaces, and also experienced some issues due the environment mismatch.\footnote{As reported in Urban Circuit Technical Interchange Meeting https://youtu.be/ymBrec7HY4A?t=19667.}

As described in Section \ref{ss:exploration} and \cite{williams_2020}, the same frontier logic was deployed in the UGVs and UAVs. In the UGVs, this was set up as explore-sync missions, which were very successful. The total distance traversed by the ground robots over the four runs was 7902\,m. In the Beta course, a ATR UGV traversed 1337\,m and observed seven artefacts, mostly operating on explore-sync. In the Alpha course, a SuperDroid UGV descended a flight of stairs and explored autonomously, covering 782\,m and detecting three artefacts. Narrow doorways were often missed due to the resolution of the traversability analysis used for exploration.

The frontier selection logic performed well when exploring large, complex areas. It was less effective when it came to declaring a complex area as completed, and selecting between distant regions. The interruption caused when robots returned to synchronise data was a significant issue, as robots often did not return to the area being explored. This also caused robots to cluster near the entry where synchronisation was finally obtained, as can be seen in the lower-left corner of the upper floor in Figure~\ref{fig:urban_alpha_2}. This was subsequently addressed through the MRTA approach (Section \ref{ss:taskallocation}), which considered clusters of frontiers rather than individual frontiers, such that tasks were more distributed.

\begin{figure}[t]
    \centering
    \begin{subfigure}{0.49\linewidth}
    \centering
    \includegraphics[height=45mm]{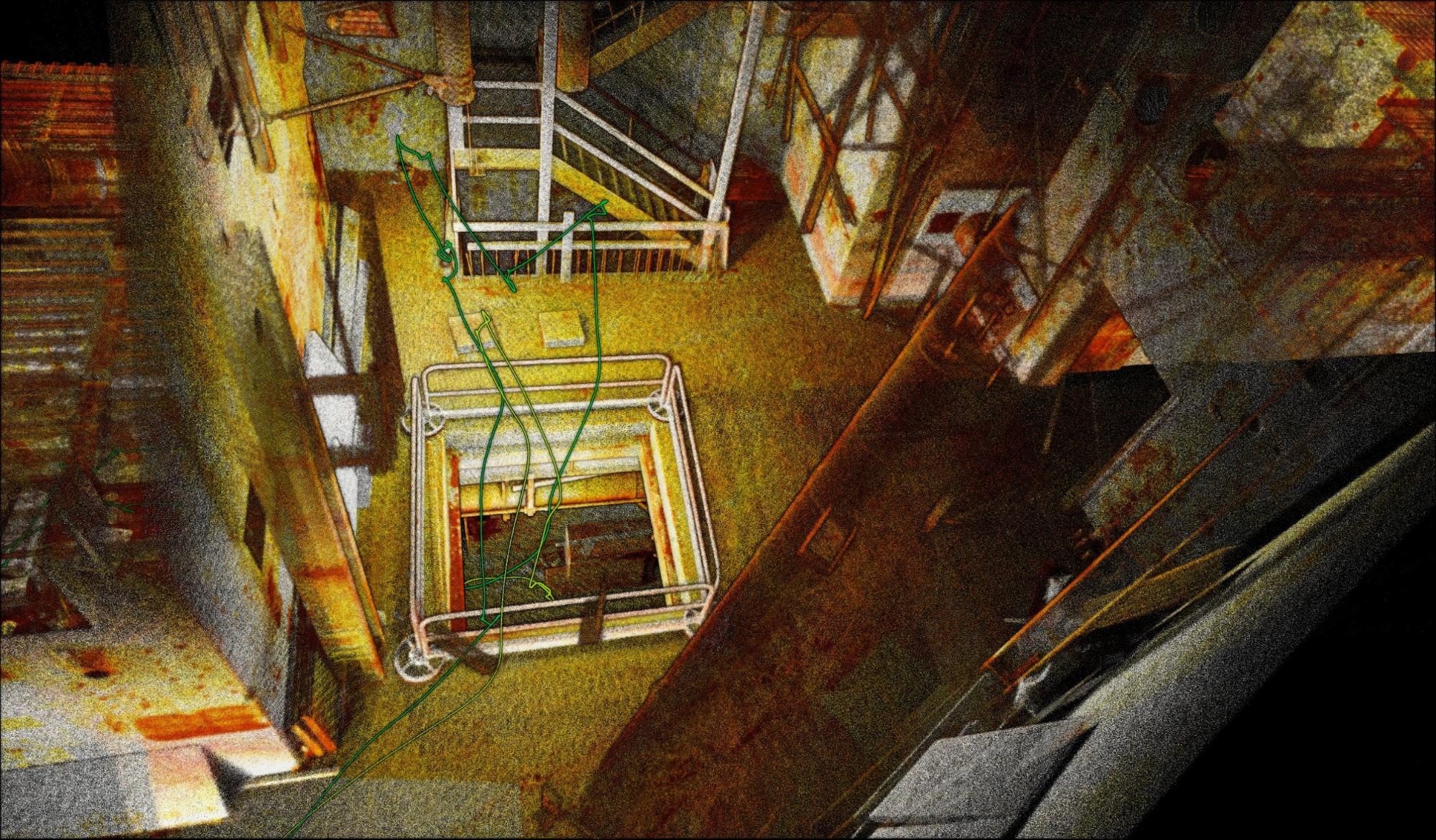}
    \caption{}
    \label{fig:drone_shaft_upper}
    \end{subfigure}
    \begin{subfigure}{0.49\linewidth}
    \centering
    \label{fig:drone_shaft_lower}
    \includegraphics[height=45mm]{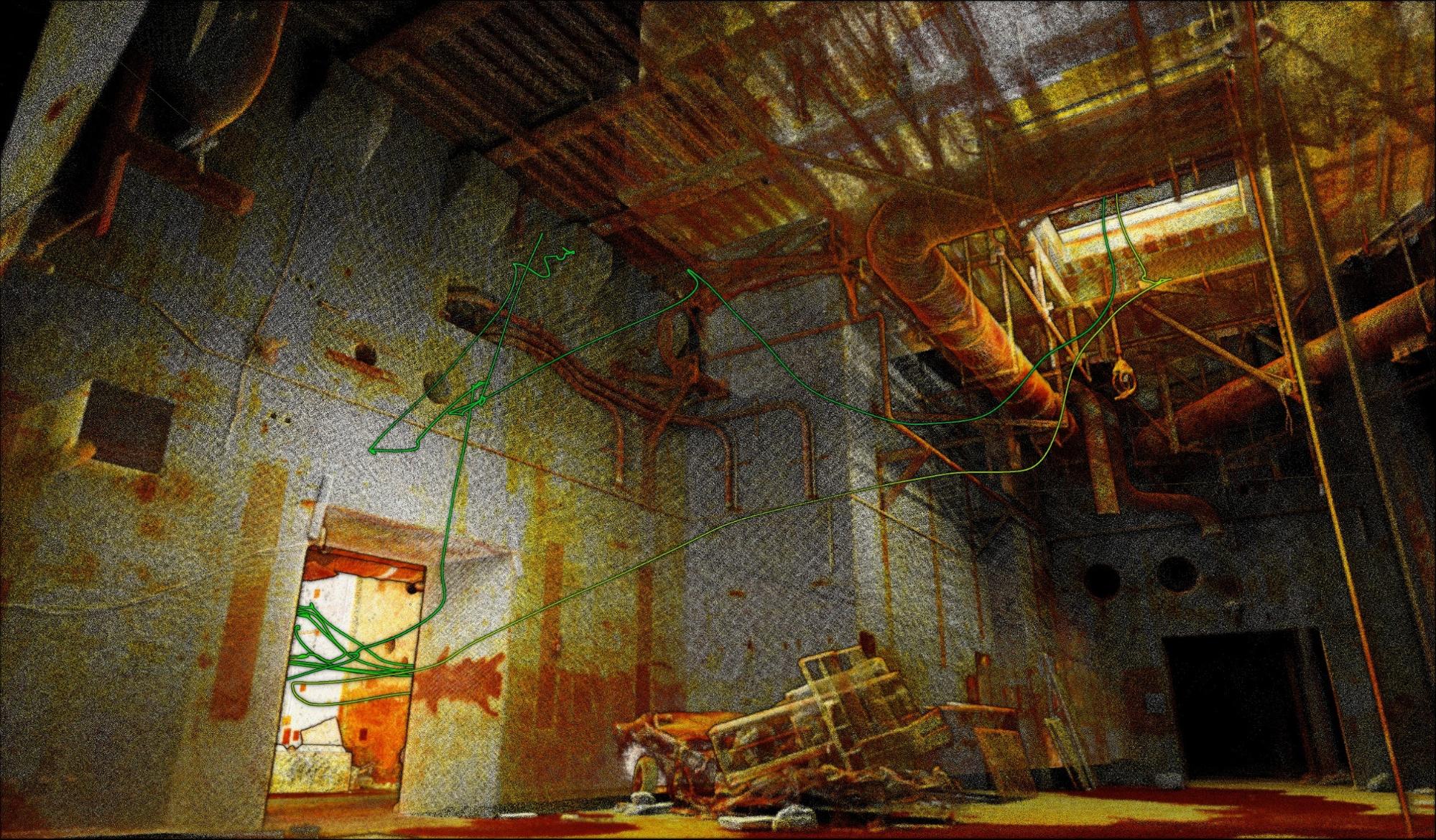}
    \caption{}
    \end{subfigure}
    \caption{Post-processed, intensity-coloured point cloud generated from UAV in mission descending through shaft in the Beta course, with trajectory overlaid. (a) shows upper level where mission commenced, while (b) shows lower level with trajectory descending through shaft and entering adjacent rooms.}
    \label{fig:drone_shaft}
\end{figure}

The course layouts and robot trajectories for the second runs of the alpha and beta courses are shown in Figure~\ref{fig:urban_trajectories}. Launching the UAV in situ was very successful and allowed exploration of areas that could not be covered by the ground agents. In one instance, a vertical shaft was identified by the operator, and the UAV was launched, descended and autonomously traversed the lower level, before returning; the trajectory taken is shown in Figures \ref{fig:urban_trajectories} and \ref{fig:drone_shaft}.

The artefact detector for the UGVs reported 13.3 detections per minute for the first run which overwhelmed the operator. This was mitigated by increasing the thresholds to reduce the number of detections sent to the operator to approximately 5-7 detections per minute. The small number of detections per minute was due to an inefficient user interface for inspecting artefact detections. Improvements to this interface were prioritised after the event to increase the number of detections an operator could process per minute.

The introduction of artefact tracking on the UGVs resulted in a 70\% reduction in the number of artefact detections reported to the operator by a single UGV during the first run.

A total of 15 out of 80 artefacts were correctly detected and localised over the duration of the event. Two artefacts were detected by both the UGVs and the UAVs, 12 were detected by the UGVs and one artefact was identified by the operator upon inspection of the map. Three of the artefacts were cellphones detected using the UGV bluetooth/wifi system. One gas artefact was also correctly identified using the UGV gas sensor. A further six artefacts were possible if the agents had been able to navigate through doorways. Five artefacts were also in the field of view but too far away for detection or otherwise were not captured in camera frames. One artefact was detected by UAV but crashed before sending report.

Agent specific details of perception performance during each run at the Urban Circuit can be found in Table~\ref{tab:agent-course-summary}. We note that the above analysis and Table~\ref{tab:agent-course-summary} is generated by post-processing the recorded data with the perception system configured to generate logging information. The post-processing is not guaranteed to produce identical results to what occurred during an event but we expect the variations to be minor. The configuration of the perception system in the post-processing is identical to the configuration used during the run. This can be seen by comparing the number of detections sent during the first attempt at the Beta course and the remaining Urban course attempts.

The median error for correctly reported artefacts was 0.58\,m; the artefact corresponding to this median happened to be the longest (line of sight) distance from the gate, at 99\,m. Team CSIRO received the award for the most accurate artefact detection, which had an error of 0.22\,m. The median error as a proportion of the line of sight distance from the gate was 0.83\%. These statistics include reports that are inherently less accurate than visual detections, such as bluetooth and gas.

\begin{figure}[!b]
    \centering
    \includegraphics[width=100mm]{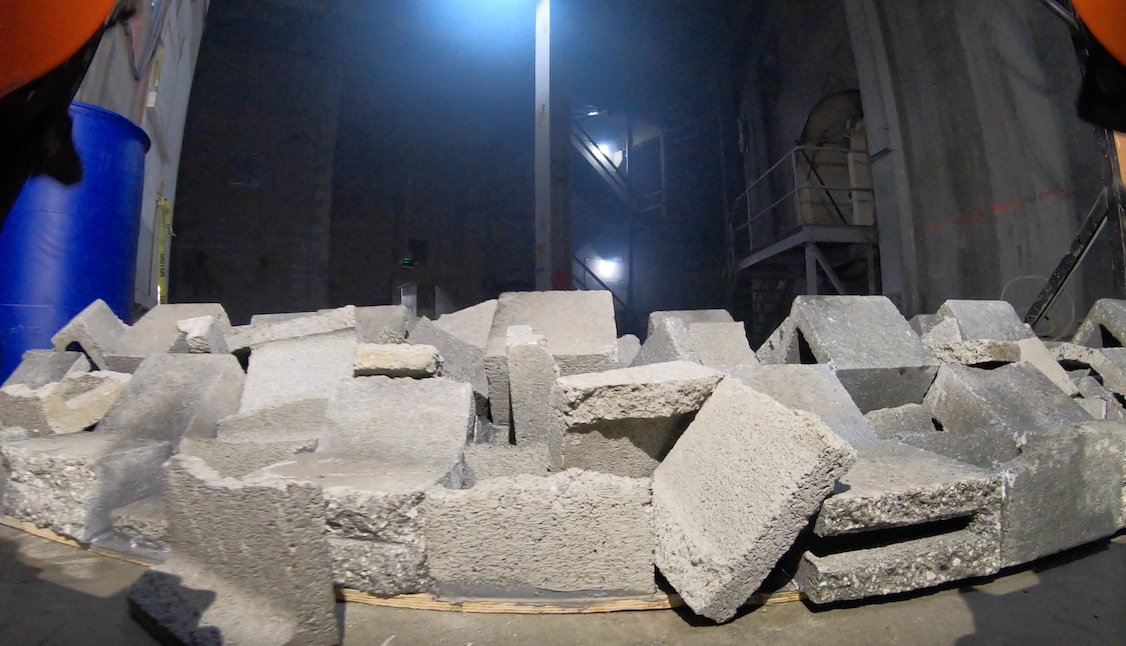}
    \caption{A traversable barrier on the Urban Circuit alpha course, which prevented further access for the UGV systems. The BIA5 Titan robot, from which the picture was taken, is physically capable of traversing the barrier, but the autonomous navigation was far too conservative.}
    \label{fig:urban_obstacle}
\end{figure}

The obstacles shown in Figure~\ref{fig:urban_obstacle} were not autonomously traversed due to assumptions of environment characteristics, and prior test environments. This further motivated the effort on traversability analysis that was made in preparation for the cave event.

\subsection{Cave Circuit}
\label{ss:cave}
\begin{figure}[ht]
    \centering
    \begin{subfigure}{.49\linewidth}
    \includegraphics[width=0.9\columnwidth, height=0.9\columnwidth*9/16]{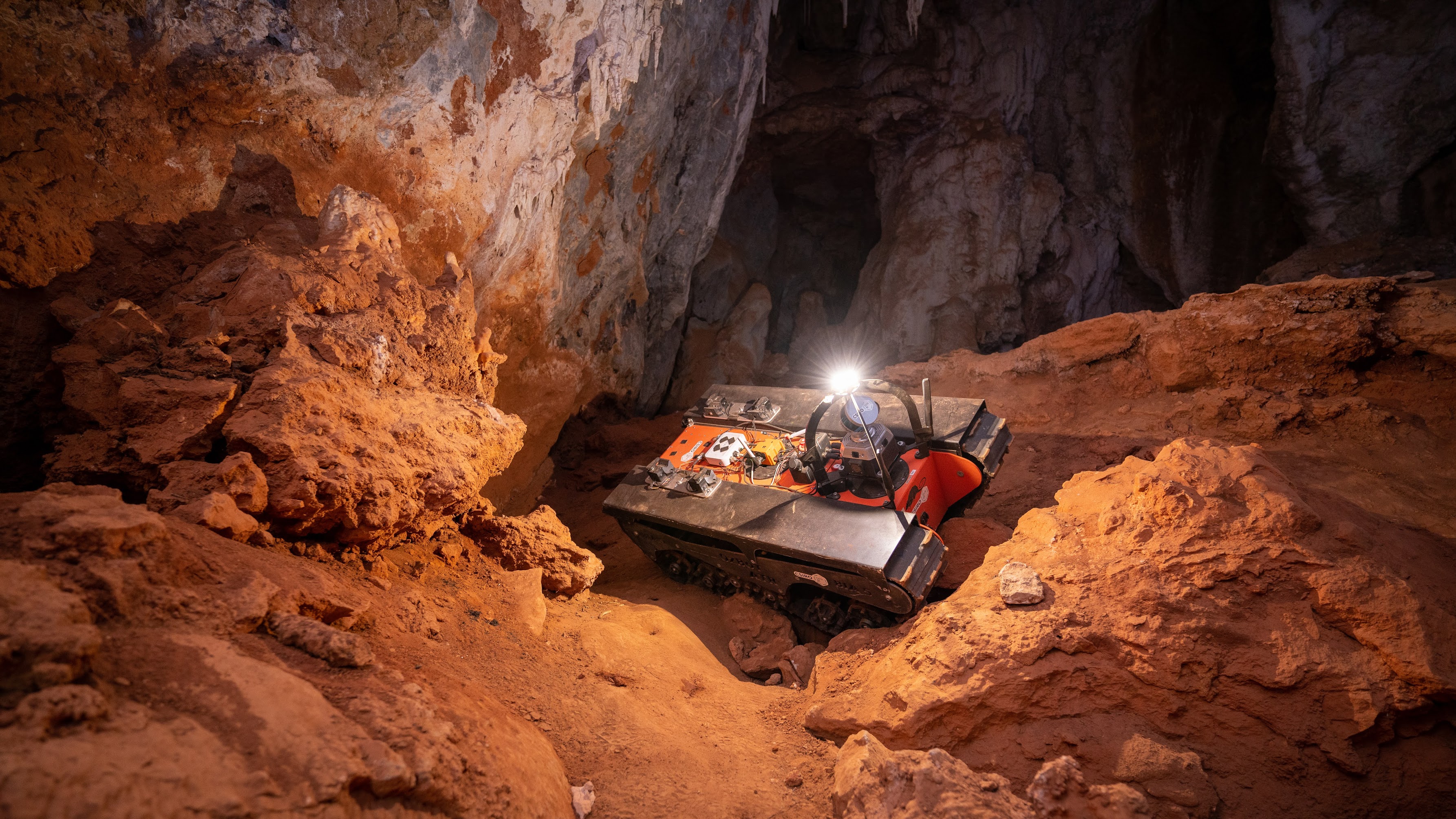}
    \caption{}
    \label{fig:starting_gate}
    \end{subfigure}
    \begin{subfigure}{.49\linewidth}
    \label{fig:titan_drone_pumpkin}
    \includegraphics[width=0.9\columnwidth, height=0.9\columnwidth*9/16]{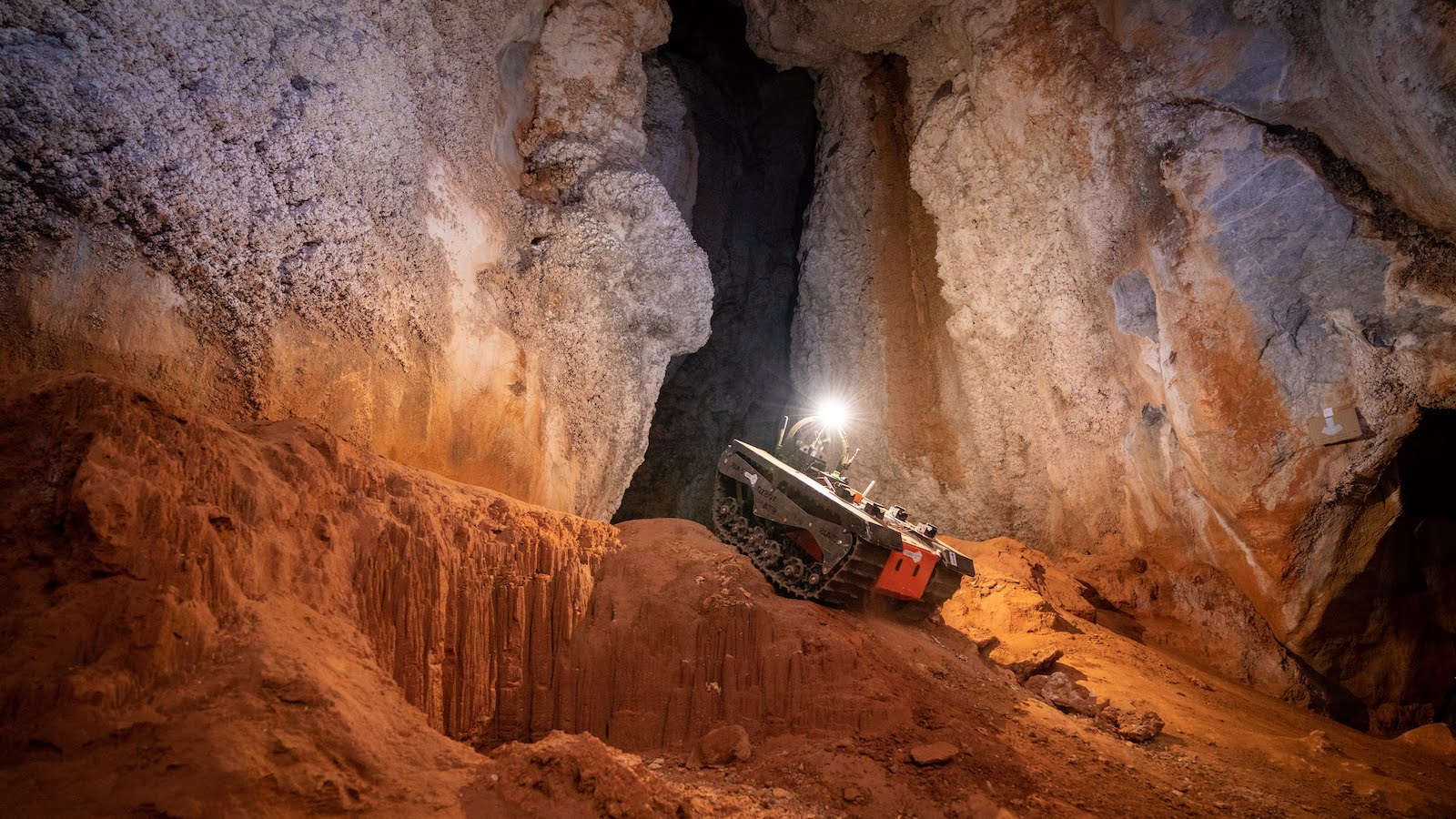}
    \caption{}
    \end{subfigure}
    \begin{subfigure}{.49\linewidth}
    \includegraphics[width=0.9\columnwidth, height=0.9\columnwidth*3/4]{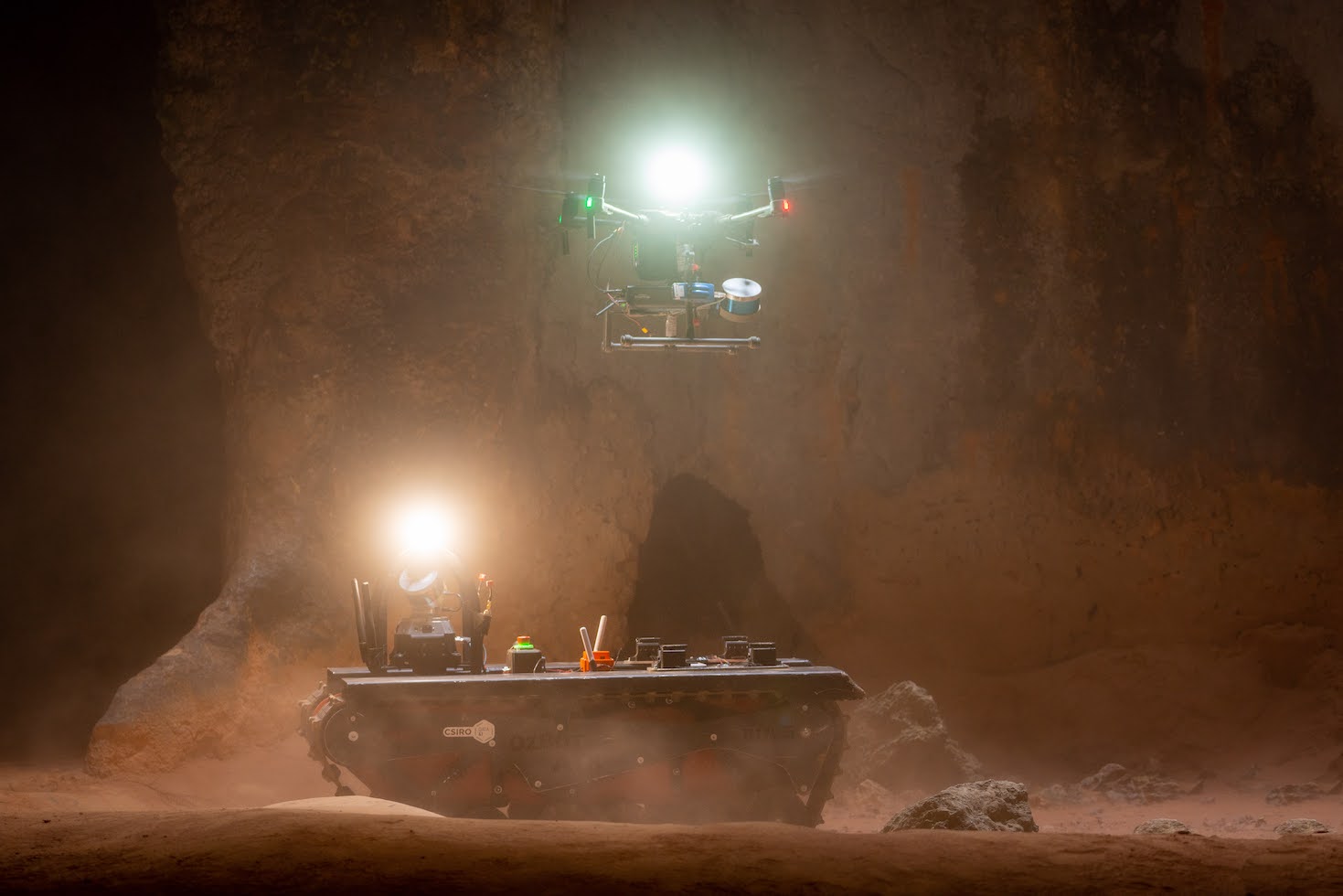}
    \caption{}
    \label{fig:titan_drone_launch}
    \end{subfigure}
    \begin{subfigure}{.49\linewidth}
    \includegraphics[width=0.9\columnwidth, height=0.9\columnwidth*3/4]{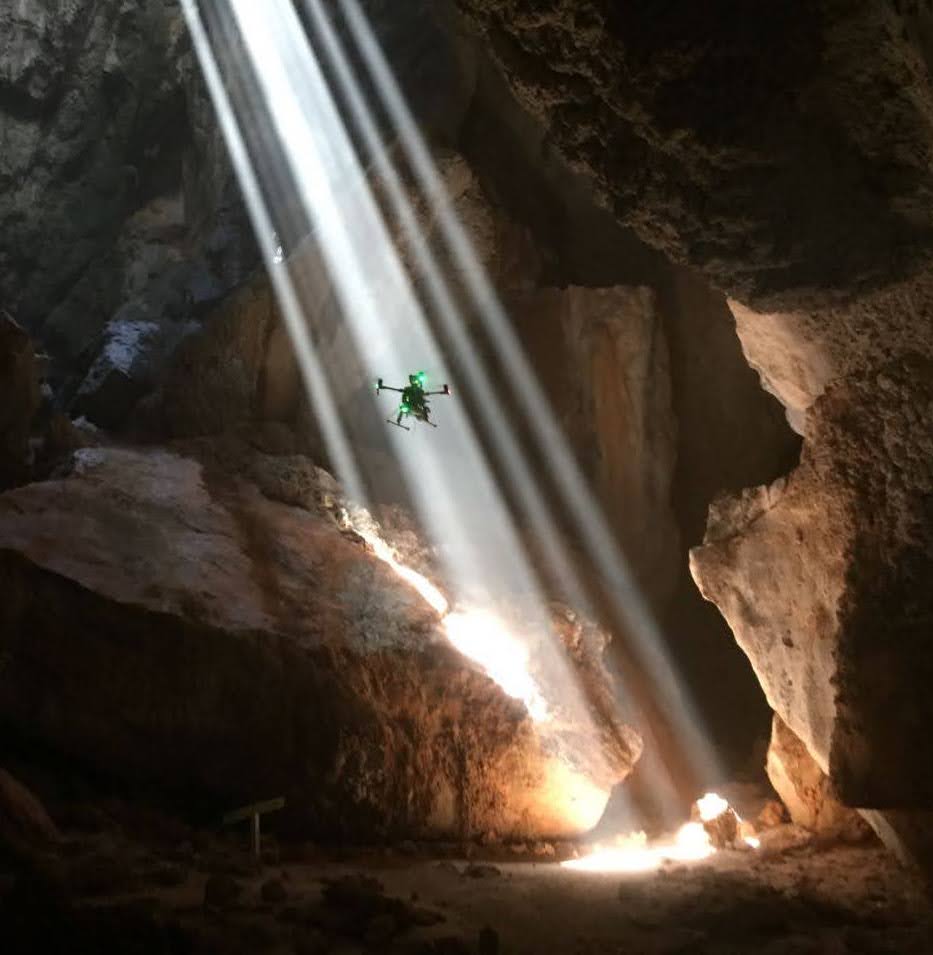}
    \caption{}
    \label{fig:drone_sunlight}
    \end{subfigure}
    \caption{Team CSIRO Data61 deployed a set of heterogeneous ground and air robots, with homogeneous sensing payloads, at the local Cave event. In (a) The BIA5 ATR is navigating down a steep rocky slope after exploring a ledge several metres above the cavern floor, and in (b) is traversing up a slope, avoiding steep drop offs on either side. (c) Shows the Emesent UAV launching from the back of the UGV, and (d) shows the UAV exploring over high rocks in a cavern. The parts of the cave system shown in (a) and (b) were in complete darkness, while the areas in (c) and (d) had openings admitting natural light.}
    \label{fig:cave_event}
\end{figure}

The field trip to the Carpentaria Caves in Chillagoe, Queensland was originally planned as a preparatory step for the DARPA Cave Circuit event. When the latter was cancelled, plans were altered to build a more intense trial, mimicking DARPA test conditions. The robot team employed consisted again of two Hovermaps, carried on two ATR UGVs, along with the newly developed DTR (Section \ref{sec:pumpkin}). Photos of the agents and terrain are shown in Figure~\ref{fig:cave_event}.

The terrain in the cave was more challenging than previously encountered, which caused significant difficulties. The superpixel-based global navigation graph described in Section \ref{sss:multiagentglobal} performed well at identifying terrain that was able to be traversed, and this analysis was also utilised in the frontier exploration. However, the cost of traversing difficult terrain was not sufficiently modelled or penalised, and consequently global paths and frontier selections often preferred short distances that involved unnecessarily traversing challenging terrain. Subsequently, risk-based selection has been made a priority. Changes have also been made to allow local navigation more flexibility in executing the global path.

The local navigation approach described in Section \ref{ss:ugvnavigation} and \cite{hines_virtual_2021} performed exceptionally well in reaching difficult waypoints and avoiding negative obstacles. Vehicles rolled several times, but each of these were traced slip, subsidence, and operator teleoperation with insufficient understanding of the environment. Minimal damage was sustained, and in some instances vehicles continued after a complete roll. It was the overwhelming impression that the autonomous local navigation was more reliable in the challenging terrain than the operator teleoperating with the aid of a local cost map and cameras.

The insufficient modelling of risk in waypoint selection led to instances where the agent became stuck in a location of challenging terrain. These instances were either resolved by commanding of a different waypoint which allowed successful navigation, or operator intervention. Repeated failed navigation attempts led to tasks which were temporarily blocked from execution. This is being addressed by development of an explicit ``unstuck'' behaviour, and improved awareness of the navigation state where task blacklisting is performed.

The UAV exhibited excellent exploration performance in the cave environment, allowing coverage of more terrain than was possible using the ground agents. The mode of operation was altered to utilise the UAV as a \textit{scout}, rather than a focused tool to analysis identified inaccessible locations.

\begin{figure}[!t]
    \centering
    \includegraphics[width=\columnwidth]{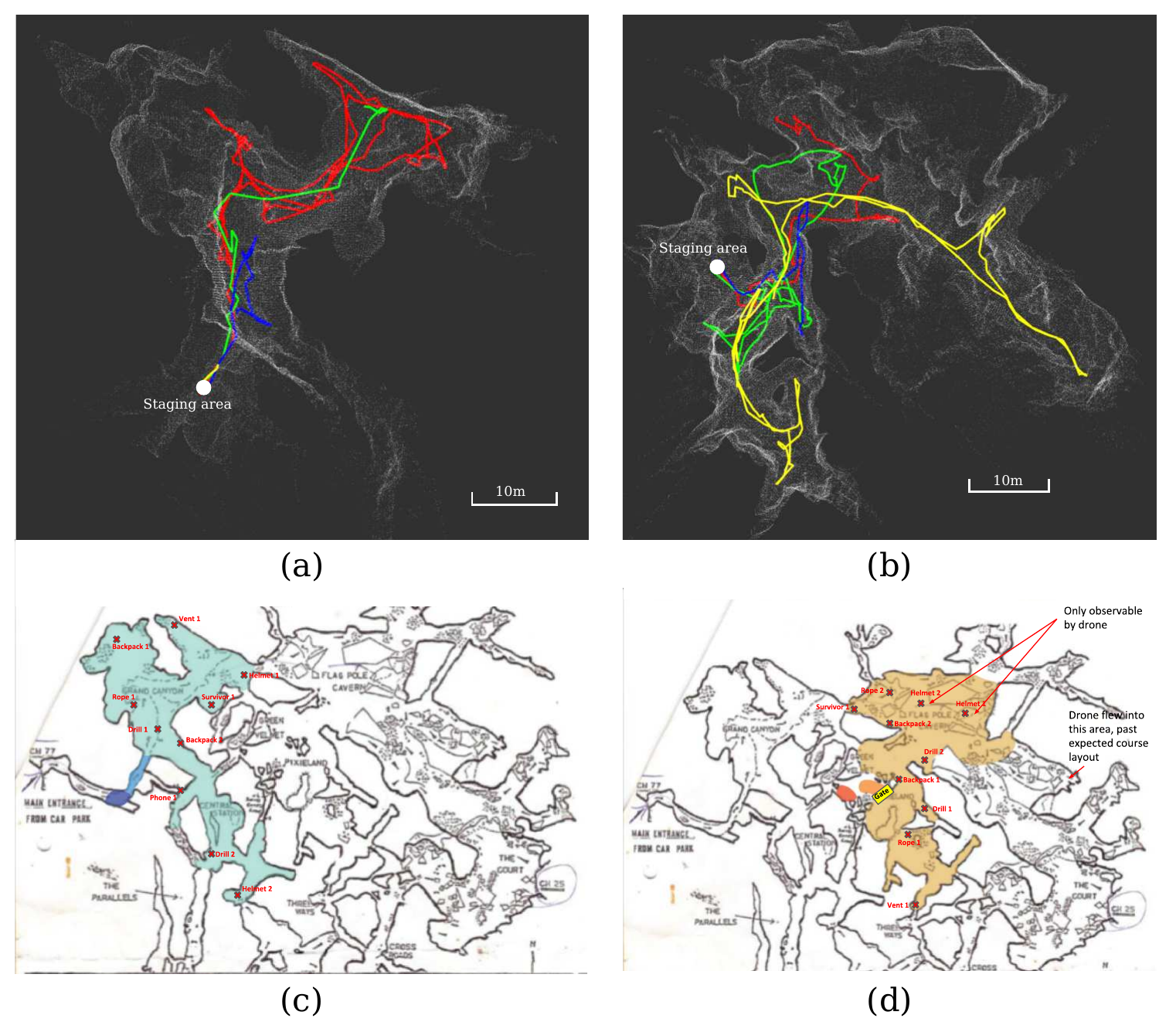}
    \caption{Maps of (a) alpha and (b) beta courses in cave testing, and robot trajectories. Green and red show large ATR UGVs, blue shows small DTR UGV, and yellow shows Hovermap UAV. (c) and (d) show maps of the larger respective areas with the intended course areas shaded, and artefact positions marked. In alpha course, the segment to the bottom right was not explored, due to constrained access through a steep narrow and rocky path. The beta course was almost entirely explored by the combination of UAV and UGVs, but the large cavern at the top was not explored completely, instead the UAV flew through a narrower opening and beyond the intended course area (right section of the yellow trajectory in (b)).}
    \label{fig:cave_trajectories}
\end{figure}

The courses and trajectories are illustrated in Figure~\ref{fig:cave_trajectories}. In alpha course, the red agent traversed 367\,m, and four out of ten artefacts were detected (the UAV did not operate in this run). In beta course, the green agent traversed 231\,m, while the UAV (yellow) traversed 400\,m. Again, four out of ten artefacts were detected.

MRTA (described in Section \ref{ss:taskallocation}) accounted for the majority of missions. Agent separation generally improved, due to improvements in inter-agent coordination, and representation of tasks at a higher level than individual frontiers. The courses were limited to parts of the cave that could be safely accessed by humans, and from which robots could be retrieved. Because of the distribution of impassible segments in the cave, this led to a smaller scale course in comparison to Tunnel and Urban circuit events. Consequently, good communications links were maintained throughout; the challenge of the test was extreme traversability rather than scale.

A total of 16 out of 44 artefacts were correctly detected and located across the four evaluations. Of these, 14 were detected by the UGVs and the remaining two were identified by the operator upon inspection of the map. Three additional artefacts were in field of view but were to far away to be detected. One artefact was missed because the detector thresholds were too high.

Agent specific details of perception performance during each run at the Cave Circuit can be found in Table~\ref{tab:agent-course-summary}. Again, we note that the above analysis of the perception system performance and the data present in Table~\ref{tab:agent-course-summary} was generated by post-processing the recorded data with the perception system configured to generate logging information.

\begin{table}[]
    \centering
    \begin{tabular}{|l|l|c|c|c|c|c|c|c|}
         \hline
         Course & Agent & \#Raw & \#Sent & \shortstack{\# Missed \\ (Perception)} & \shortstack{\# Missed \\ (Comms) } & \shortstack{ Distance \\ (m) } & \shortstack{Duration \\ (mins) } \\
         \hline\hline
Urban Beta 1            & Atlas         &  3468   &  296 &   &   &  151 &   54.8 \\
Urban Beta 1            & Kitten        &  3012   &  820 &   & 2 &  670 &   60.0 \\
Urban Beta 1            & Winslow       &  2058   &  422 &   &   &  509 &   36.7 \\
Urban Beta 1            & Hovermap      &    15   &   15 &   &   &  287 &   11.0 \\     
\hline
Urban Alpha 1           & Atlas         &   554   &   68 &   &   &  379 &   23.1 \\
Urban Alpha 1           & Beachhead     &   262   &   55 & 1 &   &  414 &   54.0 \\
Urban Alpha 1           & Kitten        &   173   &   36 &   &   &  118 &   57.6 \\
Urban Alpha 1           & Rat           &   783   &  182 & 3 & 1 &  907 &   60.0 \\
Urban Alpha 1           & Winslow       &   494   &   32 &   &   &  104 &   32.6 \\
Urban Alpha 1           & Hovermap      &   384   &   55 &   & 1 &  174 &    4.3 \\
\hline
Urban Beta 2            & Atlas         &   932   &  191 & 1 &   &  490 &   40.4 \\
Urban Beta 2            & Rat           &   981   &  303 & 1 &   & 1337 &   60.0 \\
Urban Beta 2            & Winslow       &  4571   &   56 &   &   &  131 &   24.1 \\
Urban Beta 2            & Hovermap      &     55  &   55 &   &   &  367 &   12.0 \\
\hline
Urban Alpha 2           & Atlas         &  2890   &  167 &   &   &  782 &   51.0 \\
Urban Alpha 2           & Beachhead     &   996   &  115 &   &   &  288 &   23.0 \\
Urban Alpha 2           & Kitten        &   696   &  161 &   &   &  610 &   33.9 \\
Urban Alpha 2           & Rat           &  1179   &  139 &   &   &  514 &   59.4 \\
Urban Alpha 2           & Winslow       &  1972   &  121 &   &   &  498 &   28.1 \\
Urban Alpha 2           & Hovermap1     &   135   &  112 &   &   &    0 &   12.0 \\
Urban Alpha 2           & Hovermap2     &    22   &   19 & 2 &   &  289 &   11.2  \\
\hline\hline
Cave Alpha 1            & Rat           &   162   &   17 &   &   &  179 &   27.4 \\
Cave Alpha 1            & Kitten        &   357   &   35 &   &   &  205 &   23.6 \\
Cave Alpha 1            & Pumpkin       &   103   &   22 &   &   &  100 &   26.2 \\
\hline
Cave Alpha 2            & Kitten        &   226   &   27 & 1 &   &  367 &   37.6 \\
Cave Alpha 2            & Pumpkin       &    64   &    7 & 1 &   &   94 &   25.6 \\
\hline
Cave Beta 1             & Rat           &   127   &   20 &   &   &  231 &   40.0 \\
Cave Beta 1             & Kitten        &   403   &   19 & 1 &   &  195 &   36.0 \\
Cave Beta 1             & Pumpkin       &  1336   &    9 &   &   &   82 &   31.6 \\
Cave Beta 1             & Hovermap      &    17   &    4 &   &   &  262 &   10.6 \\
\hline
Cave Beta 2             & Rat           &  1280   &    9 &   &   &   87 &   48.4 \\
Cave Beta 2             & Kitten        &   261   &   18 & 2 &   &  206 &   44.9 \\
Cave Beta 2             & Pumpkin       &    57   &   12 &   &   &  136 &   38.2 \\
Cave Beta 2             & Hovermap      &   179   &    5 &   &   &  281 &   12.2 \\
         \hline
    \end{tabular}
    \caption{A summary of agent, communications and perception performance for the Urban and Cave events. The ``\# Raw'' column indicates the number of artefact detections made by an agent during a run. The number of unique tracked artefacts sent to the base station is shown in the ``\# Sent'' column. The number of competition points missed due to perception failure and communication issues are shown in the two ``\# Missed'' columns. The last two columns show the distance traversed in the duration the agent was active.}
    \label{tab:agent-course-summary}
\end{table}

\section{Conclusions}
\label{sec:conclusions}

The CSIRO Data61 Robot fleet was deployed in three challenging terrain types for the DARPA SubT Phase I and Phase II circuit events. Through these competitions, the team focused on increasing levels of autonomy and coordination rather than creating better communication systems, and exploited the heterogeneity of simpler UGVs and the UAVs over fielding more capable, but more complex, individual agents with legs. Creating a modular perception system based around a spinning lidar for the UGVs, complemented the Emesent Hovermap system on the UAVs, and allowed rapid experimentation and adoption of UGV platforms. The homogeneous sensing on each agent and associated Wildcat SLAM, allowed for meaningful sharing of information between agents through a global map computed by each agent. This map provided the basis for creating higher levels of autonomy including sharing navigation and artefact information between agents, as well as autonomous task allocation. Consistently, the most effective robotic agents at each event (in terms of scoring points) where those commanded to use the highest levels of autonomy available at the time. The most important uses of fleet heterogeneity was to increase course coverage: The UGVs could carry the UAVs and breach narrow parts of the course, in turn enabling the UAVs to demonstrate rapid access to shafts or terrain not traversable by the UGV systems. Smaller UGVs also provided the ability to access the narrowest parts of courses (such as stairwells) not traversable by either the larger UGVs or the UAVs and also were used as extra communication relays. Harnessing this heterogeneity was achieved through the shared global map, with agents gaining coordination by providing situational awareness to the operator, and by the the multi-robot task allocation system. 

Follow-on work which is currently in progress includes:
\begin{itemize}
\item Incorporating terrain risk into frontier selection and global navigation in order to prioritise exploration of easily accessible areas before challenging terrain
\item Improving the coordination of exploration between agents, rewards of exploration tasks, and operator interaction to prioritise exploration
\item Introducing additional behaviours to autonomously pass through narrow openings, to utilise when agents are stuck, and to allow vehicles to navigate cooperatively in close proximity
\item Introducing a live colourised point cloud (based on \cite{VecCox18}) to enhance operator situational awareness
\item Improving the perception capability of the UAV by improving aircraft and gimbal motion, or introducing multiple cameras
\item Developing \textit{smart nodes}, packaging a computational platform running the Mule peer-to-peer software with the Rajant drop node, allowing data from a passing agent to be stored and transmitted to another agent which becomes close at a later time
\item Generally improving platform robustness and stability; e.g., avoiding mounting of the drop nodes which increase the platform footprint and reduce the maximum slope
\end{itemize}

The focus on autonomy, coordination and platform heterogeneity will form the basis for deploying the fleet in even larger scale environments for the SubT final competition in September 2021.

\subsubsection*{Acknowledgements}
The authors would like to thank those in our institutions who either directly, or through prior efforts, allowed us to build upon their knowledge. Many CSIRO QCAT site staff were critical in enabling us to build and access testing facilities locally. We deeply appreciate the access to additional urban sites through Hutchinson Builders, and the cave system by the Chillagoe Caving Club Australia. Finally we appreciate the open dialogue and helpfulness from the other other teams during competition, especially Team PLUTO who directly guided us to using the Rajant Radios.    

\bibliographystyle{apalike}
\bibliography{references}

\begin{thebibliography}{}

\bibitem[Balakirsky et~al., 2007]{BalCar07}
Balakirsky, S., Carpin, S., Kleiner, A., Lewis, M., Visser, A., Wang, J., and
  Ziparo, V.~A. (2007).
\newblock Towards heterogeneous robot teams for disaster mitigation: Results
  and performance metrics from robocup rescue.
\newblock {\em Journal of Field Robotics}, 24(11‐12):943--967.

\bibitem[Bertsekas, 1990]{bertsekas1990auction}
Bertsekas, D.~P. (1990).
\newblock The auction algorithm for assignment and other network flow problems:
  A tutorial.
\newblock {\em Interfaces}, 20(4):133--149.

\bibitem[{BIA5}, 2020]{bia5}
{BIA5} (2020).
\newblock {Robotic Solutions}.
\newblock https://bia5.com/robotics-solutions/.

\bibitem[Bjelonic et~al., 2018]{bjelonic_weaver:_2018}
Bjelonic, M., Kottege, N., Homberger, T., Borges, P., Beckerle, P., and Chli,
  M. (2018).
\newblock Weaver: {Hexapod} robot for autonomous navigation on unstructured
  terrain.
\newblock {\em Journal of Field Robotics}, 35(7):1063--1079.

\bibitem[{Bosse} et~al., 2003]{BosNew03}
{Bosse}, M., {Newman}, P., {Leonard}, J., {Soika}, M., {Feiten}, W., and
  {Teller}, S. (2003).
\newblock An atlas framework for scalable mapping.
\newblock In {\em 2003 IEEE International Conference on Robotics and Automation
  (Cat. No.03CH37422)}, volume~2, pages 1899--1906 vol.2.

\bibitem[{Bosse} et~al., 2012]{BosZlo12}
{Bosse}, M., {Zlot}, R., and {Flick}, P. (2012).
\newblock Zebedee: Design of a spring-mounted 3-d range sensor with application
  to mobile mapping.
\newblock {\em IEEE Transactions on Robotics}, 28(5):1104--1119.

\bibitem[Bouman et~al., 2020]{BouFad20}
Bouman, A., Ginting, M.~F., Alatur, N., Palieri, M., Fan, D.~D., Touma, T.,
  Pailevanian, T., Kim, S.-K., Otsu, K., Burdick, J., and Agha-Mohammadi, A.
  (2020).
\newblock Autonomous {S}pot: Long-range autonomous exploration of extreme
  environments with legged locomotion.
\newblock In {\em Proc IEEE/RSJ International Conference on Intelligent
  Robotics and Systems (IROS)}, pages 1057--1064.

\bibitem[Butzke et~al., 2012]{ButDan12}
Butzke, J., Daniilidis, K., Kushleyev, A., Lee, D.~D., Likhachev, M., Phillips,
  C., and Phillips, M. (2012).
\newblock The university of pennsylvania magic 2010 multi-robot unmanned
  vehicle system.
\newblock {\em Journal of Field Robotics}, 29(5):745--761.

\bibitem[Choi et~al., 2009]{choi2009consensus}
Choi, H.-L., Brunet, L., and How, J.~P. (2009).
\newblock Consensus-based decentralized auctions for robust task allocation.
\newblock {\em IEEE transactions on robotics}, 25(4):912--926.

\bibitem[{CSIRO Robotics and Autonomous Systems Group}, 2020]{ohm}
{CSIRO Robotics and Autonomous Systems Group} (2020).
\newblock {Occupancy Homogeneous Map}.
\newblock https://github.com/csiro-robotics/ohm.

\bibitem[Dang et~al., 2020]{DanTra20}
Dang, T., Tranzatto, M., Khattak, S., Mascarich, F., Alexis, K., and Hutter, M.
  (2020).
\newblock Graph-based subterranean exploration path planning using aerial and
  legged robots.
\newblock {\em Journal of Field Robotics}, 37(8):1363--1388.

\bibitem[DARPA, 2020]{darpasubt}
DARPA (2020).
\newblock {DARPA Subterranean Challenge}.
\newblock https://www.subtchallenge.com/.

\bibitem[Deng et~al., 2009]{ImageNet2009}
Deng, J., Dong, W., Socher, R., Li, L.-J., Li, K., and Fei-Fei, L. (2009).
\newblock {ImageNet: A Large-Scale Hierarchical Image Database}.
\newblock In {\em CVPR09}.

\bibitem[Elfes et~al., 2017]{Elfes2017}
Elfes, A., Steindl, R., Talbot, F., Kendoul, F., Sikka, P., Lowe, T., Kottege,
  N., Bjelonic, M., Dungavell, R., Bandyopadhyay, T., Hoerger, M., Tam, B., and
  Rytz, D. (2017).
\newblock The multilegged autonomous explorer {(MAX)}.
\newblock In {\em IEEE International Conference on Robotics and Automation},
  pages 1050--1057.

\bibitem[{Emesent}, 2020]{emesent}
{Emesent} (2020).
\newblock {Autonomy Level 2}.
\newblock https://www.emesent.io/autonomy-level-2/.

\bibitem[Gerkey and Matari{\'c}, 2004]{gerkey2004formal}
Gerkey, B.~P. and Matari{\'c}, M.~J. (2004).
\newblock A formal analysis and taxonomy of task allocation in multi-robot
  systems.
\newblock {\em The International journal of robotics research}, 23(9):939--954.

\bibitem[{Ghost Robotics}, 2020]{ghostrobotics}
{Ghost Robotics} (2020).
\newblock {Vision 60 quadruped robot}.
\newblock https://www.ghostrobotics.io/robots.

\bibitem[Gregory et~al., 2016]{gregory2016application}
Gregory, J., Fink, J., Stump, E., Twigg, J., Rogers, J., Baran, D., Fung, N.,
  and Young, S. (2016).
\newblock Application of multi-robot systems to disaster-relief scenarios with
  limited communication.
\newblock In {\em Field and Service Robotics}, pages 639--653. Springer.

\bibitem[Hines et~al., 2021]{hines_virtual_2021}
Hines, T., Stepanas, K., Talbot, F., Sa, I., Lewis, J., Hernandez, E., Kottege,
  N., and Hudson, N. (2021).
\newblock Virtual {Surfaces} and {Attitude} {Aware} {Planning} and {Behaviours}
  for {Negative} {Obstacle} {Navigation}.
\newblock {\em IEEE Robotics and Automation Letters}, 6(2):4048--4055.

\bibitem[Howard et~al., 2006]{HowPar06}
Howard, A., Parker, L.~E., and Sukhatme, G.~S. (2006).
\newblock Experiments with a large heterogeneous mobile robot team:
  Exploration, mapping, deployment and detection.
\newblock {\em The International Journal of Robotics Research},
  25(5--6):431--447.

\bibitem[Hsieh et~al., 2007]{HsiCow07}
Hsieh, M.~A., Cowley, A., Keller, J.~F., Chaimowicz, L., Grocholsky, B., Kumar,
  V., Taylor, C.~J., Endo, Y., Arkin, R.~C., Jung, B., Wolf, D.~F., Sukhatme,
  G.~S., and MacKenzie, D.~C. (2007).
\newblock Adaptive teams of autonomous aerial and ground robots for situational
  awareness.
\newblock {\em Journal of Field Robotics}, 24(11-‐12):991--1014.

\bibitem[Huang et~al., 2016]{Huang16}
Huang, J., Rathod, V., Sun, C., Zhu, M., Korattikara, A., Fathi, A., Fischer,
  I., Wojna, Z., Song, Y., Guadarrama, S., and Murphy, K. (2016).
\newblock Speed/accuracy trade-offs for modern convolutional object detectors.
\newblock {\em CoRR}, abs/1611.10012.

\bibitem[Huang et~al., 2019]{HuaLu19}
Huang, Y.-W., Lu, C.-L., Chen, K.-L., Ser, P.-S., Huang, J.-T., Shen, Y.-C.,
  Chen, P.-W., Chang, P.-K., Lee, S.-C., and Wang, H.-C. (2019).
\newblock Duckiefloat: a collision-tolerant resource-constrained blimp for
  long-term autonomy in subterranean environments.

\bibitem[{Katz} and {Tal}, 2015]{KatTal15}
{Katz}, S. and {Tal}, A. (2015).
\newblock On the visibility of point clouds.
\newblock In {\em {ICCV}}, pages 1350--1358.

\bibitem[Kim et~al., 2021]{KimBou21}
Kim, S.-K., Bouman, A., Salhotra, G., Fan, D.~D., Otsu, K., Burdick, J., and
  akbar Agha-mohammadi, A. (2021).
\newblock {PLGRIM}: Hierarchical value learning for large-scale exploration in
  unknown environments.

\bibitem[Kleiner and Ziparo, 2006]{KleZip06}
Kleiner, A. and Ziparo, V. (2006).
\newblock {R}obo{C}up{R}escue - simulation league team {R}escue{R}obots
  {F}reiburg ({G}ermany).

\bibitem[Lin et~al., 2014]{MSCOCO2014}
Lin, T.-Y., Maire, M., Belongie, S., Hays, J., Perona, P., Ramanan, D.,
  Doll{\'a}r, P., and Zitnick, C.~L. (2014).
\newblock Microsoft coco: Common objects in context.
\newblock In Fleet, D., Pajdla, T., Schiele, B., and Tuytelaars, T., editors,
  {\em Computer Vision -- ECCV 2014}, pages 740--755, Cham. Springer
  International Publishing.

\bibitem[Liu and Nejat, 2013]{LiuNej13}
Liu, Y. and Nejat, G. (2013).
\newblock Robotic urban search and rescue: A survey from the control
  perspective.
\newblock {\em Journal of Intelligent and Robotic Systems}, 72:147--165.

\bibitem[{Mangelson} et~al., 2018]{ManDom18}
{Mangelson}, J.~G., {Dominic}, D., {Eustice}, R.~M., and {Vasudevan}, R.
  (2018).
\newblock Pairwise consistent measurement set maximization for robust
  multi-robot map merging.
\newblock In {\em 2018 IEEE International Conference on Robotics and Automation
  (ICRA)}, pages 2916--2923.

\bibitem[Miller et~al., 2020]{Miller20}
Miller, I., Cohen, A., Kulkarni, A., Laney, J., Taylor, C., Kumar, V., Cladera,
  F., Cowley, A., Skandan, S., Lee, E., Lipschitz, L., Bhat, A., Rodrigues, N.,
  and Zhou, A. (2020).
\newblock Mine tunnel exploration using multiple quadrupedal robots.
\newblock {\em IEEE Robotics and Automation Letters}, PP:1--1.

\bibitem[NVIDIA, 2020]{TensorRT_2020}
NVIDIA (2020).
\newblock Tensorrt.
\newblock \url{https://github.com/NVIDIA/TensorRT}.

\bibitem[{Ohradzansky} et~al., 2020]{OhrMil20}
{Ohradzansky}, M.~T., {Mills}, A.~B., {Rush}, E.~R., {Riley}, D.~G., {Frew},
  E.~W., and {Sean Humbert}, J. (2020).
\newblock Reactive control and metric-topological planning for exploration.
\newblock In {\em 2020 IEEE International Conference on Robotics and Automation
  (ICRA)}, pages 4073--4079.

\bibitem[Otte et~al., 2017]{otte2017multi}
Otte, M., Kuhlman, M., and Sofge, D. (2017).
\newblock Multi-robot task allocation with auctions in harsh communication
  environments.
\newblock In {\em 2017 International Symposium on Multi-Robot and Multi-Agent
  Systems (MRS)}, pages 32--39. IEEE.

\bibitem[Parker, 1999]{Par99}
Parker, L.~E. (1999).
\newblock Adaptive heterogeneous multi-robot teams.
\newblock {\em Neurocomputing}, 28(1):75--92.

\bibitem[Parker et~al., 2016]{ParRus16}
Parker, L.~E., Rus, D., and Sukhatme, G.~S. (2016).
\newblock {\em Multiple Mobile Robot Systems}, pages 1335--1384.
\newblock Springer International Publishing.

\bibitem[Qin et~al., 2018]{Qin2018}
Qin, H., Meng, Z., Meng, W., Chen, X., Sun, H., Lin, F., and Jr, M. (2018).
\newblock Autonomous exploration and mapping system using heterogeneous uavs
  and ugvs in gps-denied environments.
\newblock {\em IEEE Transactions on Vehicular Technology}.

\bibitem[Recchiuto and Sgorbissa, 2018]{RecTom18}
Recchiuto, C.~T. and Sgorbissa, A. (2018).
\newblock Post-disaster assessment with unmanned aerial vehicles: A survey on
  practical implementations and research approaches.
\newblock {\em Journal of Field Robotics}, 35(4):459--490.

\bibitem[Sandler et~al., 2018]{Sandler18}
Sandler, M., Howard, A.~G., Zhu, M., Zhmoginov, A., and Chen, L. (2018).
\newblock Inverted residuals and linear bottlenecks: Mobile networks for
  classification, detection and segmentation.
\newblock {\em CoRR}, abs/1801.04381.

\bibitem[Shivakumar et~al., 2019]{shivakumar2019pst900}
Shivakumar, S.~S., Rodrigues, N., Zhou, A., Miller, I.~D., Kumar, V., and
  Taylor, C.~J. (2019).
\newblock Pst900: Rgb-thermal calibration, dataset and segmentation network.

\bibitem[Steindl et~al., 2020]{steindl_2020}
Steindl, R., Molnar, T., Talbot, F., Hudson, N., Tam, B., Murrell, S., and
  Kottege, N. (2020).
\newblock Bruce {-} design and development of a dynamic hexapod robot.
\newblock In {\em Australasian Conference on Robotics and Automation {(ACRA)}}.

\bibitem[Tiderko et~al., 2016]{Tiderko16}
Tiderko, A., Hoeller, F., and R{\"o}hling, T. (2016).
\newblock {\em The ROS Multimaster Extension for Simplified Deployment of
  Multi-Robot Systems}, pages 629--650.
\newblock Springer International Publishing, Cham.

\bibitem[{Tychsen-Smith} and {Petersson}, 2017]{Tychsen_Smith_2017}
{Tychsen-Smith}, L. and {Petersson}, L. (2017).
\newblock Denet: Scalable real-time object detection with directed sparse
  sampling.
\newblock In {\em 2017 IEEE International Conference on Computer Vision
  (ICCV)}, pages 428--436.

\bibitem[{Tychsen-Smith} and {Petersson}, 2018]{Tychsen_Smith_2018}
{Tychsen-Smith}, L. and {Petersson}, L. (2018).
\newblock Improving object localization with fitness nms and bounded iou loss.
\newblock In {\em 2018 IEEE/CVF Conference on Computer Vision and Pattern
  Recognition}, pages 6877--6885.

\bibitem[Ulam et~al., 2007]{ulam2007integrated}
Ulam, P., Endo, Y., Wagner, A., and Arkin, R. (2007).
\newblock Integrated mission specification and task allocation for robot
  teams-design and implementation.
\newblock In {\em Proceedings 2007 IEEE International Conference on Robotics
  and Automation}, pages 4428--4435. IEEE.

\bibitem[{Vechersky} et~al., 2018]{VecCox18}
{Vechersky}, P., {Cox}, M., {Borges}, P., and {Lowe}, T. (2018).
\newblock Colourising point clouds using independent cameras.
\newblock {\em IEEE Robotics and Automation Letters}, 3(4):3575--3582.

\bibitem[Williams et~al., 2020]{williams_2020}
Williams, J., Jiang, S., O’Brien, M., Wagner, G., Hernandez, E., Cox, M.,
  Pitt, A., Arkin, R., and Hudson, N. (2020).
\newblock Online {3D} frontier-based {UGV} and {UAV} exploration using direct
  point cloud visibility.
\newblock In {\em IEEE International Conference on Multisensor Fusion and
  Integration (MFI)}.

\bibitem[{Yamauchi}, 1997]{Yam97}
{Yamauchi}, B. (1997).
\newblock A frontier-based approach for autonomous exploration.
\newblock In {\em CIRA}, pages 146--151.

\bibitem[Zlot and Stentz, 2006]{zlot2006market}
Zlot, R. and Stentz, A. (2006).
\newblock Market-based multirobot coordination for complex tasks.
\newblock {\em The International Journal of Robotics Research}, 25(1):73--101.

\end{thebibliography}

\end{document}